\documentclass{article}
\usepackage{spconf,amsmath,graphicx,hyperref}

\usepackage[utf8]{inputenc} 
\usepackage[T1]{fontenc}    
\usepackage{hyperref}       
\usepackage{url}            
\usepackage{booktabs}       
\usepackage{amsfonts}       
\usepackage{nicefrac}       
\usepackage{microtype}      
\usepackage{xcolor}         
\usepackage{svg}
\svgpath{{figures/}} 

\usepackage{subfigure}

\usepackage{xcolor}
\usepackage{wrapfig}

\usepackage{titletoc}

\usepackage{amssymb}
\usepackage{mathtools}
\usepackage{amsthm}

\usepackage{amsmath,amsfonts,bm}

\def\eqref#1{equation~\ref{#1}}

\def\1{\bm{1}}

\DeclareMathAlphabet{\mathsfit}{\encodingdefault}{\sfdefault}{m}{sl}
\SetMathAlphabet{\mathsfit}{bold}{\encodingdefault}{\sfdefault}{bx}{n}

\usepackage{url}          
\usepackage{booktabs}    
\usepackage{nicefrac}    
\usepackage{microtype}   
\usepackage{xcolor}      

\usepackage{array}
\newcolumntype{C}{@{\extracolsep{3cm}}c@{\extracolsep{0pt}}}%

\usepackage{xcolor}

\definecolor{darkgreen}{rgb}{0.0, 0.5, 0.0}
\newcommand{\first}[1]{\textbf{\textcolor{red}{#1}}}
\newcommand{\second}[1]{\underline{\textcolor{blue}{#1}}}
\definecolor{mydarkgreen}{RGB}{0,120,0}

\usepackage{algorithm}
\usepackage{algorithmic}

\usepackage{amsmath}  
\usepackage{comment}  
\usepackage{graphicx}     
\usepackage{subcaption}  
\usepackage{caption}  
\usepackage{threeparttable}
\usepackage{amsfonts} 
\usepackage{enumitem}
\usepackage[table]{xcolor}
\usepackage{adjustbox}
\usepackage{tabularray}
\usepackage{nicematrix}
\usepackage{pifont}
\newcommand{\cmark}{\ding{51}}
\newcommand{\xmark}{\ding{55}}
\usepackage{multirow}

\usepackage{wrapfig}

\usepackage{tablefootnote}

\usepackage{titlesec} 

\title{Dataset-Driven Channel Masks in Transformers\\for Multivariate Time Series}
\name{
Seunghan Lee\textsuperscript{1,2*},
Taeyoung Park\textsuperscript{1$\dagger$},
Kibok Lee\textsuperscript{1$\dagger$}
\thanks{\textsuperscript{*} Work done at Yonsei University.}
\thanks{\textsuperscript{$\dagger$} Equal advising (co-corresponding authors).}
\thanks{This work was supported by National Research Foundation of Korea (NRF) grant funded by the Korea government (MSIT) (2020R1A2C1A01005949, 2022R1A4A1033384, RS-2023-00217705, RS-2024-00341749), the MSIT (Ministry of Science and ICT), Korea, under the ICAN (ICT Challenge and Advanced Network of HRD) support program (RS-2023-00259934) supervised by the IITP (Institute for Information \& Communications Technology Planning \& Evaluation), Yonsei University Research Fund (2024-22-0148).}
}

\address{
\textsuperscript{1} Department of Statistics and Data Science, Yonsei University \\
\textsuperscript{2} LG AI Research
}
\begin{document}
%
\maketitle
\begin{abstract}
Capturing channel dependency (CD) is essential for modeling multivariate time series (TS), and attention-based methods have been widely employed for this purpose.
Nonetheless, these methods
primarily focus on modifying the \textit{architecture}, often neglecting the importance of
\textit{dataset-specific} characteristics.
In this work, we introduce the concept of \textbf{partial channel dependence} (PCD) to enhance CD modeling in Transformer-based models by leveraging \textit{dataset-specific information}
to refine the CD captured by the model.
To achieve PCD, we propose \textbf{channel masks} (CMs), which are integrated into the attention matrices of Transformers via element-wise multiplication.
CMs consist of two components:
1)~a \textbf{similarity matrix} that captures relationships between the channels,
and 2)~dataset-specific and learnable \textbf{domain parameters} that refine the similarity matrix.
We validate the effectiveness of PCD across diverse tasks and datasets
with various backbones.
Code is available at this repository: \url{https://github.com/YonseiML/pcd}.
\end{abstract}


\section{Introduction}
Multivariate time series (MTS) consist of multiple interrelated channels and are prevalent in various real-world applications \cite{wei2019multivariate}.
MTS forecasting has been explored with two different strategies:
the \textit{channel-dependent} (CD) strategy~\cite{wang2024timemixer++,zhao2024rethinking,qi2024enhancing} and the \textit{channel-independent} (CI) strategy~\cite{zeng2023transformers,li2023revisiting,nie2022time,lee2023learning},
with the former emphasizing inter-channel dependencies, while the latter ignoring these dependencies and dealing with
channels individually.
These two strategies are known to have a capacity-robustness trade-off \cite{han2023capacity}, where CD models offer higher capacity but lower robustness, 
and CI models do vice versa.

Recently, 
with the rise of iTransformer \cite{liu2023itransformer},
various methods \cite{wang2024card, wang2024timemixer++} have 
adopted attention mechanisms \cite{vaswani2017attention} to capture CD.
Furthermore, with the emergence of large-scale TS datasets \cite{woo2024unified}, 
the CD strategy proves more effective than the CI strategy due to the capacity-robustness trade-off \cite{han2023capacity}, where the higher capacity of CD models benefits larger datasets.
In line with these trends, 
\textit{we highlight the importance of CD}, 
as CI can be derived from CD by 
ignoring 
unnecessary dependencies
when well captured.
Nonetheless, previous CD methods~\cite{wang2024card,yang2024vcformer,zhang2023crossformer,chen2023tsmixer,huang2023crossgnn}
have focused primarily on the \textit{model architecture}, 
neglecting
the 
importance
of the \textit{dataset}.

\begin{figure}[t]
\centering
\includegraphics[width=1.00\columnwidth]{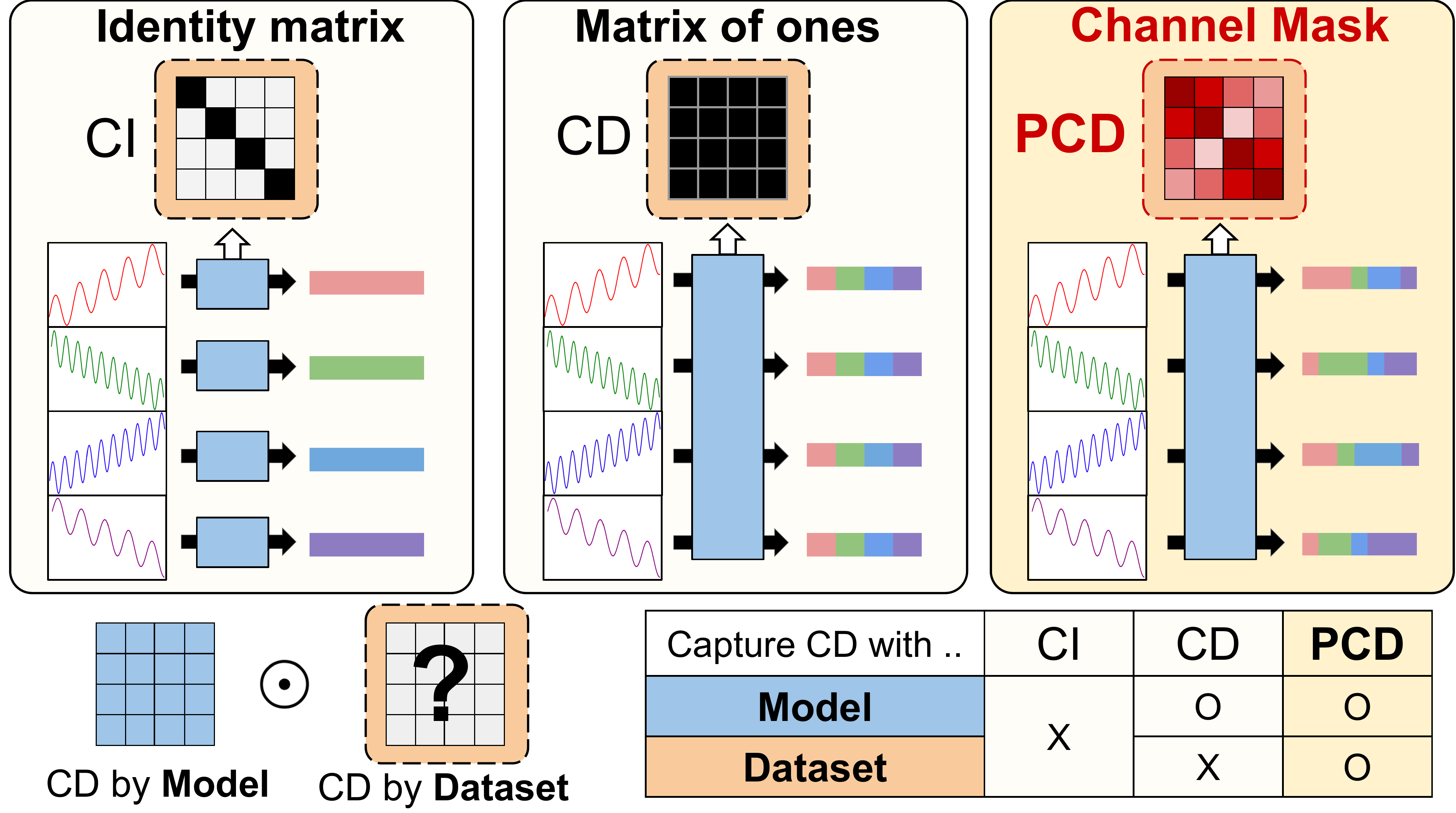}
\caption{
\textbf{CI vs. CD vs. PCD framework.} 
Under the 
\textbf{partial channel dependence (PCD)} framework,
CD captured by 
model 
is adjusted with
\textbf{channel mask} (i.e., CD captured by dataset).
}
\label{fig:intro1}
\end{figure}

\begin{figure*}[t]
\centering
\includegraphics[width=1.0\textwidth]{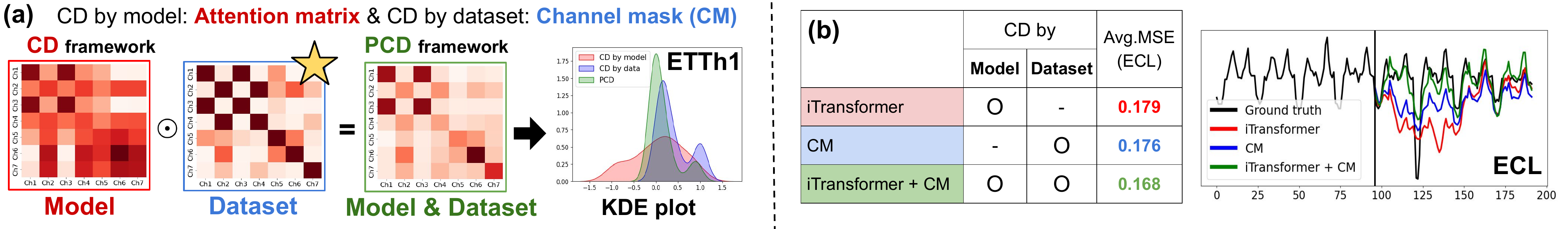} 
\caption{
\textbf{Necessity of CD by dataset.} (a) presents 
CDs
captured by 
\textcolor{red}{model}, 
\textcolor{blue}{dataset}, and 
\textcolor{darkgreen}{both}, 
along with their distributions, 
where \textit{CD by model} is adjusted with \textit{CD by dataset} within the PCD framework. 
(b) shows the TS forecasting results 
using 
\textcolor{red}{model} (iTransformer), 
\textcolor{blue}{dataset} (replacing attention matrix of iTransformer with CM), and 
\textcolor{darkgreen}{both} (iTransformer with CM),
highlighting the importance of leveraging the \textit{dataset} itself.
}
\label{fig:intro2}
\vspace{-5pt}
\end{figure*}

To this end, 
we introduce the concept of \textbf{partial channel dependence} (PCD), 
which improves CD modeling in Transformer-based models by incorporating \textit{dataset-specific} information.
Specifically, 
we propose a \textbf{channel mask} (CM) to achieve PCD, 
a \textit{dataset-specific} matrix that adjusts the CD captured by the model 
through element-wise multiplication with the attention matrix, as shown in Figure~\ref{fig:intro1} and Figure~\ref{fig:intro2} (a).
A CM containing dataset-specific information consists of 1) a \textbf{similarity matrix} between the channels, calculated from the \textit{entire dataset} in the data space
and 2) \textbf{domain parameters} specific to each dataset, which refine the similarity matrix to capture absolute dependencies.

Furthermore, 
we argue that 
PCD
is particularly crucial for TS foundation models (TSFMs) for two reasons: 
1) A TSFM is a \textit{single} model trained on \textit{multiple} datasets with varying CD, 
where each dataset may benefit from a different (CI or CD) approach
\cite{woo2024unified}, and
2) a CD architecture (e.g., attention mechanism) is essential for a TSFM due to the 
capacity-robustness trade-off \cite{han2023capacity}, 
as it is trained with \textit{large-scale} datasets.

The main contributions are summarized as:
\begin{itemize}[leftmargin=0.3cm,itemsep=-1pt,topsep=-3pt, partopsep=0pt]
    \item We introduce the concept of \textbf{partial channel dependence} (PCD), 
    where the CD captured by the Transformer-based model is 
    adjusted with the characteristics of the dataset.
    \item 
    We propose a \textbf{channel mask} (CM) to achieve PCD, consisting of 1) a \textit{similarity matrix} between the channels of the entire dataset and 2) \textit{domain parameters} 
    refining
    the similarity matrix.
    As a plug-in method, CM
    is element-wise multiplied with the attention matrix (i.e., CD captured by the model), making it applicable to any model that captures CD with an attention mechanism.
    \item We present extensive experiments across 
    five backbones, 
    including a \textcolor{black}{TSFM pretrained on multiple datasets}, showing consistent performance gains.
    For instance, applying CMs to iTransformer \cite{liu2023itransformer} results in gains across all 13 datasets, yielding an \textcolor{black}{average MSE improvement of 6.3\%}.
\end{itemize}

\section{Related Works}

\textbf{MTS forecasting models.}
For \textcolor{black}{CI models}, DLinear \cite{zeng2023transformers} and RLinear \cite{li2023revisiting} employ linear models along the time dimension, PatchTST \cite{nie2022time} divides TS into patches and feeds them into a Transformer in a CI manner,
and PITS \cite{lee2023learning} combines CI and patch independent architectures with multi-layer perceptrons (MLPs).
For \textcolor{black}{CD models}, Crossformer \cite{zhang2023crossformer} 
captures both temporal dependencies (TD) and CD using an attention mechanism,
TSMixer \cite{chen2023tsmixer} 
utilizes MLPs 
with patching 
to capture both dependencies,
and CrossGNN \cite{huang2023crossgnn} 
employs a linear complexity graph neural network to capture CD.
Beyond these methods, various methods have underscored the importance of capturing CD, where
LIFT  \cite{zhao2024rethinking} 
captures the lead-lag relationship between channels
and CDAM  \cite{qi2024enhancing} minimizes redundant information while enhancing relevant mutual information between channels.

\textbf{Transformer-based CD models.}
Recently, iTransformer  \cite{liu2023itransformer} inverts the traditional Transformer framework in TS domain by treating each channel as a token instead of each patch.
Following this framework, various methods have been proposed to capture CD with attention mechanism.
CARD \cite{wang2024card} employs channel-aligned attention to capture both TD and CD and
PRformer \cite{yu2024prformer} employs RNN and attention to capture TD and CD, respectively.
Minusformer \cite{liang2024minusformer} applies a Boosting ensemble learning to channel-wise attention.
VCformer \cite{yang2024vcformer} introduces a variable correlation attention module that modifies the standard attention.
While these works capture CDs, they either \textit{rely on input-window correlations} \cite{yang2024vcformer} or \textit{require architectural modifications} \cite{huang2023crossgnn}. 
Our approach differs by modeling CD at both local 
(input-dependent) 
and global 
(entire dataset level) 
scales within a general plug-in framework.

\begin{figure}[t]
\centering
\includegraphics[width=1\columnwidth]{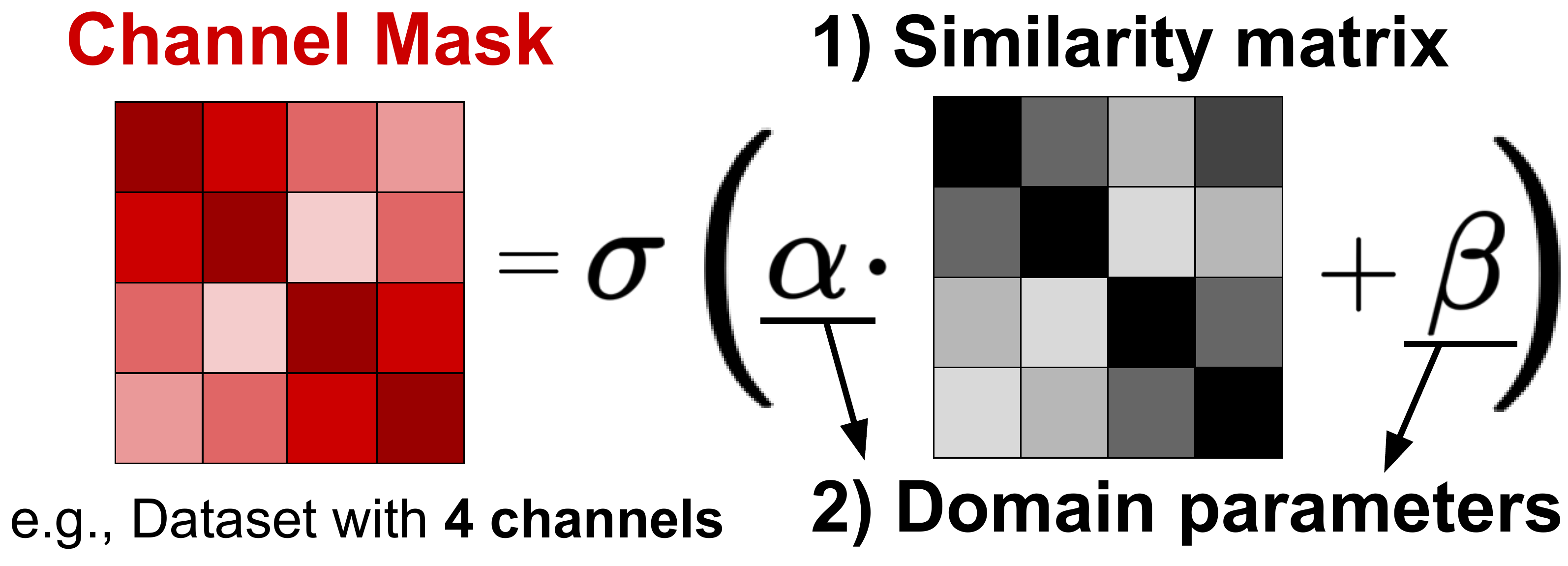}
\vspace{-15pt}
\caption{
\textbf{Channel Mask.} 
CM
consists of 1)~a \textit{similarity matrix} between channels and 2)~\textit{domain parameters} to refine the similarity matrix.
}
\label{fig:CM_component}
\vspace{-8pt}
\end{figure}

\section{Methodology}
In this section, we introduce a channel mask (CM), a simple yet effective method for achieving PCD.
A CM employs a 1)~\textit{similarity matrix}
to capture CD with the entire dataset in the data space
and 2)~\textit{domain parameters} that refine the matrix to learn absolute dependencies specific to each dataset.

\begin{figure*}[t]
\centering
\includegraphics[width=1\textwidth]{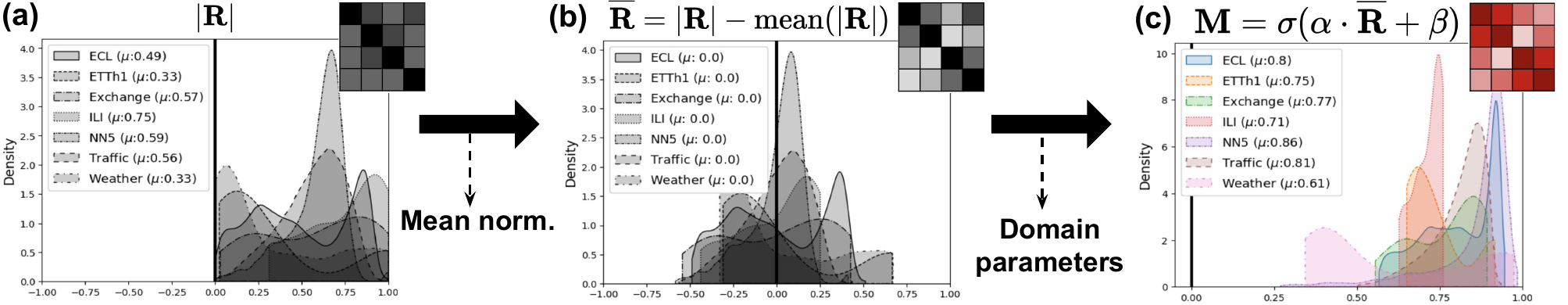} 
\caption{
\textbf{Necessity of domain parameters.}
As similarity metric is a relative measure depending on the dataset,
we employ domain parameters to adjust similarity matrix.
Specifically, we refine the matrix 
with 1) mean normalization and 2) domain parameters,
resulting in $\mathbf{M} = \sigma(\alpha \cdot \bar{\mathbf{R}} + \beta)$.
}
\label{fig:adjust}
\end{figure*}

\subsection{Channel Mask: CD by Dataset}
\label{sec:pi_task}
As shown in Figure~\ref{fig:CM_component}, 
a CM consists of two components: 1) similarity matrix ($\mathbf{R}$) between channels, and 2) 
domain parameters ($\alpha$ and $\beta$), which scale and shift the matrix according to the dataset's characteristics, along with a sigmoid function to normalize the values between 0 and 1.

\textbf{1) Similarity matrix.}
Correlation measures the relationships between channels and has been used in previous works to analyze CD \cite{yang2024vcformer,zhao2024rethinking}.
\textcolor{black}{Building on these approaches, we employ a correlation matrix ($\mathbf{R}$) as a similarity matrix.} 
However, strong relationhip implies either highly positive 
or highly negative correlation, as the values of correlation range from -1 to 1, implying no relationship with correlation of 0.
To address this issue, we use the absolute value of the matrix $|\mathbf{R}|$.
Robustness to the choice of similarity metric 
is discussed in Table~\ref{tbl:metric_abl}.

\textbf{2) Domain parameters.}
Although a similarity matrix captures dataset-specific information as prior knowledge using the statistics of a dataset,
we argue that the matrix alone might be insufficient for modeling a CM for the following two reasons:
\begin{itemize}[leftmargin=0.3cm,itemsep=1pt,topsep=-2pt, partopsep=2pt]
    \item \textbf{a) Relative measure.} 
    Similarity is a dataset-dependent measure, where different datasets exhibit different distributions of elements of $|\mathbf{R}|$, as shown in shown in Figure~\ref{fig:adjust}(a).
    \item 
    \textbf{b) Varying dataset characteristics.} 
    The relationship between correlation and CD may \textit{vary across datasets} (i.e., a same correlation can correspond to different levels of CD). 
\end{itemize}

To handle the former (i.e., relative measure), we normalize $|\mathbf{R}|$ by subtracting the mean value, resulting in $\bar{\mathbf{R}}$, as shown in Figure~\ref{fig:adjust}(b).
To handle the latter (i.e., varying datasets), 
we introduce two learnable domain parameters ($\alpha$ and $\beta$) for each dataset to refine $|\mathbf{R}|$ with affine transformation.
Using these parameters along with a sigmoid function, we model a CM 
as $\mathbf{M} = \sigma(\alpha\cdot\bar{\mathbf{R}} + \beta)$,
as shown in Figure~\ref{fig:adjust}(c).
We argue that using parameters to address 
the discrepancies 
is particularly crucial for a TSFM since it is trained on \textit{multiple} datasets, as discussed in Tables \ref{tbl:cdr_comparison} and \ref{tbl:abl_CM}.

\subsection{CD by Model \& CD by Dataset}
The proposed CM adjusts the
CD captured by the model 
by performing element-wise multiplication with the 
attention matrix,
with the general adjustment
modeled by $\mathbf{A}$:
\begin{equation}
\begin{aligned}
\operatorname{Attn}(\mathbf{Q}, \mathbf{K}, \mathbf{V}) &= \operatorname{Softmax}\left(\mathbf{A} \odot \frac{\mathbf{Q} \mathbf{K}^{\top}}{\sqrt{d_k}}\right) \cdot \mathbf{V}, \\
\text{where } \mathbf{A} &= \begin{cases}
\mathbf{I}_{C \times C} & \text{if CI,} \\
\mathbf{1}_{C \times C} & \text{if CD,} \\
\mathbf{M} = \sigma(\alpha \cdot \mathbf{\bar{R}} + \beta) & \text{if PCD,}
\end{cases}
\end{aligned}
\label{eq:main}
\end{equation}
and $C$ is the number of channels. Note that Equation~\ref{eq:main} incorporates both CI and CD frameworks within a single expression. As shown in Figure~\ref{fig:intro1},
$\mathbf{A}$ is the identity matrix ($\mathbf{I}_{C \times C}$) in the CI framework, 
while $\mathbf{A}$ is a matrix of ones ($\mathbf{1}_{C \times C}$) in the CD framework.
In contrast, PCD framework represents it as $\mathbf{M} = \sigma(\alpha \cdot \bar{\mathbf{R}} + \beta)$, enabling a more refined adjustment of CD tailored to the dataset.

\begin{table*}[t]
\centering
\begin{subtable}
\centering
\begin{adjustbox}{max width=1.0\textwidth}
\begin{NiceTabular}{c|cccc|c||cccc|c||cccc|c}
\toprule
\multirow{2.5}{*}{Average 4$H$s} 
& \multicolumn{2}{c}{\textbf{iTransformer}}  & \multicolumn{2}{c}{+ CM} & \multirow{2.5}{*}{\shortstack{ Imp. \\(MSE)}}
& \multicolumn{2}{c}{\textbf{CARD$^{L=96}$}}  & \multicolumn{2}{c}{+ CM} & 
\multirow{2.5}{*}{\shortstack{ Imp. \\(MSE)}}
& \multicolumn{2}{c}{\textbf{CARD$^{L=720}$}}  & \multicolumn{2}{c}{+ CM} & 
\multirow{2.5}{*}{\shortstack{ Imp. \\(MSE)}}
\\

\cmidrule(lr){2-3} \cmidrule(lr){4-5} 
\cmidrule(lr){7-8} \cmidrule(lr){9-10} 
\cmidrule(lr){12-13} \cmidrule(lr){14-15} 

 & MSE & MAE & MSE & MAE &  & MSE & MAE & MSE & MAE &  & MSE & MAE & MSE & MAE &  \\
\toprule

ETTh1 
& {0.457} & {0.449} &  \first{0.444} & \first{0.441}& \cellcolor{gray!20} \first{2.8\%} 
& 0.444 & 0.429 & \first{0.438} & \first{0.425} &\cellcolor{gray!20} \first{1.4\%} & 0.406 & 0.428 & \first{0.405} & \first{0.427} & \cellcolor{gray!20} \first{0.2\%}
\\
ETTh2 
&  {0.384} & {0.407} &  \first{0.383} &  \first{0.406}&\cellcolor{gray!20}  \first{0.3\%}
& 0.360 & 0.387 & \first{0.360} & \first{0.387} &\cellcolor{gray!20} \first{0.0\%}  & 0.330 & 0.378 & \first{0.328} & \first{0.377} & \cellcolor{gray!20} \first{0.6\%}
\\
ETTm1 
& {0.408} & {0.412} &  \first{0.398} &  \first{0.406}&\cellcolor{gray!20}  \first{2.5\%} 
& 0.383 & 0.382 & \first{0.379} & \first{0.381} &\cellcolor{gray!20} \first{1.1\%} & 0.350 & 0.370 & \first{0.347} & \first{0.369} & \cellcolor{gray!20} \first{0.9\%}
\\
ETTm2 
& {0.293} & {0.337} &  \first{0.289} &  \first{0.335}&\cellcolor{gray!20}  \first{1.4\%} 
& 0.272 & 0.314 & \first{0.270} & \first{0.313} &\cellcolor{gray!20} \first{0.7\%} & 0.252 & 0.309 & \first{0.250} & \first{0.308} & \cellcolor{gray!20} \first{0.8\%}
\\
PEMS03 
& {0.142} & {0.248} &  \first{0.124} &  \first{0.231}&\cellcolor{gray!20}  \first{12.7\%} 
& 0.239 & 0.329 & \first{0.221} & \first{0.318} &\cellcolor{gray!20} \first{7.6\%} & 0.116 & 0.220 & \first{0.112} & \first{0.216} & \cellcolor{gray!20} \first{3.4\%}
\\
PEMS04 
&  {0.121} & {0.232} &  \first{0.098} &  \first{0.210} &\cellcolor{gray!20}  \first{19.0\%} 
& 0.276 & 0.353 & \first{0.258} & \first{0.339} &\cellcolor{gray!20} \first{6.5\%} & 0.120 & 0.222 & \first{0.109} & \first{0.210} & \cellcolor{gray!20} \first{9.2\%}
\\
PEMS07 
& {0.102} & {0.205} &  \first{0.082} &  \first{0.183}&\cellcolor{gray!20}  \first{19.6\%} 
& 0.210 & 0.305 & \first{0.208} & \first{0.303} &\cellcolor{gray!20} \first{1.0\%} & 0.087 & 0.190 & \first{0.080} & \first{0.181} & \cellcolor{gray!20} \first{8.0\%}
\\
PEMS08 
& {0.254} & {0.306} &  \first{0.152} &  \first{0.231}&\cellcolor{gray!20}  \first{40.2\%} 
& 0.302 & 0.358 & \first{0.271} & \first{0.336} &\cellcolor{gray!20} \first{9.6\%} & 0.148 & 0.213 & \first{0.146} & \first{0.209} & \cellcolor{gray!20} \first{1.2\%}
\\
Exchange 
& {0.368} & {0.409} &  \first{0.363} &  \first{0.406}&\cellcolor{gray!20}  \first{1.4\%} 
& 0.370 & 0.407 & \first{0.365} & \first{0.404} &\cellcolor{gray!20} \first{1.4\%} & 0.425 & 0.448 & \first{0.381} & \first{0.422} & \cellcolor{gray!20} \first{10.4\%}
\\
Weather 
& {0.260} & {0.281} &  \first{0.250} & \first{0.275}&\cellcolor{gray!20}  \first{3.8\%} 
& 0.240 & 0.262 & \first{0.238} & \first{0.261} &\cellcolor{gray!20} \first{0.8\%} & 0.226 & 0.255 & \first{0.220} & \first{0.252} & \cellcolor{gray!20} \first{2.7\%}
\\
Solar 
& {0.234} & {0.261} & \first{0.228} &  \first{0.258}&\cellcolor{gray!20}  \first{2.6\%} 
& 0.304 & 0.287 & \first{0.296} & \first{0.281} &\cellcolor{gray!20} \first{2.6\%} & 0.211 & 0.243 & \first{0.204} & \first{0.231} & \cellcolor{gray!20} \first{3.3\%}
\\
\cmidrule{12-16}
ECL 
& {0.179} & {0.270} &  \first{0.168} &  \first{0.262}&\cellcolor{gray!20}  \first{6.1\%} 
& 0.177 & 0.263 & \first{0.171} & \first{0.259} & \cellcolor{gray!20} \first{3.4\%} & \multicolumn{5}{c}{\multirow{2}{*}{Out of Memory}}  
\\
Traffic 
& {0.428} & {0.282} & \first{0.422} &  \first{0.281}&\cellcolor{gray!20}  \first{1.4\%} 
& 0.453 & 0.281 & \first{0.436} & \first{0.268} & \cellcolor{gray!20} \first{3.8\%} & \multicolumn{5}{c}{ }
\\ 
\midrule
Average & \rowcolor{yellow!20} 0.279 & 0.315 &  \first{0.261} &  \first{0.302}  & \cellcolor{gray!20} 
 \first{6.3\%} 
 & 0.310 & 0.335 & \first{0.301} & \first{0.329} & \cellcolor{gray!20} \first{3.0\%} & 0.243 & 0.298 & \first{0.234} & \first{0.291} & \first{3.7\%}
 \\
\midrule
 Best Count  &  \rowcolor{yellow!20}  2 & 2 & \first{50} & \first{51} & - &  
 9 & 11 & \first{49} & \first{48} & - & 2 & 7 & \first{44} & \first{44} & - \\
\bottomrule
\toprule
\multirow{2.5}{*}{Average 4$H$s} 
& \multicolumn{2}{c}{\textbf{PRformer}}  & \multicolumn{2}{c}{+ CM} & \multirow{2.5}{*}{\shortstack{ Imp. \\(MSE)}}
& \multicolumn{2}{c}{\textbf{Minusformer$^{L=96}$}}  & \multicolumn{2}{c}{+ CM} & 
\multirow{2.5}{*}{\shortstack{ Imp. \\(MSE)}}
& \multicolumn{2}{c}{\textbf{Minusformer$^{L=336}$}}  & \multicolumn{2}{c}{+ CM} & 
\multirow{2.5}{*}{\shortstack{ Imp. \\(MSE)}}
\\

\cmidrule(lr){2-3} \cmidrule(lr){4-5} 
\cmidrule(lr){7-8} \cmidrule(lr){9-10} 
\cmidrule(lr){12-13} \cmidrule(lr){14-15} 

 & MSE & MAE & MSE & MAE &  & MSE & MAE & MSE & MAE &  & MSE & MAE & MSE & MAE &  \\
\toprule
ETTh1
&  0.444 & 0.445 & \first{0.434} & \first{0.436} & \cellcolor{gray!20} \first{2.3\%}
& 0.463 & 0.452 & \first{0.451} & \first{0.446} & \cellcolor{gray!20} \first{2.6\%} 
& 0.453 & 0.453 & \first{0.442} & \first{0.448} & \cellcolor{gray!20} \first{2.4\%} 
\\
ETTh2
&  0.345 & 0.383 & \first{0.340} & \first{0.381} & \cellcolor{gray!20} \first{1.4\%}
&  0.394 & 0.409 & \first{0.385} & \first{0.406} & \cellcolor{gray!20} \first{2.3\%} 
&  0.360 & 0.397 &  \first{0.356} & \first{0.395} & \cellcolor{gray!20} \first{1.1\%} 
\\
ETTm1
&  0.352 & 0.376 & \first{0.343} & \first{0.371} & \cellcolor{gray!20} \first{2.6\%}
& 0.416 & 0.412 & \first{0.404} & \first{0.408} & \cellcolor{gray!20} \first{1.9\%} 
& 0.369 & 0.393 &  \first{0.361} & \first{0.389} & \cellcolor{gray!20} \first{2.2\%} 
\\
ETTm2
&  0.266 & 0.310 & \first{0.262} & \first{0.317} & \cellcolor{gray!20} \first{1.5\%}
& 0.285 & 0.328 & \first{0.283} & \first{0.327} & \cellcolor{gray!20} \first{0.7\%} 
& 0.275 & 0.328 &  \first{0.263} & \first{0.322} & \cellcolor{gray!20} \first{4.4\%} 
\\
PEMS03
&  0.152 & 0.254 & \first{0.143} & \first{0.248} & \cellcolor{gray!20} \first{5.9\%}
& 0.138 & 0.245 & \first{0.114} & \first{0.223} & \cellcolor{gray!20} \first{17.4\%}
& 0.087 & 0.190 &  \first{0.086} & \first{0.188} & \cellcolor{gray!20} \first{1.1\%} 
\\
PEMS04
& 0.185 & 0.281 & \first{0.173} & \first{0.272} & \cellcolor{gray!20} \first{6.5\%}
& 0.171 & 0.270 & \first{0.120} & \first{0.230} & \cellcolor{gray!20} \first{29.8\%} 
& 0.100 & 0.200 &  \first{0.093} & \first{0.195} & \cellcolor{gray!20} \first{7.0\%} 
\\
PEMS07
&  0.140 & 0.239 & \first{0.132} & \first{0.230} & \cellcolor{gray!20} \first{5.7\%}
& 0.125 & 0.224 & \first{0.089} & \first{0.191} & \cellcolor{gray!20} \first{28.8\%} 
& 0.068 & 0.165 &  \first{0.067} & \first{0.164} & \cellcolor{gray!20} \first{1.5\%} 
\\
PEMS08
&  0.205 & 0.283 & \first{0.186} & \first{0.271} & \cellcolor{gray!20} \first{9.3\%}
& 0.170 & 0.254 & \first{0.142} & \first{0.234} & \cellcolor{gray!20} \first{22.4\%} 
& 0.111 & 0.198 &  \first{0.109} & \first{0.191} & \cellcolor{gray!20} \first{1.8\%} 
\\
Exchange
& 0.565 & 0.491 & \first{0.472} & \first{0.449} & \cellcolor{gray!20} \first{16.5\%}
& 0.508 & 0.488 & \first{0.465} & \first{0.472} & \cellcolor{gray!20} \first{8.5\%} 
& 0.533 & 0.529 &  \first{0.474} & \first{0.500} & \cellcolor{gray!20} \first{11.1\%} 
\\
Weather
&  0.224 & 0.256 & \first{0.219} & \first{0.251} & \cellcolor{gray!20} \first{2.2\%}
&  0.260 & 0.281 & \first{0.252} & \first{0.275} & \cellcolor{gray!20} \first{3.1\%} 
&  0.242 & 0.275 &  \first{0.235} & \first{0.270} & \cellcolor{gray!20} \first{2.9\%} 
\\
Solar
&  0.202 & 0.218 & \first{0.197} & \first{0.212} & \cellcolor{gray!20} \first{2.5\%}
& 0.230 & 0.253 & \first{0.228} & \first{0.252} & \cellcolor{gray!20} \first{0.9\%} 
& 0.215 & 0.258 &  \first{0.212} & \first{0.255} & \cellcolor{gray!20} \first{1.4\%} 
\\
ECL
&  0.155 & 0.246 & \first{0.149} & \first{0.242} & \cellcolor{gray!20} \first{3.9\%}
& 0.171 & 0.262 & \first{0.164} & \first{0.258} & \cellcolor{gray!20} \first{4.1\%} 
& 0.161 & 0.254 &  \first{0.157} & \first{0.252} & \cellcolor{gray!20} \first{2.5\%} 
\\
Traffic
&  0.378 & 0.236 & \first{0.348} & \first{0.227} & \cellcolor{gray!20} \first{7.9\%} 
& 0.413 & 0.272 & \first{0.405} & \first{0.268} & \cellcolor{gray!20} \first{1.9\%} 
& 0.373 & 0.260 &  \first{0.366} & \first{0.259} & \cellcolor{gray!20} \first{1.9\%} 
\\
\midrule
Average  &  \rowcolor{yellow!20}  0.278 & 0.309 & \first{0.261} & \first{0.301} & \first{6.1\%} 
& 0.288 & 0.319 & \first{0.269} & \first{0.307} & \first{6.6\%} 
& 0.258 & 0.300 & \first{0.248} & \first{0.295} & \first{3.9\%} \\
\midrule
 Best Count  &  \rowcolor{yellow!20}  
 3 & 4  & 51 & 49 & - &
 4 & 8 & \first{51} & \first{49} & -  &
 6 & 8 & \first{46} & \first{47} & - \\
\bottomrule
\end{NiceTabular}
\end{adjustbox}
\label{tbl:full_fcst2}
\end{subtable}
\caption{
\textbf{Results of TS forecasting.} 
We apply our method to various backbones with 
varying input window sizes ($L$) and four horizons ($H$) over 13 datasets,
yielding consistent performance gains.
}
\label{tbl:main}
\end{table*}

\begin{figure}[t]
\centering
\includegraphics[width=1.00\columnwidth]{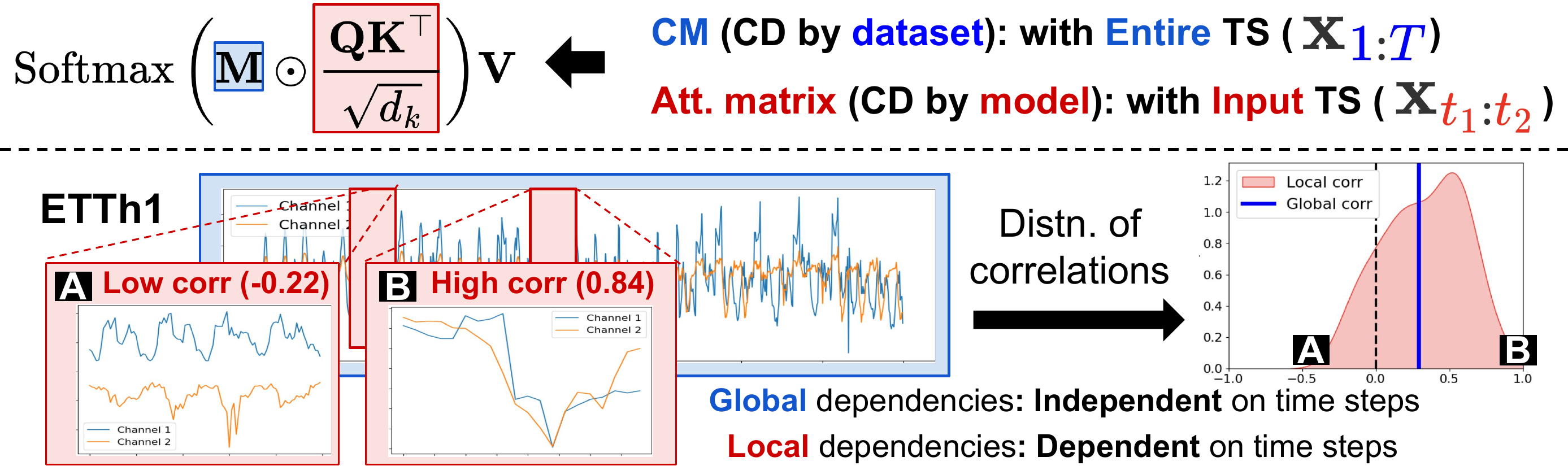} 
\caption{\textbf{Global \& local dependencies.}
CM and attention matrix are complementary in capturing global and local dependencies, respectively, since CM is constructed from the \textit{entire TS}, while the attention matrix is computed from the \textit{input TS} (segment of the entire TS).
}
\label{fig:global_local}
\vspace{-5pt}
\end{figure}
\textbf{Global \& local dependencies.}
As a similarity matrix is 
calculated based on the \textit{entire TS}, 
CM (i.e., CD by dataset), denoted by $\mathbf{M}$,
\textcolor{black}{captures the \textit{global} dependencies shared across all time steps.}
This complements the attention matrix ($\mathbf{Q} \mathbf{K}^{\top}/\sqrt{d_k}$) (i.e., CD by model),
which is calculated based on the \textit{input TS} and thus captures the \textit{local} dependencies which vary by input time step.
As shown in Figure~\ref{fig:global_local}, 
PCD framework captures both 
dependencies
through the element-wise multiplication of a CM and an attention matrix.
Furthermore, the figure illustrates two channels from ETTh1 \cite{zhou2021informer}, showing that even with the existence of global dependencies (red line), the dependency vary across time steps, emphasizing the need to capture both 
dependencies.
Further analysis on 
the importance of 
capturing both 
dependencies
is discussed in Table~\ref{tbl:global_local}.

\subsection{Channel Dependence Ratio}
\begin{figure}[t]
\centering
\includegraphics[width=1.00\columnwidth]{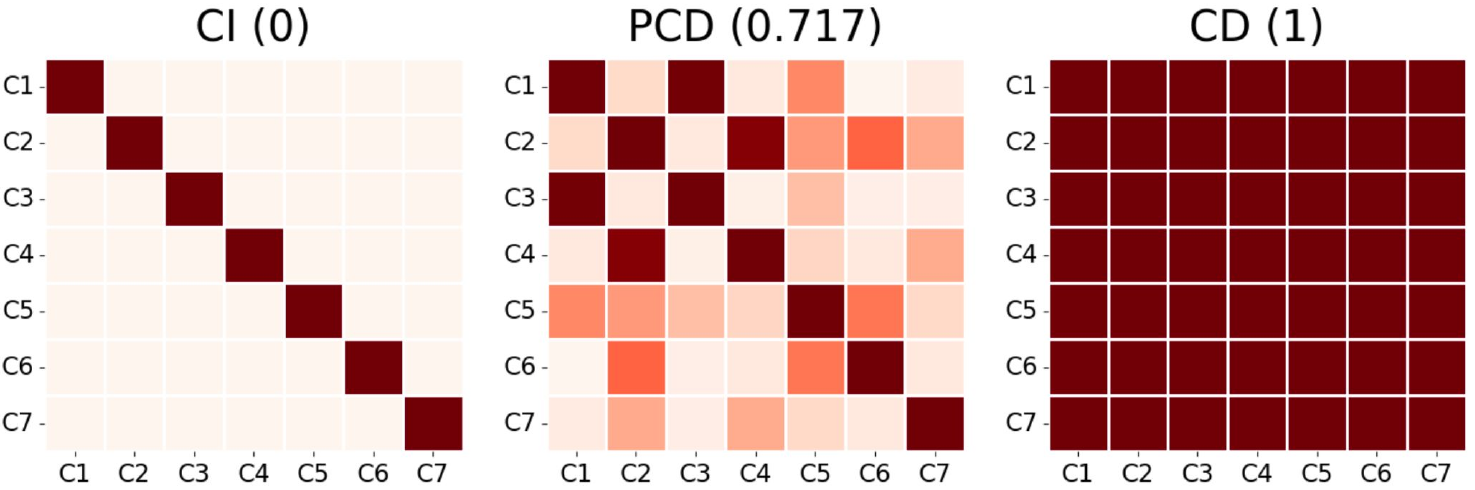} 
\caption{
CD ratio of CI, PCD, and CD.
}
\label{fig:CD_ratio_etth2}
\vspace{-8pt}
\end{figure}

To quantify the degree of CD for each dataset trained with the same model, we propose measuring the \textit{channel dependence ratio} (CD ratio), a metric based on a CM.
The CD ratio of $\mathbf{M}$, 
denoted as $r(\mathbf{M})$,
is 
the average of the off-diagonal elements 
of $\mathbf{M}$, 
excluding the autocorrelations of their respective channels.
This metric yields 
a value of 0 for CI cases 
and 1 for CD cases,
with higher values indicating a greater preference for 
channel interactions.
Note that it is a \textit{relative} metric rather than an absolute one, as it is designed to compare preferences for CD across multiple datasets with a single model.

Figure~\ref{fig:CD_ratio_etth2}
shows the visualization of $\mathbf{M}$ and its corresponding CD ratio for ETTh1 \cite{zhou2021informer}, 
with a ratio of 0.717 for PCD. 
We find that $\mathbf{M}$ effectively captures the degree of CD for each dataset, 
as datasets with higher $r(\mathbf{M})$ tend to have greater performance gains 
with CD architecture compared to CI architecture, as illustrated in Figure~\ref{fig:CD_gain}.

\section{Experiments}
We demonstrate the effectiveness of our method
in both \textcolor{black}{single-task models and TSFMs}
under supervised (SL) or self-supervised (SSL) settings, 
as:
\begin{itemize}[leftmargin=0.3cm,itemsep=1pt,topsep=-2pt, partopsep=2pt]
    \item 1) \textcolor{black}{Single-task model (SL)}: iTransformer \cite{liu2023itransformer}, CARD \cite{wang2024card}, PRformer \cite{yu2024prformer}, MinusFormer \cite{liang2024minusformer}
    \item 2) \textcolor{black}{Single-task model (SSL)}: TimeSiam \cite{dong2024timesiam} (discussed in Appendix \ref{sec:timesiam})
    \item 3) \textcolor{black}{TSFM (SL, SSL)}: UniTS \cite{gao2024units}
\end{itemize}

For UniTS, we consider four different tasks: forecasting (FCST), classification (CLS), imputation (IMP), and anomaly detection (AD), 
across various dataset sizes
including few-shot and zero-shot settings.
As evaluation metrics, we use the mean squared error (MSE) and mean absolute error (MAE) for FCST and IMP,
accuracy (Acc.) for CLS,
and F$_1$ score for AD.
Dataset statistics and implementation details 
can be found in Appendix~\ref{sec:data} and \ref{sec:impl}, respectively.

\begin{figure*}[t]    
\centering
\begin{minipage}{1\textwidth}
\centering
\begin{adjustbox}{max width=1.00\textwidth}
\begin{NiceTabular}{c|cc|c|c||c|cc|c||cc|c}
\toprule
\multicolumn{5}{c}{[1] Forecasting \& [2] Classification} & \multicolumn{4}{c}{[3] Imputation} & \multicolumn{3}{c}{[4] Anomaly Detection} \\
\midrule
\midrule
Ratio &  \multicolumn{2}{c}{Model}  & MSE   & Acc. & Ratio  &  \multicolumn{2}{c}{Model} &   MSE & \multicolumn{2}{c}{Model} &  \multicolumn{1}{c}{$\text{F}_1$}\\
\midrule
\multirow{6}{*}{5\%}   & iTransformer & FT & 0.598  &  51.4 & \multirow{6}{*}{25\%} & iTransformer &  FT &  0.186 & Anomaly Trans. & -  & 79.2  \\ 
\cmidrule{2-5} \cmidrule{7-9}
& \multirow{2}{*}{UniTS} & PT  & \cellcolor{yellow!20} 0.549  &  \cellcolor{yellow!20} 49.4 & & \multirow{2}{*}{UniTS} & PT  &  \cellcolor{yellow!20} 0.179 & TimesNet & FT  & 74.2 \\
& & FT  & \cellcolor{yellow!20} \textcolor{blue}{\underline{0.505}}  & \cellcolor{yellow!20} 53.8 & &   & FT  &  \cellcolor{yellow!20} \textcolor{blue}{\underline{0.167}} & PatchTST & FT  &  84.3 \\
\cmidrule{2-5} \cmidrule{7-9}
& \multirow{2}{*}{UniTS + CM} & PT  &  \cellcolor{yellow!20} 0.546  &  \cellcolor{yellow!20} \textcolor{red}{\textbf{54.9}} & & \multirow{2}{*}{UniTS + CM}  & PT  &  \cellcolor{yellow!20}  0.179 & iTransformer & FT   &  83.1 \\
\cmidrule{10-12}
& & FT  & \cellcolor{yellow!20} \textcolor{red}{\textbf{0.489}} & \cellcolor{yellow!20} \textcolor{blue}{\underline{54.8}} & &   & FT  &   \cellcolor{yellow!20} \textcolor{red}{\textbf{0.158}} & \multirow{2}{*}{UniTS}  & PT  &  \cellcolor{yellow!20} 81.7 \\
\cmidrule{1-5} \cmidrule{6-9}
\multirow{6}{*}{20\%} & iTransformer & FT & 0.510 & 59.9 & \multirow{7}{*}{50\%}  & iTransformer & FT &  0.226 & & FT  &  \cellcolor{yellow!20} \textcolor{blue}{\underline{85.6}} \\
\cmidrule{2-5} \cmidrule{7-9} \cmidrule{10-12}
& \multirow{2}{*}{UniTS} & PT  & \cellcolor{yellow!20} 0.525 &  \cellcolor{yellow!20} 58.9 & & \multirow{2}{*}{UniTS}  & PT  &  \cellcolor{yellow!20} 0.232  & \multirow{2}{*}{UniTS + CM} & PT  &  \cellcolor{yellow!20} 82.0 \\
& & FT  & \cellcolor{yellow!20} 0.486 &  \cellcolor{yellow!20} \textcolor{blue}{\underline{63.6}} & &   & FT  &  \cellcolor{yellow!20} \textcolor{blue}{\underline{0.213}} & & FT  &  \cellcolor{yellow!20} \textcolor{red}{\textbf{86.6}} \\
\cmidrule{2-5} \cmidrule{7-9} \cmidrule{10-12}
& \multirow{2}{*}{UniTS + CM} & PT & \cellcolor{yellow!20} \textcolor{blue}{\underline{0.453}}  & \cellcolor{yellow!20} 60.0  & & \multirow{2}{*}{UniTS + CM}  & PT  &  \cellcolor{yellow!20} 0.225 & \multicolumn{3}{>{\columncolor{gray!20}}c}{} \\
& & FT &  \cellcolor{yellow!20} \textcolor{red}{\textbf{0.425}}  & \cellcolor{yellow!20} \textcolor{red}{\textbf{64.8}} & &   & FT  &   \cellcolor{yellow!20} \textcolor{red}{\textbf{0.201}} & \multicolumn{3}{>{\columncolor{gray!20}}c}{ }\\
\bottomrule
\end{NiceTabular}
\end{adjustbox}
\captionsetup{type=table}
\caption{\textbf{Few-shot.} 
The table presents the results of four few-shot tasks
under both PT (prompt-tuning) and FT (fine-tuning) settings, 
reporting the average performance on
[1] nine forecasting, [2] six classification, [3] six imputation, and [4] four anomaly detection tasks.
}
\label{tbl:few_total}
\end{minipage}
\end{figure*}

\begin{table}[t]
\centering
\begin{adjustbox}{max width=1.00\columnwidth}
\begin{NiceTabular}{c|c|cc|c}
\toprule 
\multicolumn{2}{c}{ }  & UniTS & + CM & Imp. \\
\midrule
\multirow{2.5}{*}{\shortstack{Forecasting\\(MSE)}} & Supervised & 0.469 & \cellcolor{yellow!20} \textcolor{red}{\textbf{0.445}} & \first{5.1\%} \\ 
\cmidrule{2-5}
&  Prompt-tuning & 0.478 & \cellcolor{yellow!20} \textcolor{red}{\textbf{0.452}} & \first{5.4\%} \\ 
\midrule
\multirow{2.5}{*}{\shortstack{Classification\\(Acc.)}} & Supervised & 80.6 &\cellcolor{yellow!20} \textcolor{red}{\textbf{82.0}} & \first{1.7\%} \\ 
\cmidrule{2-5}
&  Prompt-tuning & 75.1 & \cellcolor{yellow!20} \textcolor{red}{\textbf{78.3}} & \first{4.3\%}  \\
\bottomrule
\end{NiceTabular}
\end{adjustbox}
\caption{20 FCST and 18 CLS tasks.}
\label{tbl:exp1_summary}
\vspace{-6pt}
\end{table}

\subsection{Application to Single-task Models}
To demonstrate the effectiveness of CM, 
we apply it to 
four backbones\footnote{We use the official codes of each method for reproducibility.} 
capturing CD with Transformer
for forecasting tasks on 13 datasets.
Table~\ref{tbl:main} presents the average 
performance
across four horizons ($H$), demonstrating consistent 
gains across datasets and backbones 
with varying input window sizes ($L$) based on the original paper.
Note that Minusformer selectively employs CI strategy for small datasets in its official code, whereas we use CD strategy for all datasets
for fair comparisons.
Additionally, we remove the dynamic projection module \cite{zhu2021long} from CARD, 
which generates summarized channel tokens, 
enabling 
application of CM
without significantly impacting performance.

\subsection{Application to TSFM}
\label{sec:app_TSFM}
To validate the effectiveness of our method on a TSFM, 
we apply it to UniTS, which solves diverse tasks 
without the need for fine-tuning, relying solely on prompt-tuning.

\textbf{a) Forecasting \& classification.} Table~\ref{tbl:exp1_summary} presents a summary of the results from 20 forecasting tasks and 18 classification tasks under both supervised (Sup.) and prompt-tuning (PT) settings,
with the full results 
provided in Appendix~\ref{tbl:exp1_FCST} and Appendix~\ref{sec:units1}, respectively.
The results indicate that applying our method 
enhances performance across both tasks.
Notably, our method even outperforms single-task models that are individually trained for each task (dataset).
Additionally, compared to GPT4TS \cite{zhou2023one}, 
applying our method to UniTS achieves superior performance with less than 1\% of the parameters (164.5M vs. 1.57M).

\textbf{b) Few-shot learning.}
For the tasks under the few-shot settings, 
we conduct four different tasks 
(FCST, CLS, IMP, AD),
following the experimental settings of UniTS. 
For (1) \textit{FCST and CLS}, 
we experiment 9 FCST tasks and 6 CLS tasks 
with data ratios of 5\% and 20\%. 
For (2) \textit{IMP}, 
we experiment 6 IMP tasks 
with a data ratio of 10\%, 
where the goal is to impute 25\% and 50\% of missing data points.
For (3) \textit{AD}, 
we experiment 5 AD tasks 
with a data ratio of 5\%
Table~\ref{tbl:few_total} present the results,
indicating that applying our method to UniTS yields the performance gain across all tasks, outperforming other single-task models \cite{wu2022timesnet,nie2022time,liu2023itransformer}.

\begin{figure*}[t]    
\centering
\begin{minipage}{0.47\textwidth}
\vspace{-12pt}
\centering
\begin{adjustbox}{max width=1.00\textwidth}
\begin{NiceTabular}{c|cccc|cc}
\toprule
\multirow{2.5}{*}{Dataset} & \multicolumn{2}{c}{UniTS}  & \multicolumn{2}{c}{+ CM} & \multicolumn{2}{c}{Imp.} \\
\cmidrule(lr){2-3} \cmidrule(lr){4-5} \cmidrule(lr){6-7}
& MSE & MAE & MSE & MAE & MSE & MAE \\
\midrule
Solar &  0.597 & 0.607 & \cellcolor{yellow!20} \textcolor{red}{\textbf{0.586}} & \cellcolor{yellow!20} \textcolor{red}{\textbf{0.585}} & \first{1.9\%} & \first{3.6\%} \\
River & \textcolor{red}{\textbf{1.374}} & 0.698 & \cellcolor{yellow!20}\textcolor{red}{\textbf{1.374}} & \cellcolor{yellow!20} \textcolor{red}{\textbf{0.686}}& {0.0\%} & \first{1.7\%} \\
Hospital & 1.067 & 0.797 & \cellcolor{yellow!20} \textcolor{red}{\textbf{1.020}} & \cellcolor{yellow!20} \textcolor{red}{\textbf{0.777}}& \first{4.4\%} & \first{2.5\%} \\
\midrule
Avg. & 1.013 & 0.701 & \cellcolor{yellow!20} \textcolor{red}{\textbf{0.993}} & \cellcolor{yellow!20} \textcolor{red}{\textbf{0.683}} & \first{2.0\%} & \first{2.6\%} \\
\bottomrule
\end{NiceTabular}
\end{adjustbox}
\captionsetup{type=table}
\caption{\textbf{Zero-shot.} Unseen datasets.}
\label{tbl:zero_data}
\end{minipage}
\hfill
\begin{minipage}{0.49\textwidth}
\centering
\begin{adjustbox}{max width=1.00\textwidth}
\begin{NiceTabular}{c|cccc|cc}
\toprule
\multirow{2.5}{*}{Dataset} & 
\multicolumn{2}{c}{UniTS} &
\multicolumn{2}{c}{+ CM} &
\multicolumn{2}{c}{Imp.} 
\\
\cmidrule(lr){2-3} \cmidrule(lr){4-5} \cmidrule(lr){6-7}
  & MSE & MAE & MSE & MAE & MSE & MAE\\
\midrule
ECL & 0.237 & 0.329 & \cellcolor{yellow!20} \textcolor{red}{\textbf{0.231}} & \cellcolor{yellow!20} \textcolor{red}{\textbf{0.323}} & \textcolor{red}{\textbf{2.5}\%} & \textcolor{red}{\textbf{1.8}\%} \\
ETTh1   & 0.495 & 0.463 & \cellcolor{yellow!20} \textcolor{red}{\textbf{0.492}} & \cellcolor{yellow!20} 0.463 & \textcolor{red}{\textbf{0.6}\%} & 0.0\% \\
Traffic  & 0.632 & 0.372 & \cellcolor{yellow!20} \textcolor{red}{\textbf{0.592}} & \cellcolor{yellow!20} \textcolor{red}{\textbf{0.369}} & \textcolor{red}{\textbf{6.3}\%} & \textcolor{red}{\textbf{0.8}\%} \\
Weather  & 0.335 & 0.336 & \cellcolor{yellow!20} 0.335 & \cellcolor{yellow!20} 0.336 & 0.0\% & 0.0\%  \\
\bottomrule
\end{NiceTabular}
\end{adjustbox}
\captionsetup{type=table}
\caption{\textbf{Zero-shot.} New forecasting horizons.}
\label{tbl:zero_horizon}
\vspace{15pt}
\end{minipage}
\begin{minipage}{1.000\textwidth}
        \centering
        \vspace{3pt}
        \begin{adjustbox}{max width=1.00\textwidth}
        \begin{NiceTabular}{c|c|ccccccccccccc|c}
        \toprule
         \multirow{2.5}{*}{\shortstack{CD by \textbf{model}\\($\mathbf{Q} \mathbf{K}^{\top}/\sqrt{d_k}$)}} & \multirow{2.5}{*}{\shortstack{CD by \textbf{dataset}\\($\mathbf{M}$, ours)}} & \multicolumn{13}{c}{Average MSE across four horizons} 
        &\multirow{2.5}{*}{Avg.}\\
         \cmidrule(lr){3-15}
          &   &  ETTh1 & ETTh2 & ETTm1 & ETTm2 & PEMS03 & PEMS04 & PEMS07 & PEMS08 & Exchange & Weather & Solar & ECL & Traffic & \\
        \midrule
        \cmark &  & \second{0.457} & \second{0.384} & \second{0.408} & \second{0.293} & \second{0.142} & 0.121 & 0.102  & 0.254 & \second{0.368} & 0.260 & 0.234 & 0.179 & \second{0.428} & 0.279\\
        \midrule
          & \cmark & 0.466 & \first{0.383} & \first{0.398} & \first{0.289} & 0.206 & \second{0.116} & \second{0.101} & \second{0.162} & \first{0.363} & \second{0.259} & \second{0.233} & \second{0.176} & 0.429 & \second{0.275}\\
        \midrule
        \cmark & \cmark &  \first{0.444} & \first{0.383} & \first{0.398} & \first{0.289} & \first{0.124} & \first{0.098} & \first{0.082} & \first{0.152} & \first{0.363} & \first{0.250} & \first{0.228} & \first{0.168} & \first{0.422} & \first{0.261}\\
        \bottomrule
        \end{NiceTabular}
        \end{adjustbox}
        \captionsetup{type=table}
        \caption{
        \textbf{Effectiveness of CD by model and CD by dataset}. 
        Using both 
        the attention matrix (i.e., CD by model)
        and
        CM (i.e., CD by dataset) to capture local and global dependencies yields the best results, with CM alone outperforming the attention matrix in some cases.
        }
        \vspace{5pt}
        \label{tbl:global_local}
    \end{minipage}
\end{figure*}

\textbf{c) Zero-shot learning.}
We perform FCST under two types of zero-shot settings:
1)~\textit{Zero-shot dataset}: We evaluate our model on an unseen dataset that was not included during training.
2)~\textit{Zero-shot task}: We assess the model's ability to predict a new forecasting horizon that was not encountered during training, by adding the mask tokens at the end of the TS to predict the desired future time steps.

For the FCST on unseen datasets, 
we evaluate our method using three datasets \cite{solar,mcleod2013optimal,hyndman2008forecasting}.
Table~\ref{tbl:zero_data} presents the results,
demonstrating consistent improvements by incorporating CMs.
For the FCST with new forecasting horizons, 
we predict additional 384 time steps (by adding 24 masked tokens of length 16 at the end of the TS) on top of the base forecasting horizon of 96.
Table~\ref{tbl:zero_horizon} presents the results
with four different datasets \cite{zhou2021informer,wu2021autoformer}, 
showing performance gains on three 
datasets.

\subsection{Ablation Studies}
\begin{table}[t]
\centering
\begin{adjustbox}{max width=1.00\columnwidth}
\begin{NiceTabular}{cc|c|cc|c}
\toprule
 \multicolumn{2}{c}{Components} & \multirow{2.5}{*}{$\mathbf{A}$} &\multicolumn{2}{c}{FCST (20)} & CLS (18)\\ 
\cmidrule(lr){1-2} \cmidrule(lr){4-5} \cmidrule(lr){6-6}
 Sim. & Dom. &  & MSE &MAE & Acc.\\
\midrule
  &   & $\mathbf{I}$  & 0.502 & 0.408 & 75.4\%\\
 &   & $\mathbf{1}$  & 0.478 & 0.393 & 75.1\%\\
 \cmark &  & $|\mathbf{R}|$ & 0.474 & 0.390 &  \textcolor{blue}{\underline{78.8\%}} \\
 \cmark & & $\mathbf{\bar{R}}$ &  \textcolor{blue}{\underline{0.471}} &  \textcolor{blue}{\underline{0.388}} & 78.1\%\\
 & \cmark
 & $\sigma\left(\alpha \cdot \mathbf{I}+\beta\right)$ & 0.497 & 0.406 & 76.2\%\\
 \cmark& \cmark& $\sigma\left(\alpha \cdot \mathbf{\bar{R}}+\beta\right)$ & \textcolor{red}{\textbf{0.452}} & \textcolor{red}{\textbf{0.384}} &  \textcolor{red}{\textbf{80.6\%}} \\
\bottomrule
\end{NiceTabular}
\end{adjustbox}
\caption{Ablation study of CM.}
\label{tbl:abl_CM}
\end{table}

\textbf{Components of CM.} 
To demonstrate the effectiveness of a CM, we conduct an ablation study of
using the similarity matrix (Sim.) and the domain parameters (Dom.). 
Table~\ref{tbl:abl_CM} presents the result
with 20 FCST and 18 CLS tasks with UniTS under the prompt-tuning setting,
indicating that incorporating both components yields the best results.
Note that, to isolate the effect of the domain parameters, we replace $\bar{\mathbf{R}}$ with an identity matrix ($\mathbf{I}$) in the fifth row of Table~\ref{tbl:abl_CM}.

\textbf{CD by model (attention matrix) \& CD by dataset (CM).}
To demonstrate the importance of capturing 
both 
CD by model 
and 
CD by dataset, 
we conduct an ablation study, as shown in Table~\ref{tbl:global_local}. 
Specifically, 
we use the attention weights $\mathbf{W}$ in $\operatorname{Attn}(\mathbf{Q}, \mathbf{K}, \mathbf{V})=\operatorname{Softmax}\left(\mathbf{W} \right)\cdot\mathbf{V}$ in the following manner: 
1) $\mathbf{Q} \mathbf{K}^{\top}/\sqrt{d_k}$ (i.e., attention matrix) for CD by model, 
2) $\mathbf{M}$ (i.e., channel mask) for CD by dataset, and 
3) $\mathbf{M} \odot \mathbf{Q} \mathbf{K}^{\top}/\sqrt{d_k}$ for both.
The results show the average MSE across four horizons, 
indicating that using both components yields the best results.
Notably, using only CMs 
provides better performance than 
using only attention matrices in some datasets.

\section{Analysis}
\begin{figure}[t]
\centering
\includegraphics[width=1.00\columnwidth]{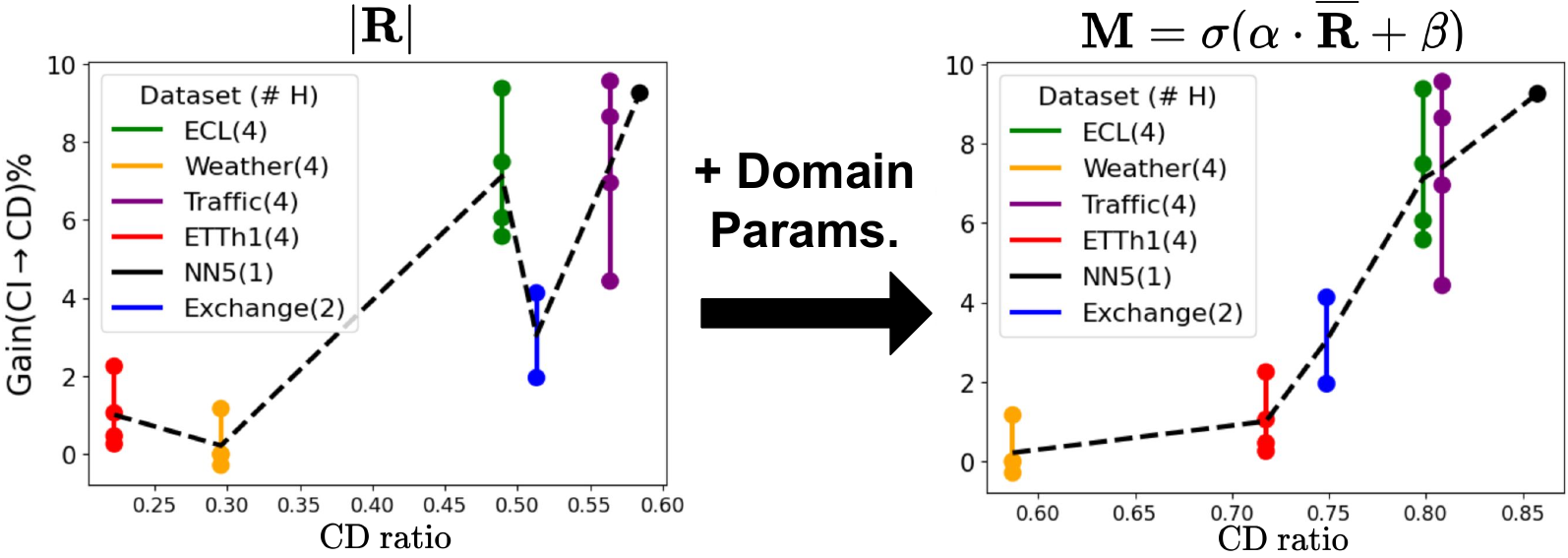} 
\caption{
CD ratio vs. Gain from (CI $\rightarrow$ CD).
}
\label{fig:CD_gain}
\end{figure}
\textbf{Effectiveness of domain parameters.}
To demonstrate the importance of domain parameters
in reflecting the degree of CD,
we compare
1) the CD ratio 
and 
2) the performance gain achieved with the CD framework against the CI framework with UniTS.
Figure~\ref{fig:CD_gain} 
shows that the gain is highly correlated with the CD ratio of a CM with the domain parameters ($r(\mathbf{M})$), but less so without them ($r(|\mathbf{R}|)$).

\begin{figure*}[t]
    \begin{minipage}{0.635\textwidth}
        \vspace{6pt}
        \centering
        \begin{adjustbox}{max width=1.00\textwidth}
        \begin{NiceTabular}{cc|ccccccc}
\toprule
$\alpha,\beta$ & CD ratio & \textcolor{blue}{Weather} & ILI & \textcolor{darkgreen}{ETTh1} & Exchange & ECL & Traffic & \textcolor{red}{NN5} \\
\cmidrule(lr){1-1} \cmidrule(lr){2-2} \cmidrule(lr){3-9}
\xmark & $r(|\mathbf{R}|)$  & 0.296 & 0.708 & 0.222 & 0.513 & 0.489 & 0.564 & 0.584 \\
\cmark & $r(\mathbf{M})$ & \textcolor{blue}{0.587} & 0,706 & \textcolor{darkgreen}{0.717} & 0.749 & 0.800 & 0.808 & \textcolor{red}{0.857} \\
\midrule
\multicolumn{2}{c}{\multirow{2}{*}{
}} & \multicolumn{7}{c}{Low  \quad $\xleftarrow{\hspace{2.7cm}}$ $r(\mathbf{M})$ $\xrightarrow{\hspace{2.7cm}}$ \quad High} \\
\multicolumn{2}{c}{ } & \multicolumn{7}{c}{Low  \quad $\xleftarrow{\hspace{1.16cm}}$ Dependencies btw channels $\xrightarrow{\hspace{1.16cm}}$ \quad High}
\\
\bottomrule
\end{NiceTabular}
        \end{adjustbox}
        \captionsetup{type=table}
        \caption{\textbf{CD ratio with and without domain parameters.} Applying domain parameters ($\alpha,\beta$) aligns the CD ratio of CM 
        ($r(\mathbf{M})$) 
        with the dependencies between channels, which is as also supported in Figure~\ref{fig:cdr_comparison}.}
        \label{tbl:cdr_comparison}
        \vspace{12pt}
    \end{minipage}
    \begin{minipage}{0.345\textwidth}
        \vspace{-8pt}
        \centering
        \includegraphics[width=1.000\textwidth]{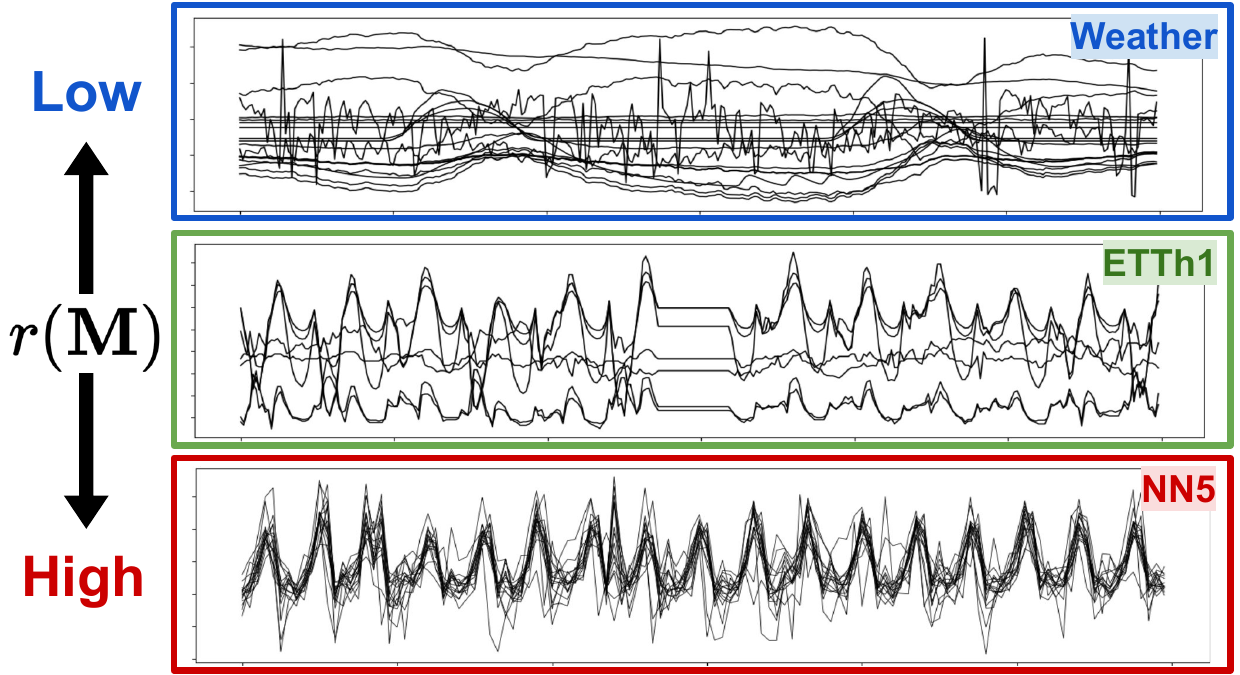}
        \caption{TS visualization by $r(\mathbf{M})$.}
        \label{fig:cdr_comparison}
    \end{minipage}
\begin{minipage}{1.00\textwidth}
        \centering
        \vspace{10pt}
        \includegraphics[width=1.000\textwidth]{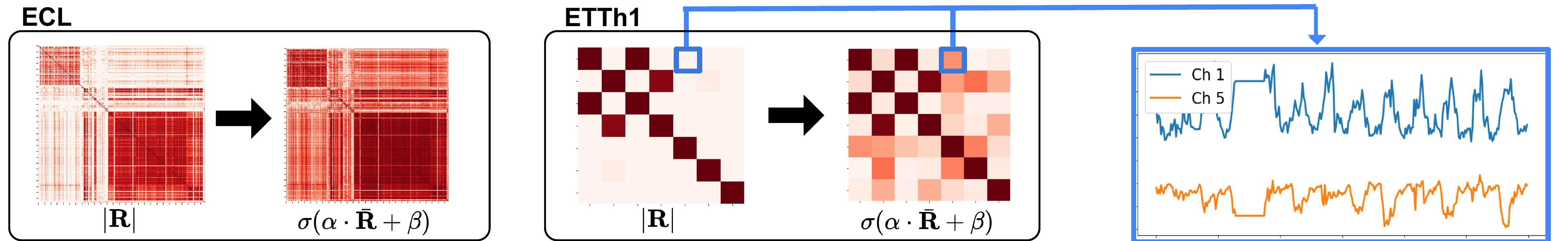}
        \caption{\textbf{
CM with and without domain parameters}. 
The figure shows CMs w/o domain parameters (e.g., similarity matrices) and CMs 
of two datasets, 
with each color scaled from 0 (light) to 1 (dark). 
The CM of ETTh1 reveals hidden relationships using domain parameters that are not captured by the correlation matrix.
}
        \label{fig:CM_viz}
    \end{minipage}
\end{figure*}

\textbf{CD ratio comparison.}
Table~\ref{tbl:cdr_comparison} 
presents the CD ratios 
of CMs with and without\footnote{For a CM w/o domain parameters, we use the absolute correlation matrix ($|\mathbf{R}|$) instead of its mean-scaled version ($\bar{\mathbf{R}}$) to ensure a fair comparison with $\mathbf{M}$, which is also scaled between 0 and 1.} domain parameters ($r(\mathbf{M})$ and $r(|\mathbf{R}|)$),
when using UniTS.
The results show that while datasets with higher $r(|\mathbf{R}|)$ generally have higher $r(\mathbf{M})$, this relationship is not consistent; for instance, Weather \cite{wu2021autoformer} exhibits lower CD despite having a stronger correlation compared to ETTh1 \cite{zhou2021informer}. 
Figure~\ref{fig:cdr_comparison} supports these findings by visualizing the channels of the datasets, revealing that the channels of ETTh1 tend to be more dependent on each other than those of Weather. 
These results underscore the importance of using domain parameters to adjust $|\mathbf{R}|$ for learning absolute dependencies specific to each dataset.
Furthermore, datasets with a larger number of channels 
tend to have higher $r(\mathbf{M})$, 
aligning with the prior work \cite{ahamed2024timemachine} emphasizing CD over CI for datasets with more channels.

\begin{table}[t]
\centering
\begin{adjustbox}{max width=1.00\columnwidth}
\begin{NiceTabular}{c|c|cc}
\toprule
Method & Dataset & MSE & MAE \\
\midrule
\multicolumn{2}{c}{UniTS} & 1.006 & 0.701 \\
\midrule
\multirow{3}{*}{+ CM} 
& Forecasting + Classification & \second{0.995} & \second{0.684} \\
& Forecasting & \first{0.993} & \first{0.683}\\
& Closest & \first{0.993} & \first{0.683}\\
\bottomrule
\end{NiceTabular}
\end{adjustbox}
\caption{$\alpha, \beta$ for unseen datasets.}
\label{tb:alpha_zeroshot}
\end{table}

\textbf{Domain parameters for unseen dataset.}
For an unseen dataset, selecting the appropriate domain parameters 
is challenging, as these 
are not learned during training. 
To address this issue, we propose three strategies: 
1) averaging the parameters across all datasets, 
2) averaging the parameters from the forecasting datasets, 
and 3) selecting parameters from the dataset with the closest $r(\bar{\mathbf{R}})$,
where Table~\ref{tb:alpha_zeroshot} demonstrates the robustness of these strategies.

\textbf{Visualization of CM.}
Figure~\ref{fig:CM_viz} shows the CMs of 
ECL and ETTh1,
illustrating the dependencies between the channels of each dataset.
The CM of ETTh1 reveals a hidden relationship between the first and fifth channels when using domain parameters, which is not identified 
by $|\mathbf{R}|$
alone.

\begin{table}[t]
\centering
\begin{adjustbox}{max width=1.00\columnwidth}
\begin{NiceTabular}{c|c|cccc}
\toprule
\multirow{2.5}{*}{Dataset} & \multirow{2.5}{*}{w/o CM} & \multicolumn{4}{c}{w/ CM} \\
\cmidrule{3-6}
& & Euc. & Cos. & DTW & Corr.\\
\midrule
ETTh1 & 0.457 & \second{0.445} & 0.446 & \first{0.444} & \first{0.444}\\
ETTh2 & \second{0.384} & \second{0.384} & \second{0.384} & 0.385 & \first{0.383}\\
ETTm1 & 0.408 & 0.402 &  0.403 & \second{0.401} & \first{0.398}\\
ETTm2 & 0.293 & 0.292 & \second{0.290} & 0.292 & \first{0.289}\\
PEMS03 & 0.142 & 0.146 & \second{0.134} & - & \first{0.124}\\
PEMS04 & 0.121 & 0.111 & \second{0.105} & - & \first{0.098}\\
PEMS07 & 0.102 & 0.092 & \second{0.087} & - & \first{0.082}\\
PEMS08 & 0.254 & \second{0.163} & 0.179 & - & \first{0.152}\\
Exchange & 0.368 & \second{0.364} & \first{0.363} & \second{0.364} & \first{0.363}\\
Weather & 0.260 & 0.256 & 0.255 & \second{0.254} & \first{0.250}\\
Solar & 0.234 & 0.232 & \second{0.229} & - & \first{0.228}\\
ECL & 0.179 & 0.173 & \second{0.171} & - & \first{0.168}\\
Traffic & \second{0.428} & 0.432 & 0.443 & - & \first{0.422}\\
\midrule
\rowcolor{yellow!20} Avg. & 0.279 & 0.269 & \second{0.268} & - & \first{0.261}\\
\bottomrule
\end{NiceTabular}
\end{adjustbox}
\caption{Various metrics for similarity.}
\label{tbl:metric_abl}
\end{table}

\textbf{Various TS metrics.}
To demonstrate the effectiveness of CMs using metrics beyond 
(Pearson) correlation, 
we apply CMs to iTransformer with three different metrics:
1) Euclidean distance (Euc.), which we min-max normalize to the range (0,1) and subtract from 1 to convert it into a similarity metric; 2) cosine similarity (Cos.), for which we take the absolute value, following the same intuition as correlation; and 3) dynamic time warping (DTW), where we apply the same process as with Euc.
Table~\ref{tbl:metric_abl} presents 
average MSE for 4$H$s,
indicating that CMs yield a performance gain regardless of the metric, with the best performance achieved with correlation.
Note that we use DTW only for datasets with 
$C < 100$
due to its computational complexity.

\begin{figure*}[t]
    \centering
    \begin{minipage}{0.425\textwidth}
        \vspace{-10pt}
        \centering
        \includegraphics[width=1.000\textwidth]{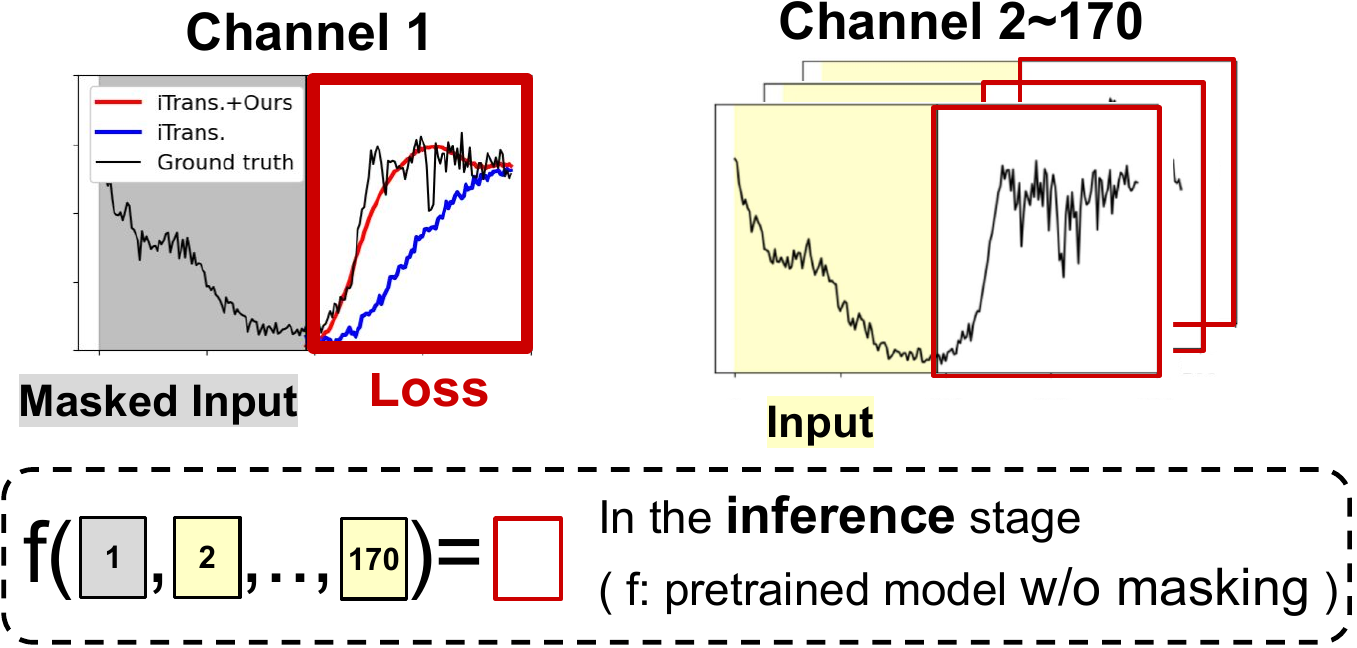}
        \caption{Masked channel prediction.}
        \label{fig:mcp}
    \end{minipage}
    \hfill
    \begin{minipage}{0.52\textwidth}
        \centering
        \begin{adjustbox}{max width=1.00\textwidth}
        \begin{NiceTabular}{c|cc|c|cc|c}
        \toprule
        \multirow{2.5}{*}{$H$} & \multicolumn{3}{c}{PEMS04 ($C=307$)} & \multicolumn{3}{c}{PEMS08 ($C=170$)}\\
        \cmidrule(lr){2-4} \cmidrule(lr){5-7} 
        & iTrans. & + CM &  Imp.
        & iTrans. & + CM &  Imp.\\
        \midrule
        12 & 
        0.549 & \cellcolor{yellow!20} \first{0.300} & \first{45.4\%} &
        0.628 & \cellcolor{yellow!20} \first{0.200} & \first{68.1\%}
        \\
        24 & 
        0.718 & \cellcolor{yellow!20} \first{0.351} & \first{51.1\%} &
        0.678 & \cellcolor{yellow!20} \first{0.241} & \first{64.5\%}
        \\
        48 & 
        0.750 & \cellcolor{yellow!20} \first{0.409} & \first{45.5\%} &
        1.197 & \cellcolor{yellow!20} \first{1.059} & \first{11.5\%}
        \\
        96 & 
        0.758 & \cellcolor{yellow!20} \first{0.513} & \first{32.3\%} &
        1.375 & \cellcolor{yellow!20} \first{1.217} & \first{11.5\%}
        \\
        \midrule
        Avg.  & 0.694 & \cellcolor{yellow!20} \first{0.393} & \first{43.3\%} &0.970 & \cellcolor{yellow!20} \first{0.679} & \first{29.9\%} \\
        \bottomrule
        \end{NiceTabular}
        \end{adjustbox}
        \captionsetup{type=table}
        \caption{Results of masked channel prediction.}
        \label{tbl:mcp}
        \vspace{15pt}
    \end{minipage}
    \begin{minipage}{0.59\textwidth}
        \centering
        \begin{adjustbox}{max width=1.00\textwidth}
        \begin{NiceTabular}{c|ccc|ccc}
        \toprule
        \multirow{2.5}{*}{$L,H=96$} & \multicolumn{3}{c}{Weather ($C=21$)} & \multicolumn{3}{c}{ECL ($C=321$)} \\
        \cmidrule(lr){2-4} \cmidrule(lr){5-7}
         & iTrans. & - & + CM & iTrans. & - & + CM \\
        \cmidrule{1-1} \cmidrule(lr){2-2} \cmidrule(lr){3-3} \cmidrule(lr){4-4} \cmidrule(lr){5-5} \cmidrule(lr){6-6} \cmidrule(lr){7-7} 
        CD by model (Att. matrix) & \cmark &  & \cmark & \cmark &  & \cmark \\
        CD by dataset (CM) &  & \cmark & \cmark & & \cmark  & \cmark \\
        \cmidrule{1-1} \cmidrule(lr){2-4} \cmidrule(lr){5-7}
        Train (sec/epoch)& 26.2 & 24.1 & \cellcolor{yellow!20} 26.7 & 33.2 & 26.0 & \cellcolor{yellow!20}  36.4 \\
        Inference (ms) & 11.1 & 11.1 & \cellcolor{yellow!20}  11.2 & 12.4 & 11.0 & \cellcolor{yellow!20}  13.2 \\
        \midrule
        Avg. MSE & 0.260 & \second{0.259} & \cellcolor{yellow!20} 
 \first{0.250} & 0.179 &  \second{0.176} & \cellcolor{yellow!20}  \first{0.168} \\
        \bottomrule
        \end{NiceTabular}
        \end{adjustbox}
        \captionsetup{type=table}
        \caption{Efficiency analysis.}
        \label{tbl:efficiency}
    \end{minipage}
    \hfill
    \begin{minipage}{0.365\textwidth}
        \centering
        \includegraphics[width=1.000\textwidth]{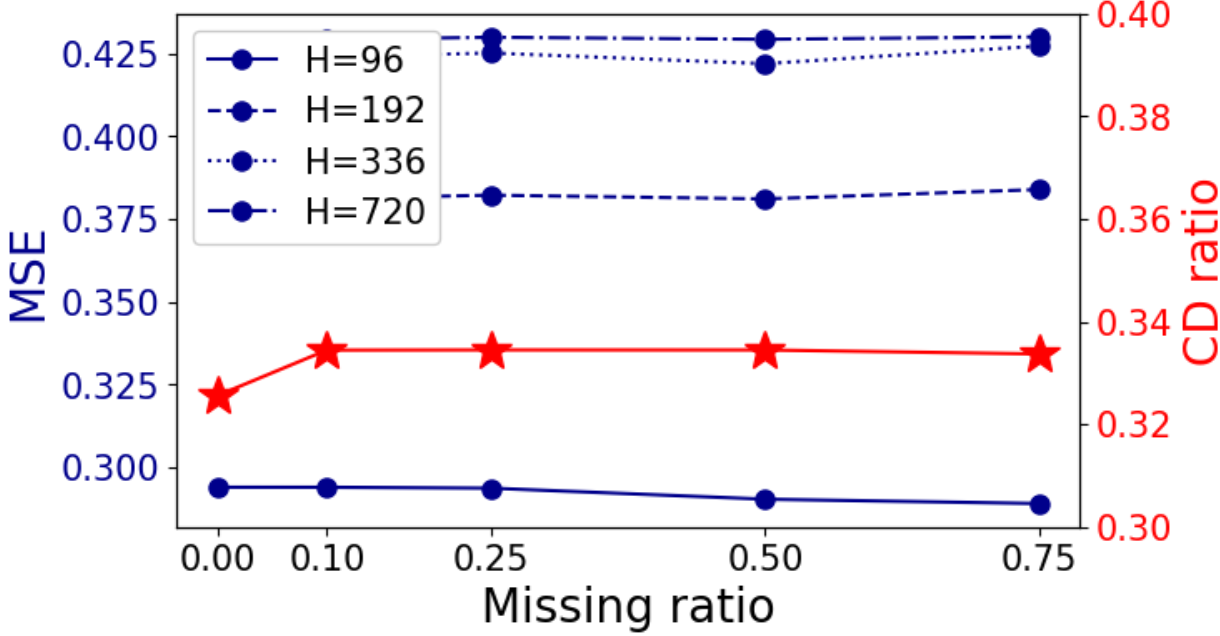}
        \caption{Robustness to missingness.}
        \label{fig:ts_missing}
    \end{minipage}
\end{figure*}

\textbf{Masked channel prediction.}
To evaluate the model's ability to capture CD, we introduce a novel evaluation method, \textit{masked channel prediction}, which involves predicting the future values of the masked channel 
using 
the historical values of the 
unmasked channels. 
Specifically, we calculate the average loss for each channel when masked once, with the loss for the $c$-th channel expressed as:
\begin{equation}
L_{(c)}(y,\hat{y}) = \text{MSE}(y[:,c],\hat{y}_{(c)}[:,c]), 
 \text{where } \hat{y}_{(c)}=f(x_{(c)}),
\end{equation}
where $x_{(c)}$ is $x$ with the $c$-th channel masked, 
and $\hat{y}_{(c)}$ is the predicted output using $x_{(c)}$ as the input.
Note that 
masked channel prediction is an \textit{evaluation method} that does not require additional training,
and instead uses a model pretrained without any masking.

To assess the effectiveness of CMs in capturing CD,
we experiment masked channel prediction
with iTransformer with and without CMs, 
imputing the historical values of the masked channels with
there average values, which are essentially zero with normalization.
The results in Table~\ref{tbl:mcp}
demonstrate significant improvements by incorporating CMs.
Furthermore, Figure~\ref{fig:mcp} visualizes the predicted results for PEMS08 \cite{liu2022scinet},
where models with CMs predict masked channels better than models without CMs.
We provide more results 
in Appendix~\ref{sec:appendix_mcp}.

\textbf{Efficiency analysis.}
To demonstrate the efficiency of CMs, 
we compare the training and inference times of iTransformer on two datasets \cite{wu2021autoformer} 
with varying numbers of channels, 
using 
1) only attention matrices, 
2) only CMs, and
3) both.
Training time is measured per epoch and inference time is measured per data instance. 
Table~\ref{tbl:efficiency} indicates that incorporating CMs \textit{does not significantly impact computational time}, 
even with datasets containing a large number of channels. 
It is important to note that similarity matrices can be precomputed offline, making CMs practical for use.

\textbf{Robustness to missingness.}
To demonstrate the robustness of CMs to missing values, we analyze scenarios where some values are randomly missing at ratios of 10\%, 25\%, 50\% and 75\%,
with the missing values linearly interpolated using adjacent values.
Figure~\ref{fig:ts_missing} shows the result on ETTh2 \cite{zhou2021informer} with iTransformer, indicating that both $r(|\mathbf{R}|)$
and the performance remain robust
to the missingness. 

\begin{table}[t]
\centering
\begin{adjustbox}{max width=1.00\columnwidth}
\begin{NiceTabular}{c|c|c|c}
\toprule
\multicolumn{2}{c}{Domain parameters} & Channel mask ($\mathbf{M}$) & Asym. \\ 
\midrule
Scalar & $\textcolor{red}{\alpha}, \textcolor{red}{\beta} \in \mathbb{R}^1$ & $\sigma\left(\textcolor{red}{\alpha} \cdot \mathbf{\bar{R}}+\textcolor{red}{\beta}\right)$ & \xmark \\
\midrule
\multirow{2}{*}{Vector} & $\textcolor{darkgreen}{\mathbf{E}} \in \mathbb{R}^d$ & $\operatorname{Norm}(\textcolor{darkgreen}{\mathbf{E}} \textcolor{darkgreen}{\mathbf{E}^T}) \odot \bar{\mathbf{R}}$ & \xmark \\
 & $\textcolor{darkgreen}{\mathbf{E}_1},\textcolor{darkgreen}{\mathbf{E}_2} \in \mathbb{R}^d$   & $\operatorname{Norm}(\textcolor{darkgreen}{\mathbf{E}_1} \textcolor{darkgreen}{\mathbf{E}_2^T}) \odot \bar{\mathbf{R}}$ & \cmark \\
\midrule
Matrix & $\textcolor{blue}{\mathbf{A}} \in \mathbb{R}^{C\times C}$   & $\textcolor{blue}{\mathbf{A}} \odot \bar{\mathbf{R}}$ & \cmark \\
\bottomrule
\end{NiceTabular}
\end{adjustbox}
\caption{Extension of domain parameters.}
\label{tbl:alter_domain}
\vspace{-8pt}
\end{table}

\begin{figure*}[t]
    \centering
    \begin{minipage}{1.000\textwidth}
        \centering
        \begin{adjustbox}{max width=1.00\textwidth}
        \begin{NiceTabular}{c|ccccccccccccc|c}
        \toprule
        \multirow{2.5}{*}{}
        & \multicolumn{13}{c}{Average MSE across four horizons} & \multirow{2.5}{*}{Avg.}
        \\ 
         \cmidrule(lr){2-14}
         &  ETTh1 & ETTh2 & ETTm1 & ETTm2 &  PEMS03 & PEMS04 & PEMS07 & PEMS08 &  Exchange & Weather & Solar & ECL & Traffic &  \\
        \midrule
        $\alpha, \beta$ & \first{0.444} & \first{0.383} & \first{0.398} & \first{0.289} & \first{0.124} & \first{0.098} & \first{0.082} & \first{0.152} & \first{0.363} & \first{0.250} & \second{0.228} & \first{0.168} & {0.422} & \first{0.261}\\
         $\mathbf{E}$ & \second{0.452}  & {0.391} & \second{0.402}  & \second{0.291} & 0.150  & 0.106 & 0.096  & 0.202 & \second{0.364}  & \second{0.255} & 0.234  & \second{0.177} & \second{0.416}  & 0.272  \\
          $\mathbf{E_1},\mathbf{E_2}$ & \second{0.452}  & {0.391} & \second{0.402}  & \second{0.291} & 0.152  & 0.105 & \second{0.095}  & 0.205 & \second{0.364}  & \second{0.255} & 0.233  & \second{0.177} & \first{0.415}  & 0.272  \\
        $\mathbf{A}$ & 0.454  & {0.391} & \second{0.402}  & \second{0.291} & \second{0.138}  & \second{0.099} & 0.102  & \second{0.182} & \second{0.364}  & 0.259 & \first{0.226}  & \second{0.177} & {0.418}  & \second{0.269}  \\
        \midrule
        - & 0.457  & \second{0.384} & 0.408  & 0.293 & 0.142 & 0.121 & 0.102  & 0.254 & 0.368  & 0.260 & 0.234  & 0.179 & 0.428  & 0.279 \\
        \bottomrule
        \end{NiceTabular}
        \end{adjustbox}
        \captionsetup{type=table}
        \caption{
        \textbf{Various domain parameters.} Using \textit{scalar} parameters ($\alpha,\beta$),
        which scale/shift the similarity matrix,
        yields the best results, compared to \textit{vector} parameters ($\mathbf{E}$ or $\mathbf{E}_1,\mathbf{E}_2$) and \textit{matrix} parameters ($\mathbf{A}$).
        }
        \label{tb:extension_domain}
    \end{minipage}
\end{figure*}

\textbf{Various options for domain parameters.}
The proposed domain parameters ($\alpha$ and $\beta$) 
are scalars that adjust $\bar{\mathbf{R}}$ 
by changing its elements 
monotonically.
For further flexibility, 
we design alternative options for the parameters:
1) a vector $\mathbf{E}$ for each channel
and 2) a matrix $\mathbf{A}$ for each dataset.
Both options are used to construct an adjustment matrix 
that is element-wise multiplied to
$\bar{\mathbf{R}}$, as shown in Table~\ref{tbl:alter_domain}.

The first option serves as identifiable vectors for each channel, 
with the adjustment matrix constructed based on the inner product between these vectors 
and normalized with
$\operatorname{Norm}(\cdot) = \operatorname{Softmax}\left(\operatorname{ReLU}\left( \cdot \right)\right)$, 
while the second option 
acts as the adjustment matrix itself.
For the vector parameters, 
we also implement an asymmetric matrix version that requires two different vectors for each channel: one for the inner vector ($\mathbf{E}_1$) and the other for the outer vector ($\mathbf{E}_2$), as described in the previous work \cite{wu2019graph}.
Table~\ref{tb:extension_domain} shows that using scalar parameters achieves the best performance, demonstrating the efficiency of CMs by requiring only two additional parameters per dataset.

\section{Discussion}
\textbf{A. Scope of our work.}
This paper focuses on enhancing Transformer-based methods capturing 
CD, \textit{rather than improving TSFMs}.
We argue that \textit{this focus is broadly impactful} for two reasons:
\begin{itemize}[leftmargin=0.3cm,itemsep=0pt,topsep=-2pt, partopsep=0pt]
\item \textbf{a) Transformer for CD.} Since the introduction of iTransformer \cite{liu2023itransformer}, various methods have adopted complex attention mechanisms to capture CD, while simpler algorithms are used to capture TD; in this paper, we examine six representative methods \cite{liu2023itransformer,liang2024minusformer,wang2024card,yu2024prformer,gao2024units,dong2024timesiam} that follow this framework.
\item \textbf{b) Large-scale TS datasets.} 
The emergence of large-scale TS datasets further highlights the potential of our work, which emphasizes the importance of CD,
as the capacity--robustness trade-off \cite{han2023capacity} indicates that large datasets tend to favor CD-based models,
whereas smaller datasets benefit more from CI approaches.
\end{itemize}

Although we demonstrate applicability within UniTS \cite{gao2024units}, we leave the application to various TSFMs as future work due to its broader scope.

\begin{table}[t]
\centering
\begin{adjustbox}{max width=1.00\columnwidth}
\begin{NiceTabular}{c|c|c}
\toprule
\multirow{2.5}{*}{\shortstack{Univariate \\ 
(e.g., flattening)
}} & \multicolumn{2}{c}{Multivariate}\\
\cmidrule{2-3}
 & CI & CD \\
\midrule
\cite{liutimer}, \cite{rasul2023lag}, \cite{talukder2024totem}, \cite{woo2024unified} & 
\cite{ansari2024chronos}, \cite{das2024decoder}, \cite{goswami2024moment} & 
\cite{ekambaram2024tiny}, \cite{feng2024only}, \cite{gao2024units}, \cite{yuan2024unist} \\
\bottomrule
\end{NiceTabular}
\end{adjustbox}
\caption{Categorization of TSFMs.}
\label{tb:cat_TSFM}
\vspace{-8pt}
\end{table}

\textbf{B. Application to other TSFMs.}
It is important to clarify that our focus is \textit{not on improving TSFM}, but on enhancing general TS models that \textit{capture CD with Transformer}.
As shown in Table~\ref{tb:cat_TSFM}, various recent methods rely on attention-based approaches specifically to model CD. 

Among these methods, we select UniTS \cite{gao2024units} as the sole TSFM baseline in this study for the following reasons:
1) GTT \cite{feng2024only} 
reformulates TS forecasting as an image task for next curve shape prediction
and does not provide hyperparameter details or scripts for reproducibility.
2) UniST \cite{yuan2024unist} is tailored for urban spatio-temporal prediction, taking spatial inputs in a 4D grid format
which makes it incompatible with general TS benchmarks.
3) TTM \cite{ekambaram2024tiny}
is built on TSMixer \cite{chen2023tsmixer}, which falls outside the scope of our work.
In contrast, UniTS \cite{gao2024units} 
handles four diverse tasks with a single architecture
through prompt-tuning,
making
it a representative choice for demonstrating the 
effectiveness
of CM.

Furthermore, Moirai \cite{woo2024unified}, a CI method that applies time-axis attention after channel flattening, can still incorporate our CM by reshaping the CM from $C \times C$ to $C \cdot P \times C \cdot P$ with duplicated intra-channel values, where $P$ is the number of patches, to match the shape of Moirai’s attention matrix. Nonetheless, we exclude Moirai for the reasons detailed in Appendix \ref{sec:why_not_moirai}.

\vspace{4pt}
\textbf{C. Effectiveness of CM under non-stationary TS.}
\begin{itemize}[leftmargin=0.3cm,itemsep=-5pt,topsep=-2.8pt, partopsep=0pt]
\item \textbf{a) Distribution shifts.}
Adjusting the CM based on distribution shifts \textit{would conflict with its intended purpose}, potentially diminishing its ability to capture stable, dataset-specific characteristics. Therefore, we rely on the attention matrix to capture these changes. 
Furthermore, the effectiveness of CM is evident under such shifts, as our datasets exhibit significant distribution shifts \cite{han2023capacity}.
\item \textbf{b) Time-lagged dependencies.} Since channels may exhibit time-lagged dependencies \cite{zhao2024rethinking}, we include metrics such as DTW in Table \ref{tbl:metric_abl}, 
which explicitly account for time lags between channels. Importantly, the potential challenge posed by lagged relationships pertains to the \textit{choice of similarity metric}, not to the \textit{use of similarity itself}, which is our primary focus. Nonetheless, as shown in Table \ref{tbl:metric_abl}, our method demonstrates robust performance regardless of the metric.
\end{itemize}

\vspace{4pt}
\textbf{D. Comparison with other plug-in methods.}
The proposed CM is a \textit{plug-in} method that can be applied orthogonally to other plug-in methods, including
(Granger-causality based) LIFT \cite{chen2024similarity} and PRReg \cite{han2023capacity}.
Therefore, the key point is that \textit{applying our method to various backbones consistently yields performance improvements},
rather than comparing with other 
plug-ins. 
While a comparison between these methods is still possible, their baseline models differ from ours, as our method employs more recent Transformer-based architectures, making a direct comparison infeasible. 
Furthermore, their code is not implemented as a plug-in method but is rather tightly coupled with backbones.

\section{Conclusion}
In this work, we introduce 
PCD, 
where CD captured by a Transformer-based model is adjusted using a 
CM, 
which is a plug-in method that leverages dataset-specific information to capture CD. 
Experimental results demonstrate 
that utilizing dataset-specific information is crucial for TS modeling, 
enhancing performance across various backbones.
As our method is limited to Transformer-based models, we aim to develop a method to achieve PCD that \textit{does not rely on the 
architecture}.
We hope this work 
emphasizes 
the importance of utilizing dataset-specific information across diverse domains.

\section{Acknowledgements}
This work was supported by the National Research Foundation of Korea (NRF) grant funded by the Korea government (MSIT) (2022R1A4A1033384, RS-2023-00217705, RS-2024-00341749, RS-2026-25487501), the MSIT (Ministry of Science and ICT), Korea, under
the ICAN (ICT Challenge and Advanced Network of HRD) support program (RS-2023-00259934), Developing the Next-Generation General AI with Reliability, Ethics, and Adaptability (RS-2025-02283048),
and the Ministry of Trade, Industry and Resources (MOTIR), Korea, under the project Industrial Technology Infrastructure Program (RS-2024-00466693).


\bibliographystyle{IEEEbib}
\bibliography{icassp2026_conference}

@article{godahewa2021monash,
  title={Monash time series forecasting archive},
  author={Godahewa, Rakshitha and Bergmeir, Christoph and Webb, Geoffrey I and Hyndman, Rob J and Montero-Manso, Pablo},
  journal={arXiv:2105.06643},
  year={2021}
}

@inproceedings{abdulaal2021practical,
  title={Practical approach to asynchronous multivariate time series anomaly detection and localization},
  author={Abdulaal, Ahmed and Liu, Zhuanghua and Lancewicki, Tomer},
  booktitle={Proceedings of the 27th ACM SIGKDD conference on knowledge discovery \& data mining},
  pages={2485--2494},
  year={2021}
}

@inproceedings{mathur2016swat,
  title={SWaT: A water treatment testbed for research and training on ICS security},
  author={Mathur, Aditya P and Tippenhauer, Nils Ole},
  booktitle={2016 international workshop on cyber-physical systems for smart water networks (CySWater)},
  pages={31--36},
  year={2016},
  organization={IEEE}
}

@inproceedings{hundman2018detecting,
  title={Detecting spacecraft anomalies using lstms and nonparametric dynamic thresholding},
  author={Hundman, Kyle and Constantinou, Valentino and Laporte, Christopher and Colwell, Ian and Soderstrom, Tom},
  booktitle={Proceedings of the 24th ACM SIGKDD international conference on knowledge discovery \& data mining},
  pages={387--395},
  year={2018}
}

@inproceedings{su2019robust,
  title={Robust anomaly detection for multivariate time series through stochastic recurrent neural network},
  author={Su, Ya and Zhao, Youjian and Niu, Chenhao and Liu, Rong and Sun, Wei and Pei, Dan},
  booktitle={Proceedings of the 25th ACM SIGKDD international conference on knowledge discovery \& data mining},
  pages={2828--2837},
  year={2019}
}

@article{middlehurst2024bake,
  title={Bake off redux: a review and experimental evaluation of recent time series classification algorithms},
  author={Middlehurst, Matthew and Sch{\"a}fer, Patrick and Bagnall, Anthony},
  journal={Data Mining and Knowledge Discovery},
  pages={1--74},
  year={2024},
  publisher={Springer}
}

@article{taieb2012review,
  title={A review and comparison of strategies for multi-step ahead time series forecasting based on the NN5 forecasting competition},
  author={Taieb, Souhaib Ben and Bontempi, Gianluca and Atiya, Amir F and Sorjamaa, Antti},
  journal={Expert systems with applications},
  volume={39},
  number={8},
  pages={7067--7083},
  year={2012},
  publisher={Elsevier}
}

@article{rasul2023lag,
  title={Lag-llama: Towards foundation models for probabilistic time series forecasting},
  author={Rasul, Kashif and Ashok, Arjun and Williams, Andrew Robert and Ghonia, Hena and Bhagwatkar, Rishika and Khorasani, Arian and Bayazi, Mohammad Javad Darvishi and Adamopoulos, George and Riachi, Roland and Hassen, Nadhir and others},
  journal={arXiv:2310.08278},
  year={2023}
}

@book{hyndman2008forecasting,
  title={Forecasting with exponential smoothing: the state space approach},
  author={Hyndman, Rob and Koehler, Anne B and Ord, J Keith and Snyder, Ralph D},
  year={2008},
  publisher={Springer Science \& Business Media}
}

@article{mcleod2013optimal,
  title={Optimal deseasonalization for monthly and daily geophysical time series},
  author={McLeod, AI and Gweon, Hyukjun},
  journal={Journal of Environmental statistics},
  volume={4},
  number={11},
  pages={1--11},
  year={2013}
}

@inproceedings{wang2024timemixer++,
  title={Timemixer++: A general time series pattern machine for universal predictive analysis},
  author={Wang, Shiyu and Li, Jiawei and Shi, Xiaoming and Ye, Zhou and Mo, Baichuan and Lin, Wenze and Ju, Shengtong and Chu, Zhixuan and Jin, Ming},
  booktitle={ICLR},
  year={2025}
}

@inproceedings{wu2022timesnet,
  title={Timesnet: Temporal 2d-variation modeling for general time series analysis},
  author={Wu, Haixu and Hu, Tengge and Liu, Yong and Zhou, Hang and Wang, Jianmin and Long, Mingsheng},
  booktitle={ICLR},
  year={2023}
}

@inproceedings{dong2024timesiam,
  title={TimeSiam: A Pre-Training Framework for Siamese Time-Series Modeling},
  author={Dong, Jiaxiang and Wu, Haixu and Wang, Yuxuan and Qiu, Yunzhong and Zhang, Li and Wang, Jianmin and Long, Mingsheng},
  booktitle={ICML},
  year={2024}
}

@inproceedings{zhang2023crossformer,
  title={Crossformer: Transformer utilizing cross-dimension dependency for multivariate time series forecasting},
  author={Zhang, Yunhao and Yan, Junchi},
  booktitle={ICLR},
  year={2023}
}

@inproceedings{zhao2024rethinking,
  title={Rethinking Channel Dependence for Multivariate Time Series Forecasting: Learning from Leading Indicators},
  author={Zhao, Lifan and Shen, Yanyan},
  booktitle={ICLR},
  year={2024}
}

@article{qi2024enhancing,
  title={Enhancing Multivariate Time Series Forecasting with Mutual Information-driven Cross-Variable and Temporal Modeling},
  author={Qi, Shiyi and Wen, Liangjian and Li, Yiduo and Yang, Yuanhang and Li, Zhe and Rao, Zhongwen and Pan, Lujia and Xu, Zenglin},
  journal={arXiv:2403.00869},
  year={2024}
}

@article{huang2023crossgnn,
  title={Crossgnn: Confronting noisy multivariate time series via cross interaction refinement},
  author={Huang, Qihe and Shen, Lei and Zhang, Ruixin and Ding, Shouhong and Wang, Binwu and Zhou, Zhengyang and Wang, Yang},
  journal={NeurIPS},
  year={2023}
}

@article{li2023revisiting,
  title={Revisiting long-term time series forecasting: An investigation on linear mapping},
  author={Li, Zhe and Qi, Shiyi and Li, Yiduo and Xu, Zenglin},
  journal={arXiv:2305.10721},
  year={2023}
}

@article{yang2024vcformer,
  title={VCformer: Variable Correlation Transformer with Inherent Lagged Correlation for Multivariate Time Series Forecasting},
  author={Yang, Yingnan and Zhu, Qingling and Chen, Jianyong},
  journal={arXiv:2405.11470},
  year={2024}
}

@inproceedings{jin2023time,
  title={Time-llm: Time series forecasting by reprogramming large language models},
  author={Jin, Ming and Wang, Shiyu and Ma, Lintao and Chu, Zhixuan and Zhang, James Y and Shi, Xiaoming and Chen, Pin-Yu and Liang, Yuxuan and Li, Yuan-Fang and Pan, Shirui and others},
  year={2024},
  booktitle={ICLR}
}

@inproceedings{zhou2023one,
  title={One fits all: Power general time series analysis by pretrained lm},
  author={Zhou, Tian and Niu, Peisong and Sun, Liang and Jin, Rong and others},
  year={2023},
  booktitle={NeurIPS}
}

@inproceedings{kinga2015method,
  title={A method for stochastic optimization},
  author={Kinga, D and Adam, Jimmy Ba and others},
  booktitle={ICLR},
  volume={5},
  pages={6},
  year={2015},
  organization={San Diego, California;}
}

@inproceedings{lai2018modeling,
  title={Modeling long-and short-term temporal patterns with deep neural networks},
  author={Lai, Guokun and Chang, Wei-Cheng and Yang, Yiming and Liu, Hanxiao},
  booktitle={The 41st international ACM SIGIR conference on research \& development in information retrieval},
  year={2018}
}

@inproceedings{liu2022scinet,
  title={Scinet: Time series modeling and forecasting with sample convolution and interaction},
  author={Liu, Minhao and Zeng, Ailing and Chen, Muxi and Xu, Zhijian and Lai, Qiuxia and Ma, Lingna and Xu, Qiang},
  booktitle={NeurIPS},
  year={2022}
}

@inproceedings{gao2024units,
  title={Units: Building a unified time series model},
  author={Gao, Shanghua and Koker, Teddy and Queen, Owen and Hartvigsen, Thomas and Tsiligkaridis, Theodoros and Zitnik, Marinka},
  year={2024},
  booktitle={NeurIPS}
}

@inproceedings{liutimer,
  title={Timer: Generative Pre-trained Transformers Are Large Time Series Models},
  author={Liu, Yong and Zhang, Haoran and Li, Chenyu and Huang, Xiangdong and Wang, Jianmin and Long, Mingsheng},
  year={2024},
  booktitle={ICML}
}

@inproceedings{woo2024unified,
  title={Unified training of universal time series forecasting transformers},
  author={Woo, Gerald and Liu, Chenghao and Kumar, Akshat and Xiong, Caiming and Savarese, Silvio and Sahoo, Doyen},
  year={2024},
  booktitle={ICML}
}

@article{zhu2021long,
  title={Long-short transformer: Efficient transformers for language and vision},
  author={Zhu, Chen and Ping, Wei and Xiao, Chaowei and Shoeybi, Mohammad and Goldstein, Tom and Anandkumar, Anima and Catanzaro, Bryan},
  journal={NeurIPS},
  year={2021}
}

@article{yu2024prformer,
  title={PRformer: Pyramidal Recurrent Transformer for Multivariate Time Series Forecasting},
  author={Yu, Yongbo and Yu, Weizhong and Nie, Feiping and Li, Xuelong},
  journal={arXiv:2408.10483},
  year={2024}
}

@article{liang2024minusformer,
  title={Minusformer: Improving Time Series Forecasting by Progressively Learning Residuals},
  author={Liang, Daojun and Zhang, Haixia and Yuan, Dongfeng and Zhang, Bingzheng and Zhang, Minggao},
  journal={arXiv:2402.02332},
  year={2024}
}

@inproceedings{goswami2024moment,
  title={Moment: A family of open time-series foundation models},
  author={Goswami, Mononito and Szafer, Konrad and Choudhry, Arjun and Cai, Yifu and Li, Shuo and Dubrawski, Artur},
  year={2024},
  booktitle={ICML}
}

@article{talukder2024totem,
  title={TOTEM: TOkenized Time Series EMbeddings for General Time Series Analysis},
  author={Talukder, Sabera and Yue, Yisong and Gkioxari, Georgia},
  journal={arXiv:2402.16412},
  year={2024}
}

@inproceedings{wang2024card,
  title={CARD: Channel aligned robust blend transformer for time series forecasting},
  author={Wang, Xue and Zhou, Tian and Wen, Qingsong and Gao, Jinyang and Ding, Bolin and Jin, Rong},
  booktitle={ICLR},
  year={2024}
}

@book{wei2019multivariate,
  title={Multivariate time series analysis and applications},
  author={Wei, William WS},
  year={2019},
  publisher={John Wiley \& Sons}
}

@article{han2023capacity,
  title={The Capacity and Robustness Trade-off: Revisiting the Channel Independent Strategy for Multivariate Time Series Forecasting},
  author={Han, Lu and Ye, Han-Jia and Zhan, De-Chuan},
  journal={arXiv:2304.05206},
  year={2023}
}

@article{ahamed2024timemachine,
  title={Timemachine: A time series is worth 4 mambas for long-term forecasting},
  author={Ahamed, Md Atik and Cheng, Qiang},
  journal={arXiv:2403.09898},
  year={2024}
}

@inproceedings{nie2022time,
  title={A time series is worth 64 words: Long-term forecasting with transformers},
  author={Nie, Yushan and Nguyen, Nam H and Sinthong, Pattarawat and Kalagnanam, Jayant},
  booktitle={ICLR},
  year={2023}
}

@inproceedings{liu2023itransformer,
  title={itransformer: Inverted transformers are effective for time series forecasting},
  author={Liu, Yong and Hu, Tengge and Zhang, Haoran and Wu, Haixu and Wang, Shiyu and Ma, Lintao and Long, Mingsheng},
  booktitle={ICLR},
  year={2024}
}

@inproceedings{lee2023learning,
  title={Learning to embed time series patches independently},
  author={Lee, Seunghan and Park, Taeyoung and Lee, Kibok},
  booktitle={ICLR},
  year={2024}
}

@article{chen2024similarity,
  title={From similarity to superiority: Channel clustering for time series forecasting},
  author={Chen, Jialin and Lenssen, Jan Eric and Feng, Aosong and Hu, Weihua and Fey, Matthias and Tassiulas, Leandros and Leskovec, Jure and Ying, Rex},
  journal={Advances in Neural Information Processing Systems},
  volume={37},
  pages={130635--130663},
  year={2024}
}

@inproceedings{zeng2023transformers,
  title={Are transformers effective for time series forecasting?},
  author={Zeng, Ailing and Chen, Muxi and Zhang, Lei and Xu, Qiang},
  booktitle={AAAI},
  year={2023}
}

@article{chen2023tsmixer,
  title={Tsmixer: An all-mlp architecture for time series forecasting},
  author={Chen, Si-An and Li, Chun-Liang and Yoder, Nate and Arik, Sercan O and Pfister, Tomas},
  journal={TMLR},
  year={2023}
}

@inproceedings{vaswani2017attention,
  title={Attention is all you need},
  author={Vaswani, Ashish and Shazeer, Noam and Parmar, Niki and Uszkoreit, Jakob and Jones, Llion and Gomez, Aidan N and Kaiser, {\L}ukasz and Polosukhin, Illia},
  booktitle={NeurIPS},
  year={2017}
}

@inproceedings{wu2021autoformer,
  title={Autoformer: Decomposition transformers with auto-correlation for long-term series forecasting},
  author={Wu, Haixu and Xu, Jiehui and Wang, Jianmin and Long, Mingsheng},
  booktitle={NeurIPS},
  year={2021}
}

@inproceedings{wu2019graph,
  title={Graph wavenet for deep spatial-temporal graph modeling},
  author={Wu, Zonghan and Pan, Shirui and Long, Guodong and Jiang, Jing and Zhang, Chengqi},
  booktitle={IJCAI},
  year={2019}
}

@misc{solar,
  author = {NREL},
  title = {Solar power data for integration studies},
  howpublished = {\url{https://www.nrel.gov/grid/solar-power-data.html}},
  year = {2006}
}

@inproceedings{zhou2021informer,
  title={Informer: Beyond efficient transformer for long sequence time-series forecasting},
  author={Zhou, Haoyi and Zhang, Shanghang and Peng, Jieqi and Zhang, Shuai and Li, Jianxin and Xiong, Hui and Zhang, Wancai},
  booktitle={AAAI},
  year={2021}
}

@article{feng2024only,
  title={General Time Transformer: an Encoder-only Foundation Model
for Zero-Shot Multivariate Time Series Forecasting},
  author={Feng, Cheng and Huang, Long and Krompass, Denis},
  journal={CIKM},
  year={2024}
}

@inproceedings{yuan2024unist,
  title={Unist: A prompt-empowered universal model for urban spatio-temporal prediction},
  author={Yuan, Yuan and Ding, Jingtao and Feng, Jie and Jin, Depeng and Li, Yong},
  booktitle={Proceedings of the 30th ACM SIGKDD Conference on Knowledge Discovery and Data Mining},
  pages={4095--4106},
  year={2024}
}

@article{dau2019ucr,
  title={The UCR time series archive},
  author={Dau, Hoang Anh and Bagnall, Anthony and Kamgar, Kaveh and Yeh, Chin-Chia Michael and Zhu, Yan and Gharghabi, Shaghayegh and Ratanamahatana, Chotirat Ann and Keogh, Eamonn},
  journal={IEEE/CAA Journal of Automatica Sinica},
  volume={6},
  number={6},
  pages={1293--1305},
  year={2019},
  publisher={IEEE}
}

@article{bagnall2018uea,
  title={The UEA multivariate time series classification archive, 2018},
  author={Bagnall, Anthony and Dau, Hoang Anh and Lines, Jason and Flynn, Michael and Large, James and Bostrom, Aaron and Southam, Paul and Keogh, Eamonn},
  journal={arXiv:1811.00075},
  year={2018}
}

@article{ansari2024chronos,
  title={Chronos: Learning the language of time series},
  author={Ansari, Abdul Fatir and Stella, Lorenzo and Turkmen, Caner and Zhang, Xiyuan and Mercado, Pedro and Shen, Huibin and Shchur, Oleksandr and Rangapuram, Syama Sundar and Arango, Sebastian Pineda and Kapoor, Shubham and others},
  journal={arXiv:2403.07815},
  year={2024}
}

@inproceedings{das2024decoder,
  title={A decoder-only foundation model for time-series forecasting},
  author={Das, Abhimanyu and Kong, Weihao and Sen, Rajat and Zhou, Yichen},
  booktitle={ICML},
  year={2024}
}

@article{ekambaram2024tiny,
  title={Tiny time mixers (ttms): Fast pre-trained models for enhanced zero/few-shot forecasting of multivariate time series},
  author={Ekambaram, Vijay and Jati, Arindam and Dayama, Pankaj and Mukherjee, Sumanta and Nguyen, Nam and Gifford, Wesley M and Reddy, Chandra and Kalagnanam, Jayant},
  journal={NeurIPS},
  volume={37},
  pages={74147--74181},
  year={2024}
}

\newpage
\appendix
\onecolumn
\startcontents[appendices]
\printcontents[appendices]{}{1}{\section*{Appendix}\setcounter{tocdepth}{2}}

\numberwithin{table}{section}
\numberwithin{figure}{section}
\numberwithin{equation}{section}

\newpage
\section{Dataset Description}
\label{sec:data}
\subsection{Dataset for Single-task Models}
For TS forecasting in a single-task setting, we evaluate the effectiveness of our proposed method using 13 datasets, with their statistics described in Table~\ref{tab:single_task_data_FCST}.
We adhere to the same data processing and train-validation-test split protocol as iTransformer~\cite{liu2023itransformer}, ensuring that the training, validation, and test sets are separated in chronological order. The input length is consistently set to 96 across all datasets.
Note that $N$ and $C$ denote the size of the dataset and number of channels in a dataset, respectively.

\begin{table*}[h]
\centering
\vspace{20pt}
\begin{adjustbox}{max width=1.00\textwidth}
\begin{NiceTabular}{l|c|c|c}
\toprule
Dataset & $C$  & Prediction Length &$(N_\text{train},N_\text{val},N_\text{test})$ \\
\midrule
ETTh1 \cite{zhou2021informer} & 7 & {\{96, 192, 336, 720\}} & (8545, 2881, 2881) \\
\midrule
ETTh2 \cite{zhou2021informer} & 7 & {\{96, 192, 336, 720\}} & (8545, 2881, 2881) \\
 \midrule
 ETTm1 \cite{zhou2021informer} & 7 & {\{96, 192, 336, 720\}} & (34465, 11521, 11521) \\
 \midrule
 ETTm2 \cite{zhou2021informer} & 7 & {\{96, 192, 336, 720\}} & (34465, 11521, 11521) \\
\midrule
Exchange \cite{wu2021autoformer} & 8 & {\{96, 192, 336, 720\}} & (5120, 665, 1422)\\
\midrule
Weather \cite{wu2021autoformer} & 21 & {\{96, 192, 336, 720\}} & (36792, 5271, 10540) \\
\midrule
ECL \cite{wu2021autoformer}& 321 & {\{96, 192, 336, 720\}} & (18317, 2633, 5261) \\
\midrule
Traffic \cite{wu2021autoformer} & 862 & {\{96, 192, 336, 720\}} & (12185, 1757, 3509) \\
\midrule
Solar-Energy \cite{lai2018modeling} & 137  & {\{96, 192, 336, 720\}} & (36601, 5161, 10417) 
\\
\midrule
PEMS03 \cite{liu2022scinet} & 358 & {\{12, 24, 48, 96\}} & (15617, 5135, 5135) \\
\midrule
PEMS04 \cite{liu2022scinet}& 307 & {\{12, 24, 48, 96\}} & (10172, 3375, 3375) \\
\midrule
PEMS07 \cite{liu2022scinet}& 883 & {\{12, 24, 48, 96\}} & (16911, 5622, 5622) \\
\midrule
PEMS08 \cite{liu2022scinet} & 170 & {\{12, 24, 48, 96\}} & (10690, 3548, 3548) \\
\bottomrule
\end{NiceTabular}
\end{adjustbox}
\caption{Single-task forecasting datasets.}
\label{tab:single_task_data_FCST}
\end{table*}

\clearpage
\subsection{Dataset for Time Series Foundation Models}
The datasets used in the experiment are aggregated from the Monash Forecasting Repository~\cite{godahewa2021monash}, the Time Series Classification Website~\cite{middlehurst2024bake}, and the Time Series Library~\cite{wu2022timesnet}. The combined training set includes more than 35 million time steps and over 6,000 variables (channels). Note that $N$, $L$, $C$ denote the training size, input length, and number of channels in a dataset, respectively.

\subsubsection{Multi-task Learning}
For TS forecasting and classification in a multi-task setting, we evaluate the effectiveness of our proposed method using 20 datasets for forecasting and 18 datasets for classification. The statistics of these datasets are summarized in Table~\ref{tab:multi_task_data_FCST} and \ref{tab:multi_task_data_CLS}, respectively.

\begin{table*}[h]
\centering
\vspace{10pt}
\begin{adjustbox}{max width=0.75\textwidth}
\begin{NiceTabular}{c|l|c|ccc}

\toprule
Category & Dataset & Prediction Length &$N$ & $L$ & $C$ \\
\midrule
\multirow{3}{*}{Finance} & NN5 \cite{taieb2012review} & 112 & 409 & 112 & 111\\
\cmidrule{2-6}
&\multirow{2}{*}{Exchange \cite{wu2021autoformer}}& 192  & 5024 & \multirow{2}{*}{96} & \multirow{2}{*}{8} \\
 & & 336 & 4880 &  & \\
\midrule
\multirow{8}{*}{Electricity} & \multirow{4}{*}{ECL \cite{wu2021autoformer}} & 96 & 18221 & \multirow{4}{*}{96} & \multirow{4}{*}{321} \\
& & 192  & 18125 &  &  \\
& & 336 & 17981 &  &  \\
& & 720 & 17597 &  & \\
\cmidrule{2-6}
&\multirow{4}{*}{ETTh1 \cite{zhou2021informer}} & 96  & 8449 & \multirow{4}{*}{96} & \multirow{4}{*}{7} \\
& & 192 & 8353 &  &  \\
& & 336 & 8209 &  &  \\
& & 720 & 7825 &  &  \\
\midrule
Illness & ILI \cite{wu2021autoformer} & 60 & 581 & 36 & 7 \\
\midrule
\multirow{4}{*}{Traffic} & \multirow{4}{*}{Traffic \cite{wu2021autoformer}}& 96 & 12089 & \multirow{4}{*}{96} & \multirow{4}{*}{862} \\
 & & 192  & 11993 &  & \\
 & & 336& 11849 &  &  \\
 & & 720& 11465 &  &  \\
\midrule
\multirow{4}{*}{Weather} & \multirow{4}{*}{Weather \cite{wu2021autoformer}} & 96 & 36696 & \multirow{4}{*}{96} & \multirow{4}{*}{21} \\
& & 192   & 36600 &  &  \\
& & 336 & 36456 &  &  \\
& & 720 & 36072 &  &  \\
\bottomrule
\end{NiceTabular}
\end{adjustbox}
\caption{Multi-task forecasting datasets.}
\label{tab:multi_task_data_FCST}
\end{table*}

\clearpage
\begin{table*}[h]
\centering
\vspace{20pt}
\begin{adjustbox}{max width=0.75\textwidth}
\begin{NiceTabular}{c|l|c|ccc}
\toprule
Category & Dataset & \# classes &$N$ & $L$ & $C$ \\
\midrule
Finance & SharePriceIncrease \cite{dau2019ucr}& 2 & 965 & 60 & 1 \\
\midrule
\multirow{3}{*}{Audio} & JapaneseVowels \cite{bagnall2018uea} & 9 & 270 & 29 & 12\\
& SpokenArabicDigits \cite{bagnall2018uea} & 10 & 6599 & 93 & 13\\
& Heartbeat \cite{bagnall2018uea}& 2 & 204 & 405 & 61 \\
\midrule
\multirow{2}{*}{ECG} &  ECG5000 \cite{dau2019ucr} & 5 & 500 & 140 & 1 \\
& NonInvasiveFetalECGThorax1 \cite{dau2019ucr} & 52 & 1800 & 750 & 1 \\
\midrule
\multirow{3}{*}{EEG} &Blink \cite{bagnall2018uea} & 2 & 500 & 510 & 4 \\
& FaceDetection \cite{bagnall2018uea} & 2 & 5890 & 62 & 144 \\
& SelfRegulationSCP2 \cite{bagnall2018uea} & 2 & 200 & 1152 & 7 \\
\midrule
\multirow{3}{*}{Sensors} &ElectricDevices \cite{dau2019ucr} & 7 & 8926 & 96 & 1 \\
& Trace \cite{dau2019ucr} & 4 & 100 & 275 & 1 \\
& FordB \cite{dau2019ucr} & 2 & 3636 & 500 & 1 \\
\midrule
\multirow{3}{*}{Human Activity} & MotionSenseHAR \cite{bagnall2018uea} & 6 & 966 & 200 & 12 \\
& EMOPain \cite{bagnall2018uea} & 3 & 968 & 180 & 30 \\
& UWaveGestureLibrary \cite{bagnall2018uea} & 8 & 120 & 315 & 3 \\
\midrule
\multirow{3}{*}{Traffic} & Chinatown \cite{dau2019ucr} & 2 & 20 & 24 & 1 \\
& MelbournePedestrian \cite{dau2019ucr} & 10 & 1194 & 24 & 1 \\
& PEMS-SF \cite{bagnall2018uea}& 7  & 267 & 144 & 963 \\
\bottomrule
\end{NiceTabular}
\end{adjustbox}
\caption{Multi-task classification datasets.}
\label{tab:multi_task_data_CLS}
\end{table*}

\clearpage
\subsubsection{Few-shot Learning}
For TS forecasting, classification, imputation, and anomaly detection in a few-shot setting, we evaluate the effectiveness of our proposed method using nine datasets for forecasting, six datasets for classification, four datasets for imputation, and five datasets for anomaly detection. The statistics of these datasets related to forecasting and classification are summarized in Table \ref{tab:fewshot_data_FCST}, Table \ref{tab:fewshot_data_CLS}, \ref{tab:fewshot_data_imp}, and \ref{tab:fewshot_data_ad}, respectively.

\begin{table*}[h]
\centering
\vspace{20pt}
\begin{adjustbox}{max width=0.95\textwidth}
\begin{NiceTabular}{c|l|c|ccc}
\toprule
Category & Dataset & Prediction Length & $N$ & $L$ & $C$ \\
\midrule
\multirow{8}{*}{Electricity} & \multirow{4}{*}{ETTh2 \cite{zhou2021informer}} & 96 & 8449 & \multirow{4}{*}{96} & \multirow{4}{*}{7} \\
& & 192 & 8353 & & \\
& & 336 & 8209 & & \\
& & 720 & 7825 & & \\
\cmidrule{2-6}
& \multirow{4}{*}{ETTm1 \cite{zhou2021informer}} & 96 & 34369 & \multirow{4}{*}{96} & \multirow{4}{*}{7} \\
& & 192 & 34273 & & \\
& & 336 & 34129 & & \\
& & 720 & 33745 & & \\
\midrule
Weather & SaugeenRiverFlow \cite{mcleod2013optimal} & 24 & 18921 & 48 & 1 \\
\bottomrule
\end{NiceTabular}
\end{adjustbox}
\caption{Few-shot forecasting datasets.}
\label{tab:fewshot_data_FCST}
\end{table*}

\begin{table*}[h]
\centering
\vspace{20pt}
\begin{adjustbox}{max width=0.95\textwidth}
\begin{NiceTabular}{c|l|c|ccc}
\toprule
Category & Dataset & \# classes & $N$ & $L$ & $C$ \\
\midrule    
ECG & ECG200 \cite{dau2019ucr} & 2 & 100 & 96 & 1 \\
\midrule
EEG & SelfRegulationSCP1 \cite{bagnall2018uea} & 2 & 268 & 896 & 6 \\
\midrule
\multirow{3}{*}{\shortstack{\\\\Human\\Activity}}  & RacketSports \cite{bagnall2018uea} & 4& 151 & 30 & 6 \\
& Handwriting \cite{bagnall2018uea} & 26 & 150 & 152 & 3 \\
& Epilepsy \cite{bagnall2018uea} & 4& 137 & 207 & 3 \\
\midrule
Sensor & StarLightCurves \cite{dau2019ucr} & 3& 1000 & 1024 & 1 \\
\bottomrule
\end{NiceTabular}
\end{adjustbox}
\caption{Few-shot classification datasets.}
\label{tab:fewshot_data_CLS}
\end{table*}

\vspace{20pt}
\begin{figure*}[h]
    \centering
    \begin{minipage}{0.49\textwidth}
        \centering
        \begin{adjustbox}{max width=1.00\textwidth}
            \begin{NiceTabular}{c|l|cc}
            \toprule
            Category & Dataset & $L$ & $C$ \\ 
            \midrule
            \multirow{3}{*}{Electricity} & ETTm1 \cite{zhou2021informer} & 96 & 7 \\
            &ETTh1 \cite{zhou2021informer} & 96 & 7 \\
            &ECL \cite{wu2021autoformer} & 96 & 321 \\
            \midrule
            Weather & Weather \cite{wu2021autoformer} & 96 & 21\\
            \bottomrule
            \end{NiceTabular}
        \end{adjustbox}
        \captionsetup{type=table}
        \caption{Few-shot imputation datasets.}
        \label{tab:fewshot_data_imp}
    \end{minipage}
    \hfill
    \begin{minipage}{0.49\textwidth}
        \centering
        \begin{adjustbox}{max width=1.00\textwidth}
            \begin{NiceTabular}{c|l|cc}
            \toprule
            Category & Dataset & $L$ & $C$  \\
            \midrule
            \multirow{2}{*}{Machine} & SMD~\cite{su2019robust}  & 96 & 38 \\
            & PSM~\cite{abdulaal2021practical}  & 96 & 25 \\
            \midrule
            \multirow{2}{*}{Spacecraft} &  MSL~\cite{hundman2018detecting}  & 96 & 55 \\
            & SMAP~\cite{hundman2018detecting}  & 96 & 25 \\
            \midrule
            Infrastructure  & SWaT~\cite{mathur2016swat}  & 96 & 51 \\
            \bottomrule
            \end{NiceTabular}
        \end{adjustbox}
        \captionsetup{type=table}
        \caption{Few-shot anomaly detection datasets.}
        \label{tab:fewshot_data_ad}
    \end{minipage}
\end{figure*}

\clearpage

\subsubsection{Zero-shot Learning} 
For TS forecasting in a zero-shot setting, we evaluate the effectiveness of our proposed method using six datasets. Three of these datasets are used for the zero-shot setting with unseen datasets, while the remaining four datasets are used for the zero-shot setting with new prediction lengths. The statistics for the three unseen datasets are summarized in Table~\ref{tab:zeroshot_data_FCST}.

\begin{table*}[h]
\centering
\vspace{15pt}
\begin{adjustbox}{max width=0.95\textwidth}
\begin{NiceTabular}{c|l|c|cc}
\toprule
Category & Dataset & Prediction Length & $L$ & $C$  \\
\midrule
Electricity & Solar \cite{solar} & 64 & 128 & 137 \\
Weather & SaugeenRiverFlow \cite{mcleod2013optimal} & 128 & 256 & 1 \\
Healthcare & Hospital \cite{hyndman2008forecasting}& 16 & 32 & 767  \\
\bottomrule
\end{NiceTabular}
\end{adjustbox}
\caption{Zero-shot forecasting datasets.}
\label{tab:zeroshot_data_FCST}
\end{table*}

\clearpage
\section{Implementation Details}
\label{sec:impl}
It is important to note that we follow the experimental settings of iTransformer for single-task and UniTS for multi-task settings, respectively. 
For the implementation, we use the official code repositories of both methods, running the provided scripts without modifications. However, for UniTS in the prompt tuning setting, we encountered an issue where the model failed to converge using the provided script. This was resolved by setting the hidden dimension to $D=32$, which we applied uniformly across both UniTS and its integration with our method.
The following sections outline the specific settings we adhered to.

\vspace{20pt}
\subsection{Implementation for Single-task Models}
\textbf{iTransformer.} 
Following iTransformer~\cite{liu2023itransformer}, we use the Adam optimizer~\cite{kinga2015method} and L2 loss for model optimization. The batch size is consistently set to 32, and the number of training epochs is fixed at 10. Since our approach is plug-and-play, we do not adjust any hyperparameters for our method; instead, we use the same hyperparameters employed by iTransformer.
For all other single-task models, including CARD \cite{wang2024card}, Minusformer \cite{liang2024minusformer}, and PRformer \cite{yu2024prformer}, we follow the experimental setups provided in their original works.

\vspace{20pt}
\subsection{Implementation for Time Series Foundation Models}
\textbf{Model architecture.} In a multi-task setting, the UniTS network consists of three UniTS blocks, along with one \texttt{GEN} tower and one \texttt{CLS} tower. 
For each data source, specific prompt and task tokens are assigned, with forecasting tasks on the same source but with varying forecast lengths using the same prompt and \texttt{GEN} token. 
To enable zero-shot learning on new datasets, a shared prompt and \texttt{GEN} token are applied across all data sources. 
The embedding dimensions are set to 64 for the supervised version, and 32 for the prompt-tuning version, and all blocks in UniTS retain the same feature shape.

\textbf{Model training.} 
In multi-task settings, models are trained jointly on multiple tasks following a unified protocol. 
To match the largest dataset, samples from each dataset are repeated within each epoch. 
Supervised training is conducted over 5 epochs with gradient accumulation, yielding an effective batch size of 1024. 
The initial learning rate is set at 3.2e-2 and is adjusted using a multi-step decay schedule. 
For self-supervised pretraining, the models training with an are trained for 10 epochs with effective batch size of 4096, starting with a learning rate of 6.4e-3, which is adjusted using a cosine decay schedule.

\vspace{20pt}
\subsection{Construction of Correlation Matrix}
For constructing the correlation matrix for CM, we used the datasets corresponding to the training period for forecasting datasets and the training instances for classification datasets. Specifically, for a forecasting dataset with shape ($C, L_{\text{train}} + L_{\text{val}} + L_{\text{test}}$), we compute the correlation matrix with shape ($C, C$) using only the training period with shape ($C, L_{\text{train}}$). For a classification dataset with shape ($N_{\text{train}} + N_{\text{val}} + N_{\text{test}}, C, L$), we compute the correlation matrix with shape ($C, C$) using only the training instances with shape ($N_{\text{train}}, C, L$) by averaging across the instances.

\clearpage
\section{Application to Single-task Models}
\label{sec:itransformer_full}
To demonstrate the effectiveness of our method on a model with a single-task setting, we apply it to the TS forecasting task using four backbones on 13 datasets.

\begin{figure*}[h]
    \vspace{5pt}
    \centering
    \begin{minipage}{0.44\textwidth}
        \centering
        \begin{adjustbox}{max width=1.00\textwidth}
        \begin{NiceTabular}{c|c|cc|cc}
        \toprule
        \multicolumn{2}{c}{\multirow{2.5}{*}{Metric}} & 
        \multicolumn{2}{c}{iTransformer} &
        \multicolumn{2}{c}{+ CM} 
        \\
        \cmidrule(lr){3-4} \cmidrule(lr){5-6}
        \multicolumn{2}{c}{ }  & {MSE} & {MAE}  & {MSE} & {MAE}  \\
        \toprule
        \multirow{5.5}{*}{ETTh1}
        &  96 &  {0.387} & {0.405} & \rowcolor{gray!20} \textcolor{red}{\textbf{0.385}} & \textcolor{red}{\textbf{0.404}}\\
        &  192 & {0.441} & {0.436} & \rowcolor{gray!20} \textcolor{red}{\textbf{0.438}} & \textcolor{red}{\textbf{0.434}}\\
        &  336 & {0.491} & {0.462} & \rowcolor{gray!20} \textcolor{red}{\textbf{0.475}} & \textcolor{red}{\textbf{0.454}}\\
        &  720 & {0.509} & {0.494} & \rowcolor{gray!20} \textcolor{red}{\textbf{0.477}} & \textcolor{red}{\textbf{0.474}}\\
        \cmidrule{2-6}
        &  Avg.& {0.457} & {0.449} & \rowcolor{gray!20} \textcolor{red}{\textbf{0.444}} & \textcolor{red}{\textbf{0.441}}\\
        \midrule
        \multirow{5.5}{*}{ETTh2}
        &  96 &  {0.301} & {0.350} & \rowcolor{gray!20} \textcolor{red}{\textbf{0.295}} & \textcolor{red}{\textbf{0.347}}\\
        &  192 &  {0.381} & {0.399} & \rowcolor{gray!20} \textcolor{red}{\textbf{0.380}} & \textcolor{red}{\textbf{0.397}}\\
        &  336 &  \textcolor{red}{\textbf{0.423}} & \textcolor{red}{\textbf{0.432}} & \rowcolor{gray!20} 0.427 & 0.434\\
        &  720 &  \textcolor{red}{\textbf{0.430}} & {0.446} & \rowcolor{gray!20} 0.432 & \textcolor{red}{\textbf{0.445}}\\
        \cmidrule{2-6}
        &  Avg.&  {0.384} & {0.407} & \rowcolor{gray!20} \textcolor{red}{\textbf{0.383}} & \textcolor{red}{\textbf{0.406}}\\
        \midrule
        \multirow{5.5}{*}{ETTm1}
        &  96 &  {0.342} & {0.377} & \rowcolor{gray!20} \textcolor{red}{\textbf{0.331}} & \textcolor{red}{\textbf{0.369}}\\
        &  192 & {0.383} & {0.396} & \rowcolor{gray!20} \textcolor{red}{\textbf{0.372}} & \textcolor{red}{\textbf{0.390}}\\
        &  336 & {0.418} & {0.418} & \rowcolor{gray!20} \textcolor{red}{\textbf{0.412}} & \textcolor{red}{\textbf{0.414}}\\
        &  720 & {0.487} & {0.456} & \rowcolor{gray!20} \textcolor{red}{\textbf{0.479}} & \textcolor{red}{\textbf{0.453}}\\
        \cmidrule{2-6}
        &  Avg.& {0.408} & {0.412} & \rowcolor{gray!20} \textcolor{red}{\textbf{0.398}} & \textcolor{red}{\textbf{0.406}}\\
        \midrule
        \multirow{5.5}{*}{ETTm2}
        &  96 &  {0.186} & \textcolor{red}{\textbf{0.272}} & \rowcolor{gray!20} \textcolor{red}{\textbf{0.184}} & \textcolor{red}{\textbf{0.272}}\\
        &  192 & {0.254} & {0.314} & \rowcolor{gray!20} \textcolor{red}{\textbf{0.251}} & \textcolor{red}{\textbf{0.311}}\\
        &  336 & {0.317} & {0.353} & \rowcolor{gray!20} \textcolor{red}{\textbf{0.312}} & \textcolor{red}{\textbf{0.350}}\\
        &  720 & {0.416} & {0.409} & \rowcolor{gray!20} \textcolor{red}{\textbf{0.412}} & \textcolor{red}{\textbf{0.408}}\\
        \cmidrule{2-6}
        &  Avg.& {0.293} & {0.337} & \rowcolor{gray!20} \textcolor{red}{\textbf{0.289}} & \textcolor{red}{\textbf{0.335}}\\
        \midrule
        \multirow{5.5}{*}{Exchange}
        &  96 &  {0.086} & {0.206} & \rowcolor{gray!20} \textcolor{red}{\textbf{0.085}} & \textcolor{red}{\textbf{0.205}}\\
        &  192 & {0.181} & {0.303} & \rowcolor{gray!20} \textcolor{red}{\textbf{0.180}} & \textcolor{red}{\textbf{0.302}}\\
        &  336 & {0.338} & {0.422} & \rowcolor{gray!20} \textcolor{red}{\textbf{0.337}} & \textcolor{red}{\textbf{0.421}}\\
        &  720 & {0.869} & {0.704} & \rowcolor{gray!20} \textcolor{red}{\textbf{0.850}} & \textcolor{red}{\textbf{0.696}}\\
        \cmidrule{2-6}
        &  Avg.& {0.368} & {0.409} & \rowcolor{gray!20} \textcolor{red}{\textbf{0.363}} & \textcolor{red}{\textbf{0.406}}\\
        \midrule
        \multirow{5.5}{*}{Weather}
        &  96 &  {0.174} & {0.215} & \rowcolor{gray!20} \textcolor{red}{\textbf{0.165}} & \textcolor{red}{\textbf{0.209}}\\
        &  192 & {0.224} & {0.258} & \rowcolor{gray!20} \textcolor{red}{\textbf{0.213}} & \textcolor{red}{\textbf{0.251}}\\
        &  336 & {0.281} & {0.298} & \rowcolor{gray!20} \textcolor{red}{\textbf{0.274}} & \textcolor{red}{\textbf{0.296}}\\
        &  720 & {0.359} & {0.351} & \rowcolor{gray!20} \textcolor{red}{\textbf{0.350}} & \textcolor{red}{\textbf{0.346}}\\
        \cmidrule{2-6}
        &  Avg.&  {0.260} & {0.281} & \rowcolor{gray!20} \textcolor{red}{\textbf{0.250}} & \textcolor{red}{\textbf{0.275}}\\
        \midrule
        \multirow{5.5}{*}{Solar}
        &  96 &  {0.201} & {0.234} & \rowcolor{gray!20} \textcolor{red}{\textbf{0.197}} & \textcolor{red}{\textbf{0.231}}\\
        &  192 & {0.238} & {0.263} & \rowcolor{gray!20} \textcolor{red}{\textbf{0.232}} & \textcolor{red}{\textbf{0.260}}\\
        &  336 & {0.248} & {0.273} & \rowcolor{gray!20} \textcolor{red}{\textbf{0.241}} & \textcolor{red}{\textbf{0.270}}\\
        &  720 & {0.249} & {0.275} & \rowcolor{gray!20} \textcolor{red}{\textbf{0.241}} & \textcolor{red}{\textbf{0.273}}\\
        \cmidrule{2-6}
        &  Avg.& {0.234} & {0.261} & \rowcolor{gray!20} \textcolor{red}{\textbf{0.228}} & \textcolor{red}{\textbf{0.258}}\\
        \bottomrule
        \end{NiceTabular}
        \end{adjustbox}
        \captionsetup{type=table}
        \label{tbl:itrans_full1}
    \end{minipage}
    \hfill
    \begin{minipage}{0.44\textwidth}
        \centering
        \begin{adjustbox}{max width=1.00\textwidth}
        \begin{NiceTabular}{c|c|cc|cc}
\toprule
\multicolumn{2}{c}{\multirow{2.5}{*}{Metric}} & 
\multicolumn{2}{c}{iTransformer} &
\multicolumn{2}{c}{+ CM} 
\\
\cmidrule(lr){3-4} \cmidrule(lr){5-6}
\multicolumn{2}{c}{ }  & {MSE} & {MAE}  & {MSE} & {MAE}  \\
\toprule
\multirow{5.5}{*}{PEMS03}
&  12 &  {0.071} & {0.174} & \rowcolor{gray!20} \textcolor{red}{\textbf{0.063}} & \textcolor{red}{\textbf{0.168}}\\
&  24 & {0.097} & {0.208} & \rowcolor{gray!20} \textcolor{red}{\textbf{0.087}} & \textcolor{red}{\textbf{0.197}}\\
&  48 & {0.161} & {0.272} & \rowcolor{gray!20} \textcolor{red}{\textbf{0.133}} & \textcolor{red}{\textbf{0.250}}\\
&  96 & {0.240} & {0.338} & \rowcolor{gray!20} \textcolor{red}{\textbf{0.212}} & \textcolor{red}{\textbf{0.316}}\\
\cmidrule{2-6}
&  Avg.& {0.142} & {0.248} & \rowcolor{gray!20} \textcolor{red}{\textbf{0.124}} & \textcolor{red}{\textbf{0.231}}\\
\midrule
\multirow{5.5}{*}{PEMS04}
&  12 &  {0.081} & {0.188} & \rowcolor{gray!20} \textcolor{red}{\textbf{0.075}} & \textcolor{red}{\textbf{0.181}}\\
&  24 &  {0.099} & {0.211} & \rowcolor{gray!20} \textcolor{red}{\textbf{0.086}} & \textcolor{red}{\textbf{0.196}}\\
&  48 &  {0.133} & {0.246} & \rowcolor{gray!20} \textcolor{red}{\textbf{0.108}} & \textcolor{red}{\textbf{0.222}}\\
&  96 &  {0.172} & {0.283} & \rowcolor{gray!20} \textcolor{red}{\textbf{0.125}} & \textcolor{red}{\textbf{0.242}}\\
\cmidrule{2-6}
&  Avg.&  {0.121} & {0.232} & \rowcolor{gray!20} \textcolor{red}{\textbf{0.098}} & \textcolor{red}{\textbf{0.210}}\\
\midrule
\multirow{5.5}{*}{PEMS07}
&  12 &  {0.067} & {0.165} & \rowcolor{gray!20} \textcolor{red}{\textbf{0.061}} & \textcolor{red}{\textbf{0.157}}\\
&  24 & {0.088} & {0.190} & \rowcolor{gray!20} \textcolor{red}{\textbf{0.076}} & \textcolor{red}{\textbf{0.179}}\\
&  48 & {0.113} & {0.218} & \rowcolor{gray!20} \textcolor{red}{\textbf{0.086}} & \textcolor{red}{\textbf{0.188}}\\
&  96 & {0.140} & {0.246} & \rowcolor{gray!20} \textcolor{red}{\textbf{0.104}} & \textcolor{red}{\textbf{0.208}}\\
\cmidrule{2-6}
&  Avg.& {0.102} & {0.205} & \rowcolor{gray!20} \textcolor{red}{\textbf{0.082}} & \textcolor{red}{\textbf{0.183}}\\
\midrule
\multirow{5.5}{*}{PEMS08}
&  12 &  {0.088} & {0.193} & \rowcolor{gray!20} \textcolor{red}{\textbf{0.085}} & \textcolor{red}{\textbf{0.190}}\\
&  24 & {0.138} & {0.243} & \rowcolor{gray!20} \textcolor{red}{\textbf{0.126}} & \textcolor{red}{\textbf{0.234}}\\
&  48 & {0.334} & {0.353} & \rowcolor{gray!20} \textcolor{red}{\textbf{0.178}} & \textcolor{red}{\textbf{0.241}}\\
&  96 & {0.458} & {0.436} & \rowcolor{gray!20} \textcolor{red}{\textbf{0.221}} & \textcolor{red}{\textbf{0.260}}\\
\cmidrule{2-6}
&  Avg.& {0.254} & {0.306} & \rowcolor{gray!20} \textcolor{red}{\textbf{0.152}} & \textcolor{red}{\textbf{0.231}}\\
\midrule
\multirow{5.5}{*}{ECL}
&  96 &  {0.148} & {0.240} & \rowcolor{gray!20} \textcolor{red}{\textbf{0.140}} & \textcolor{red}{\textbf{0.235}}\\
&  192 & {0.167} & {0.258} & \rowcolor{gray!20} \textcolor{red}{\textbf{0.158}} & \textcolor{red}{\textbf{0.252}}\\
&  336 & {0.179} & {0.272} & \rowcolor{gray!20} \textcolor{red}{\textbf{0.172}} & \textcolor{red}{\textbf{0.267}}\\
&  720 & {0.220} & {0.310} & \rowcolor{gray!20} \textcolor{red}{\textbf{0.202}} & \textcolor{red}{\textbf{0.295}}\\
\cmidrule{2-6}
&  Avg.& {0.179} & {0.270} & \rowcolor{gray!20} \textcolor{red}{\textbf{0.168}} & \textcolor{red}{\textbf{0.262}}\\
\midrule
\multirow{5.5}{*}{Traffic}
&  96 &  {0.395} & {0.268} & \rowcolor{gray!20} \textcolor{red}{\textbf{0.391}} & \textcolor{red}{\textbf{0.266}}\\
&  192 & {0.417} & {0.277} & \rowcolor{gray!20} \textcolor{red}{\textbf{0.409}} & \textcolor{red}{\textbf{0.275}}\\
&  336 & {0.433} & {0.283} & \rowcolor{gray!20} \textcolor{red}{\textbf{0.426}} & \textcolor{red}{\textbf{0.282}}\\
&  720 & {0.467} & {0.300} & \rowcolor{gray!20} \textcolor{red}{\textbf{0.460}} & \textcolor{red}{\textbf{0.300}}\\
\cmidrule{2-6}
&  Avg.& {0.428} & {0.282} & \rowcolor{gray!20} \textcolor{red}{\textbf{0.422}} & \textcolor{red}{\textbf{0.281}}\\
\midrule
\midrule
\multicolumn{2}{c}{Best Count (/52)} &\rowcolor{yellow!20} 2 & 2 & 50 & 51 \\
\bottomrule
\end{NiceTabular}
\end{adjustbox}
\captionsetup{type=table}
\label{tbl:itrans_full2}
\end{minipage}
\captionsetup{type=table}

\caption{TS forecasting results with \textbf{iTransformer} with $L=96$.}
\label{tbl:itrans_full}
\vspace{-40pt}
\end{figure*}

\clearpage
\begin{figure*}[h]
    \centering
    \begin{minipage}{0.44\textwidth}
        \centering
        \begin{adjustbox}{max width=1.00\textwidth}
        \begin{NiceTabular}{c|c|cc|cc}
        \toprule
        \multicolumn{2}{c}{\multirow{2.5}{*}{Metric}} & 
        \multicolumn{2}{c}{CARD} &
        \multicolumn{2}{c}{+ CM} 
        \\
        \cmidrule(lr){3-4} \cmidrule(lr){5-6}
        \multicolumn{2}{c}{ }  & {MSE} & {MAE}  & {MSE} & {MAE}  \\
        \toprule
        \multirow{5.5}{*}{ETTh1}
        &  96 &  0.383 & 0.391 & \rowcolor{gray!20} \first{0.378} & \first{0.386}\\
        &  192 & 0.437 & 0.420 & \rowcolor{gray!20} \first{0.431} & \first{0.417}\\
        &  336 & 0.477 & 0.440 & \rowcolor{gray!20} \first{0.472} & \first{0.438}\\
        &  720 & 0.477 & 0.463 & \rowcolor{gray!20} \first{0.471} & \first{0.459}\\
        \cmidrule{2-6}
        &  Avg.& 0.444 & 0.429 & \rowcolor{gray!20} \first{0.438} & \first{0.425}\\
        \midrule
        \multirow{5.5}{*}{ETTh2}
        &  96 &  0.281 & \first{0.329} & \rowcolor{gray!20} \first{0.279} & \first{0.329}\\
        &  192 & 0.359 & 0.381 & \rowcolor{gray!20} \first{0.355} & \first{0.377}\\
        &  336 &  \first{0.395} & \first{0.410} & \rowcolor{gray!20} 0.403 & 0.415 \\
        &  720 &  \first{0.403} & 0.427 & \rowcolor{gray!20} \first{0.403} & \first{0.426}\\
        \cmidrule{2-6}
        &  Avg.&  0.360 & 0.387 & \rowcolor{gray!20} \first{0.360} & \first{0.387}\\
        \midrule
        \multirow{5.5}{*}{ETTm1}
        &  96 &  0.314 & 0.345 & \rowcolor{gray!20} \first{0.312} & \first{0.343}\\
        &  192 & 0.361 & \first{0.368} & \rowcolor{gray!20} \first{0.360} & \first{0.368}\\
        &  336 & 0.394 & 0.390 & \rowcolor{gray!20} \first{0.389} & \first{0.388}\\
        &  720 & 0.463 & 0.426 & \rowcolor{gray!20} \first{0.455} & \first{0.425}\\
        \cmidrule{2-6}
        &  Avg.& 0.383 & 0.382 & \rowcolor{gray!20} \first{0.379} & \first{0.381}\\
        \midrule
        \multirow{5.5}{*}{ETTm2}
        &  96 &  0.170 & 0.248 & \rowcolor{gray!20} \first{0.167} & \first{0.246}\\
        &  192 & 0.234 & 0.291 & \rowcolor{gray!20} \first{0.231} & \first{0.289}\\
        &  336 & 0.293 & \first{0.328} & \rowcolor{gray!20} \first{0.291} & \first{0.328}\\
        &  720 & \first{0.391} & \first{0.388} & \rowcolor{gray!20} \first{0.391} & \first{0.388}\\
        \cmidrule{2-6}
        &  Avg.& 0.272 & 0.314 & \rowcolor{gray!20} \first{0.270} & \first{0.313}\\
        \midrule
        \multirow{5.5}{*}{Exchange}
        &  96 &  0.083 & 0.200 & \rowcolor{gray!20} \first{0.082} & \first{0.199}\\
        &  192 & 0.178 & 0.298 & \rowcolor{gray!20} \first{0.175} & \first{0.297}\\
        &  336 & 0.350 & 0.425 & \rowcolor{gray!20} \first{0.336} & \first{0.417}\\
        &  720 & 0.870 & \first{0.703} & \rowcolor{gray!20} \first{0.869} & 0.704 \\
        \cmidrule{2-6}
        &  Avg.& 0.370 & 0.407 & \rowcolor{gray!20} \first{0.365} & \first{0.404}\\
        \midrule
        \multirow{5.5}{*}{Weather}
        &  96 &  0.152 & 0.190 & \rowcolor{gray!20} \first{0.150} & \first{0.188}\\
        &  192 & 0.203 & 0.239 & \rowcolor{gray!20} \first{0.200} & \first{0.236}\\
        &  336 & 0.261 & \first{0.283} & \rowcolor{gray!20} \first{0.260} & \first{0.283}\\
        &  720 & 0.343 & 0.337 & \rowcolor{gray!20} \first{0.342} & \first{0.336}\\
        \cmidrule{2-6}
        &  Avg.&  0.240 & 0.262 & \rowcolor{gray!20} \first{0.238} & \first{0.261}\\
        \midrule
        \multirow{5.5}{*}{Solar}
        &  96 &  \first{0.244} & \first{0.254} & \rowcolor{gray!20} 0.248 & 0.255\\
        &  192 & 0.305 & 0.288 & \rowcolor{gray!20} \first{0.289} & \first{0.276}\\
        &  336 & 0.347 & 0.307 & \rowcolor{gray!20} \first{0.318} & \first{0.296}\\
        &  720 & \first{0.318} & 0.298 & \rowcolor{gray!20} 0.328 & \first{0.297}\\
        \cmidrule{2-6}
        &  Avg.& 0.304 & 0.287 & \rowcolor{gray!20} \first{0.296} & \first{0.281}\\
        \bottomrule
        \end{NiceTabular}
        \end{adjustbox}
        \captionsetup{type=table}
    \end{minipage}
    \hfill
    \begin{minipage}{0.44\textwidth}
        \centering
        \begin{adjustbox}{max width=1.00\textwidth}
        \begin{NiceTabular}{c|c|cc|cc}
\toprule
\multicolumn{2}{c}{\multirow{2.5}{*}{Metric}} & 
\multicolumn{2}{c}{CARD} &
\multicolumn{2}{c}{+ CM} 
\\
\cmidrule(lr){3-4} \cmidrule(lr){5-6}
\multicolumn{2}{c}{ }  & {MSE} & {MAE}  & {MSE} & {MAE}  \\
\toprule
\multirow{5.5}{*}{PEMS03}
&  12 & 0.091 & 0.202 & \rowcolor{gray!20} \first{0.089} & \first{0.201}\\
&  24 & \first{0.139} & \first{0.251} & \rowcolor{gray!20} \first{0.139} & 0.253\\
&  48 & 0.244 & 0.345 & \rowcolor{gray!20} \first{0.238} & \first{0.338}\\
&  96 & 0.480 & 0.517 & \rowcolor{gray!20} \first{0.420} & \first{0.478}\\
\cmidrule{2-6}
&  Avg.& 0.239 & 0.329 & \rowcolor{gray!20} \first{0.221} & \first{0.318}\\
\midrule
\multirow{5.5}{*}{PEMS04}
&  12 & \first{0.107} & 0.218 & \rowcolor{gray!20} \first{0.107} & \first{0.216}\\
&  24 & 0.173 & 0.278 & \rowcolor{gray!20} \first{0.166} & \first{0.274}\\
&  48 &  0.309 & 0.380 & \rowcolor{gray!20} \first{0.285} & \first{0.363}\\
&  96 &  0.514 & 0.534 & \rowcolor{gray!20} \first{0.473} & \first{0.502}\\
\cmidrule{2-6}
&  Avg.& 0.276 & 0.353 & \rowcolor{gray!20} \first{0.258} & \first{0.339}\\
\midrule
\multirow{5.5}{*}{PEMS07}
&  12 & 0.080 & 0.187 & \rowcolor{gray!20} \first{0.078} & \first{0.185}\\
&  24 & \first{0.134} & \first{0.241} & \rowcolor{gray!20} \first{0.134} & \first{0.241}\\
&  48 & 0.240 & 0.339 & \rowcolor{gray!20} \first{0.239} & \first{0.338}\\
&  96 & 0.387 & 0.451 & \rowcolor{gray!20} \first{0.383} & \first{0.448}\\
\cmidrule{2-6}
&  Avg.& 0.210 & 0.305 & \rowcolor{gray!20} \first{0.208} & \first{0.303}\\
\midrule
\multirow{5.5}{*}{PEMS08}
&  12 & 0.103 & 0.210 & \rowcolor{gray!20} \first{0.099} & \first{0.205}\\
&  24 & 0.165 & 0.270 & \rowcolor{gray!20} \first{0.159} & \first{0.263}\\
&  48 & 0.311 & 0.390 & \rowcolor{gray!20} \first{0.281} & \first{0.359}\\
&  96 & 0.628 & 0.561 & \rowcolor{gray!20} \first{0.545} & \first{0.517}\\
\cmidrule{2-6}
&  Avg.& 0.302 & 0.358 & \rowcolor{gray!20} \first{0.271} & \first{0.336}\\
\midrule
\multirow{5.5}{*}{ECL}
&  96 & 0.146 & 0.236 & \rowcolor{gray!20} \first{0.145} & \first{0.235}\\
&  192 & \first{0.165} & \first{0.248} & \rowcolor{gray!20} \first{0.165} & \first{0.248}\\
&  336 & 0.179 & 0.270 & \rowcolor{gray!20} \first{0.174} & \first{0.264}\\
&  720 & 0.216 & 0.299 & \rowcolor{gray!20} \first{0.205} & \first{0.290}\\
\cmidrule{2-6}
&  Avg.& 0.177 & 0.263 & \rowcolor{gray!20} \first{0.171} & \first{0.259}\\
\midrule
\multirow{5.5}{*}{Traffic}
&  96 &  0.419 & 0.269 & \rowcolor{gray!20} \first{0.402} & \first{0.254}\\
&  192 & 0.443 & 0.275 & \rowcolor{gray!20} \first{0.424} & \first{0.263}\\
&  336 & 0.459 & 0.283 & \rowcolor{gray!20} \first{0.444} & \first{0.270}\\
&  720 & 0.489 & 0.298 & \rowcolor{gray!20} \first{0.473} & \first{0.286}\\
\cmidrule{2-6}
&  Avg.& 0.453 & 0.281 & \rowcolor{gray!20} \first{0.436} & \first{0.268}\\
\midrule
\midrule
\multicolumn{2}{c}{Best Count (/52)} &\rowcolor{yellow!20} 9 & 11 & \first{49} & \first{48} \\
\bottomrule
\end{NiceTabular}
\end{adjustbox}
\captionsetup{type=table}
\end{minipage}
\captionsetup{type=table}
\caption{TS forecasting results with \textbf{CARD} with $L=96$.}
\label{tbl:CARD_full96}
\end{figure*}

\clearpage
\begin{figure*}[h]
    \centering
    \begin{minipage}{0.44\textwidth}
        \centering
        \begin{adjustbox}{max width=1.00\textwidth}
        \begin{NiceTabular}{c|c|cc|cc}
        \toprule
        \multicolumn{2}{c}{\multirow{2.5}{*}{Metric}} & 
        \multicolumn{2}{c}{CARD} &
        \multicolumn{2}{c}{+ CM} 
        \\
        \cmidrule(lr){3-4} \cmidrule(lr){5-6}
        \multicolumn{2}{c}{ }  & {MSE} & {MAE}  & {MSE} & {MAE}  \\
        \toprule
        \multirow{5.5}{*}{ETTh1}
        &  96 &  0.372 & \first{0.400} & \rowcolor{gray!20} \first{0.370} & \first{0.400}\\
        &  192 & \first{0.408} & \first{0.425} & \rowcolor{gray!20} \first{0.408} & \first{0.424}\\
        &  336 & 0.418 & 0.431 & \rowcolor{gray!20} \first{0.417} & \first{0.430}\\
        &  720 & 0.427 & 0.457 & \rowcolor{gray!20} \first{0.424} & \first{0.455}\\
        \cmidrule{2-6}
        &  Avg.& 0.406 & 0.428 & \rowcolor{gray!20} \first{0.405} & \first{0.427}\\
        \midrule
        \multirow{5.5}{*}{ETTh2}
        &  96 &  0.269 & 0.330 & \rowcolor{gray!20} \first{0.267} & \first{0.329}\\
        &  192 & 0.338 & 0.376 & \rowcolor{gray!20} \first{0.335} & \first{0.373}\\
        &  336 &  0.330 & \first{0.380} & \rowcolor{gray!20} \first{0.328} & \first{0.380}\\
        &  720 &  0.384 & \first{0.424} & \rowcolor{gray!20} \first{0.382} & \first{0.424}\\
        \cmidrule{2-6}
        &  Avg.&  0.330 & 0.378 & \rowcolor{gray!20} \first{0.328} & \first{0.377}\\
        \midrule
        \multirow{5.5}{*}{ETTm1}
        &  96 &  0.288 & \first{0.331} & \rowcolor{gray!20} \first{0.285} & \first{0.331}\\
        &  192 & 0.330 & 0.357 & \rowcolor{gray!20} \first{0.328} & \first{0.356}\\
        &  336 & 0.364 & 0.379 & \rowcolor{gray!20} \first{0.362} & \first{0.377}\\
        &  720 & 0.418 & \first{0.411} & \rowcolor{gray!20} \first{0.415} & \first{0.411}\\
        \cmidrule{2-6}
        
        &  Avg.& 0.350 & 0.370 & \rowcolor{gray!20} \first{0.347} & \first{0.369}\\
        \midrule
        \multirow{5.5}{*}{ETTm2}
        &  96 &  0.161 & 0.247 & \rowcolor{gray!20} \first{0.160} & \first{0.246}\\
        &  192 & 0.217 & 0.287 & \rowcolor{gray!20} \first{0.215} & \first{0.285}\\
        &  336 & 0.269 & 0.322 & \rowcolor{gray!20} \first{0.266} & \first{0.320}\\
        &  720 & 0.362 & 0.381 & \rowcolor{gray!20} \first{0.359} & \first{0.380}\\
        \cmidrule{2-6}
        &  Avg.& 0.252 & 0.309 & \rowcolor{gray!20} \first{0.250} & \first{0.308}\\
        \midrule
        \multirow{5.5}{*}{Exchange}
        &  96 &  0.104 & 0.228 & \rowcolor{gray!20} \first{0.100} & \first{0.224}\\
        &  192 & 0.255 & 0.365 & \rowcolor{gray!20} \first{0.246} & \first{0.357}\\
        &  336 & 0.405 & 0.464 & \rowcolor{gray!20} \first{0.381} & \first{0.449}\\
        &  720 & 0.937 & 0.735 & \rowcolor{gray!20} \first{0.796} & \first{0.657}\\
        \cmidrule{2-6}
        &  Avg.& 0.425 & 0.448 & \rowcolor{gray!20} \first{0.381} & \first{0.422}\\
        \midrule
        \multirow{5.5}{*}{Weather}
        &  96 &  0.145 & 0.188 & \rowcolor{gray!20} \first{0.144} & \first{0.187}\\
        &  192 & 0.189 & 0.230 & \rowcolor{gray!20} \first{0.188} & \first{0.229}\\
        &  336 & 0.243 & 0.274 & \rowcolor{gray!20} \first{0.240} & \first{0.271}\\
        &  720 & 0.316 & 0.329 & \rowcolor{gray!20} \first{0.308} & \first{0.322}\\
        \cmidrule{2-6}
        &  Avg.&  0.226 & 0.255 & \rowcolor{gray!20} \first{0.220} & \first{0.252}\\
        \midrule
        \multirow{5.5}{*}{Solar}
        &  96 &  0.171 & 0.211 & \rowcolor{gray!20} \first{0.164} & \first{0.210}\\
        &  192 & 0.210 & 0.234 & \rowcolor{gray!20} \first{0.201} & \first{0.229}\\
        &  336 & 0.227 & 0.246 & \rowcolor{gray!20} \first{0.225} & \first{0.242}\\
        &  720 & 0.235 & 0.280 & \rowcolor{gray!20} \first{0.228} & \first{0.244}\\
        \cmidrule{2-6}
        &  Avg.& 0.211 & 0.243 & \rowcolor{gray!20} \first{0.204} & \first{0.231}\\
        \bottomrule
        \end{NiceTabular}
        \end{adjustbox}
        \captionsetup{type=table}
    \end{minipage}
    \hfill
    \begin{minipage}{0.44\textwidth}
        \centering
        \begin{adjustbox}{max width=1.00\textwidth}
        \begin{NiceTabular}{c|c|cc|cc}
\toprule
\multicolumn{2}{c}{\multirow{2.5}{*}{Metric}} & 
\multicolumn{2}{c}{CARD} &
\multicolumn{2}{c}{+ CM} 
\\
\cmidrule(lr){3-4} \cmidrule(lr){5-6}
\multicolumn{2}{c}{ }  & {MSE} & {MAE}  & {MSE} & {MAE}  \\
\toprule
\multirow{5.5}{*}{PEMS03}
&  12 & \first{0.064} & \first{0.168} & \rowcolor{gray!20} \first{0.064} & \first{0.168}\\
&  24 & 0.095 & 0.202 & \rowcolor{gray!20} \first{0.083} & \first{0.188}\\
&  48 & 0.145 & 0.251 & \rowcolor{gray!20} \first{0.139} & \first{0.247}\\
&  96 & 0.167 & 0.264 & \rowcolor{gray!20} \first{0.163} & \first{0.261}\\
\cmidrule{2-6}
&  Avg.& 0.116 & 0.220 & \rowcolor{gray!20} \first{0.112} & \first{0.216}\\
\midrule
\multirow{5.5}{*}{PEMS04}
&  12 & 0.081 & 0.185 & \rowcolor{gray!20} \first{0.079} & \first{0.181}\\
&  24 & 0.100 & 0.205 & \rowcolor{gray!20} \first{0.098} & \first{0.201}\\
&  48 &  0.152 & 0.250 & \rowcolor{gray!20} \first{0.122} & \first{0.224}\\
&  96 &  0.146 & 0.245 & \rowcolor{gray!20} \first{0.137} & \first{0.235}\\
\cmidrule{2-6}
&  Avg.& 0.120 & 0.222 & \rowcolor{gray!20} \first{0.109} & \first{0.210}\\
\midrule
\multirow{5.5}{*}{PEMS07}
&  12 & 0.060 & 0.160 & \rowcolor{gray!20} \first{0.055} & \first{0.154}\\
&  24 & 0.079 & 0.184 & \rowcolor{gray!20} \first{0.075} & \first{0.180}\\
&  48 & 0.094 & 0.197 & \rowcolor{gray!20} \first{0.088} & \first{0.189}\\
&  96 & 0.112 & 0.219 & \rowcolor{gray!20} \first{0.103} & \first{0.202}\\
\cmidrule{2-6}
&  Avg.& 0.087 & 0.190 & \rowcolor{gray!20} \first{0.080} & \first{0.181}\\
\midrule
\multirow{5.5}{*}{PEMS08}
&  12 & 0.085 & 0.180 & \rowcolor{gray!20} \first{0.080} & \first{0.177}\\
&  24 & 0.111 & 0.199 & \rowcolor{gray!20} \first{0.109} & \first{0.198}\\
&  48 & 0.167 & 0.228 & \rowcolor{gray!20} \first{0.160} & \first{0.224}\\
&  96 & 0.241 & 0.246 & \rowcolor{gray!20} \first{0.235} & \first{0.238}\\
\cmidrule{2-6}
&  Avg.& 0.148 & 0.213 & \rowcolor{gray!20} \first{0.146} & \first{0.209}\\
\midrule
\midrule
\multicolumn{2}{c}{Best Count (/44)} &\rowcolor{yellow!20} 2 & 7 & 44 & 44 \\
\bottomrule
\end{NiceTabular}
\end{adjustbox}
\captionsetup{type=table}
\end{minipage}
\captionsetup{type=table}
\caption{TS forecasting results with \textbf{CARD} with $L=720$.}
\label{tbl:CARD_full720}
\end{figure*}

\clearpage
\begin{figure*}[h]
    \centering
    \begin{minipage}{0.44\textwidth}
        \centering
        \begin{adjustbox}{max width=1.00\textwidth}
        \begin{NiceTabular}{c|c|cc|cc}
        \toprule
        \multicolumn{2}{c}{\multirow{2.5}{*}{Metric}} & 
        \multicolumn{2}{c}{PRformer} &
        \multicolumn{2}{c}{+ CM} 
        \\
        \cmidrule(lr){3-4} \cmidrule(lr){5-6}
        \multicolumn{2}{c}{ }  & {MSE} & {MAE}  & {MSE} & {MAE}  \\
        \toprule
        \multirow{5.5}{*}{ETTh1}
        &  96 &  0.361 & 0.386 & \rowcolor{gray!20} \first{0.360} & \first{0.384}\\
        &  192 & \first{0.409} & \first{0.416} & \rowcolor{gray!20} \first{0.409} & 0.419\\
        &  336 & 0.467 & 0.457 & \rowcolor{gray!20} \first{0.456} & \first{0.446}\\
        &  720 & 0.540 & 0.522 & \rowcolor{gray!20} \first{0.508} & \first{0.497}\\
        \cmidrule{2-6}
        &  Avg.& 0.444 & 0.445 & \rowcolor{gray!20} \first{0.434} & \first{0.436}\\
        \midrule
        \multirow{5.5}{*}{ETTh2}
        &  96 &  0.284 & 0.335 & \rowcolor{gray!20} \first{0.278} & \first{0.332}\\
        &  192 & 0.336 & \first{0.374} & \rowcolor{gray!20} \first{0.334} & 0.378\\
        &  336 &  0.363 & 0.398 & \rowcolor{gray!20} \first{0.358} & \first{0.395}\\
        &  720 &  0.398 & 0.424 & \rowcolor{gray!20} \first{0.389} & \first{0.420}\\
        \cmidrule{2-6}
        &  Avg.&  0.345 & 0.383 & \rowcolor{gray!20} \first{0.340} & \first{0.381}\\
        \midrule
        \multirow{5.5}{*}{ETTm1}
        &  96 &  0.282 & 0.335 & \rowcolor{gray!20} \first{0.276} & \first{0.330}\\
        &  192 & 0.332 & 0.363 & \rowcolor{gray!20} \first{0.325} & \first{0.360}\\
        &  336 & 0.365 & 0.385 & \rowcolor{gray!20} \first{0.359} & \first{0.381}\\
        &  720 & 0.427 & 0.421 & \rowcolor{gray!20} \first{0.414} & \first{0.414}\\
        \cmidrule{2-6}
        &  Avg.& 0.352 & 0.376 & \rowcolor{gray!20} \first{0.343} & \first{0.371}\\
        \midrule
        \multirow{5.5}{*}{ETTm2}
        &  96 &  \first{0.164} & \first{0.248} & \rowcolor{gray!20} \first{0.164} & \first{0.248}\\
        &  192 & \first{0.225} & \first{0.292} & \rowcolor{gray!20} 0.230 & 0.295\\
        &  336 & 0.287 & 0.336 & \rowcolor{gray!20} \first{0.285} & \first{0.335}\\
        &  720 & 0.387 & 0.399 & \rowcolor{gray!20} \first{0.374} & \first{0.393}\\
        \cmidrule{2-6}
        &  Avg.& 0.266 & 0.319 & \rowcolor{gray!20} \first{0.262} & \first{0.317}\\
        \midrule
        \multirow{5.5}{*}{Exchange}
        &  96 &  0.093 & 0.215 & \rowcolor{gray!20} \first{0.084} & \first{0.204}\\
        &  192 & 0.211 & 0.328 & \rowcolor{gray!20} \first{0.181} & \first{0.304}\\
        &  336 & 0.396 & 0.468 & \rowcolor{gray!20} \first{0.371} & \first{0.449}\\
        &  720 & 1.558 & 0.952 & \rowcolor{gray!20} \first{1.251} & \first{0.840}\\
        \cmidrule{2-6}
        &  Avg.& 0.565 & 0.491 & \rowcolor{gray!20} \first{0.472} & \first{0.449}\\
        \midrule
        \multirow{5.5}{*}{Weather}
        &  96 &  0.146 & 0.188 & \rowcolor{gray!20} \first{0.143} & \first{0.183}\\
        &  192 & 0.192 & 0.233 & \rowcolor{gray!20} \first{0.189} & \first{0.230}\\
        &  336 & 0.245 & 0.277 & \rowcolor{gray!20} \first{0.239} & \first{0.271}\\
        &  720 & 0.310 & 0.324 & \rowcolor{gray!20} \first{0.306} & \first{0.321}\\
        \cmidrule{2-6}
        &  Avg.&  0.224 & 0.256 & \rowcolor{gray!20} \first{0.219} & \first{0.251}\\
        \midrule
        \multirow{5.5}{*}{Solar}
        &  96 &  0.170 & 0.199 & \rowcolor{gray!20} \first{0.167} & \first{0.195}\\
        &  192 & 0.193 & 0.213 & \rowcolor{gray!20} \first{0.188} & \first{0.209}\\
        &  336 & 0.217 & 0.227 & \rowcolor{gray!20} \first{0.211} & \first{0.220}\\
        &  720 & 0.229 & 0.234 & \rowcolor{gray!20} \first{0.222} & \first{0.226}\\
        \cmidrule{2-6}
        &  Avg.& 0.202 & 0.218 & \rowcolor{gray!20} \first{0.197} & \first{0.212}\\
        \bottomrule
        \end{NiceTabular}
        \end{adjustbox}
        \captionsetup{type=table}
    \end{minipage}
    \hfill
    \begin{minipage}{0.44\textwidth}
        \centering
        \begin{adjustbox}{max width=1.00\textwidth}
        \begin{NiceTabular}{c|c|cc|cc}
\toprule
\multicolumn{2}{c}{\multirow{2.5}{*}{Metric}} & 
\multicolumn{2}{c}{PRformer} &
\multicolumn{2}{c}{+ CM} 
\\
\cmidrule(lr){3-4} \cmidrule(lr){5-6}
\multicolumn{2}{c}{ }  & {MSE} & {MAE}  & {MSE} & {MAE}  \\
\toprule
\multirow{5.5}{*}{PEMS03}
&  12 & 0.077 & 0.182 & \rowcolor{gray!20} \first{0.072} & \first{0.177}\\
&  24 & 0.106 & 0.215 & \rowcolor{gray!20} \first{0.103} & \first{0.211}\\
&  48 & 0.164 & 0.271 & \rowcolor{gray!20} \first{0.161} & \first{0.269}\\
&  96 & 0.259 & 0.346 & \rowcolor{gray!20} \first{0.238} & \first{0.337}\\
\cmidrule{2-6}
&  Avg.& 0.152 & 0.254 & \rowcolor{gray!20} \first{0.143} & \first{0.248}\\
\midrule
\multirow{5.5}{*}{PEMS04}
&  12 & 0.094 & 0.199 & \rowcolor{gray!20} \first{0.088} & \first{0.193}\\
&  24 & 0.135 & 0.240 & \rowcolor{gray!20} \first{0.123} & \first{0.231}\\
&  48 &  0.213 & 0.306 & \rowcolor{gray!20} \first{0.190} & \first{0.292}\\
&  96 &  0.300 & 0.379 & \rowcolor{gray!20} \first{0.292} & \first{0.373}\\
\cmidrule{2-6}
&  Avg.& 0.185 & 0.281 & \rowcolor{gray!20} \first{0.173} & \first{0.272}\\
\midrule
\multirow{5.5}{*}{PEMS07}
&  12 & 0.071 & 0.167 & \rowcolor{gray!20} \first{0.064} & \first{0.157}\\
&  24 & 0.100 & 0.201 & \rowcolor{gray!20} \first{0.093} & \first{0.195}\\
&  48 & 0.153 & 0.255 & \rowcolor{gray!20} \first{0.145} & \first{0.249}\\
&  96 & 0.235 & 0.331 & \rowcolor{gray!20} \first{0.225} & \first{0.318}\\
\cmidrule{2-6}
&  Avg.& 0.140 & 0.239 & \rowcolor{gray!20} \first{0.132} & \first{0.230}\\
\midrule
\multirow{5.5}{*}{PEMS08}
&  12 & 0.087 & 0.189 & \rowcolor{gray!20} \first{0.083} & \first{0.185}\\
&  24 & 0.127 & 0.229 & \rowcolor{gray!20} \first{0.119} & \first{0.222}\\
&  48 & 0.211 & 0.301 & \rowcolor{gray!20} \first{0.193} & \first{0.287}\\
&  96 & 0.394 & 0.414 & \rowcolor{gray!20} \first{0.349} & \first{0.390}\\
\cmidrule{2-6}
&  Avg.& 0.205 & 0.283 & \rowcolor{gray!20} \first{0.186} & \first{0.271}\\
\midrule
\multirow{5.5}{*}{ECL}
&  96 & 0.127 & 0.216 & \rowcolor{gray!20} \first{0.124} & \first{0.214}\\
&  192 & 0.148 & 0.237 & \rowcolor{gray!20} \first{0.144} & \first{0.234}\\
&  336 & 0.161 & 0.254 & \rowcolor{gray!20} \first{0.157} & \first{0.250}\\
&  720 & 0.185 & 0.278 & \rowcolor{gray!20} \first{0.172} & \first{0.269}\\
\cmidrule{2-6}
&  Avg.& 0.155 & 0.246 & \rowcolor{gray!20} \first{0.149} & \first{0.242}\\
\midrule
\multirow{5.5}{*}{Traffic}
&  96 &  0.349 & 0.222 & \rowcolor{gray!20} \first{0.326} & \first{0.214}\\
&  192 & 0.357 & 0.225 & \rowcolor{gray!20} \first{0.332} & \first{0.220}\\
&  336 & 0.385 & 0.239 & \rowcolor{gray!20} \first{0.348} & \first{0.228}\\
&  720 & 0.421 & 0.258 & \rowcolor{gray!20} \first{0.388} & \first{0.245}\\
\cmidrule{2-6}
&  Avg.& 0.378 & 0.236 & \rowcolor{gray!20} \first{0.348} & \first{0.227}\\
\midrule
\midrule
\multicolumn{2}{c}{Best Count (/52)} &\rowcolor{yellow!20} 3 & 4 & \first{51} & \first{49} \\
\bottomrule
\end{NiceTabular}
\end{adjustbox}
\captionsetup{type=table}
\end{minipage}
\captionsetup{type=table}
\caption{TS forecasting results with \textbf{PRformer} with varying $L$ by datasets.}
\label{tbl:prformer_full}
\end{figure*}

\clearpage
\begin{figure*}[h]
    \centering
    \begin{minipage}{0.44\textwidth}
        \centering
        \begin{adjustbox}{max width=1.00\textwidth}
        \begin{NiceTabular}{c|c|cc|cc}
        \toprule
        \multicolumn{2}{c}{\multirow{2.5}{*}{Metric}} & 
        \multicolumn{2}{c}{Minusformer} &
        \multicolumn{2}{c}{+ CM} 
        \\
        \cmidrule(lr){3-4} \cmidrule(lr){5-6}
        \multicolumn{2}{c}{ }  & {MSE} & {MAE}  & {MSE} & {MAE}  \\
        \toprule
        \multirow{5.5}{*}{ETTh1}
        &  96 &  \first{0.387} & \first{0.404} & \rowcolor{gray!20} \first{0.387} & \first{0.404}\\
        &  192 & \first{0.446} & 0.437 & \rowcolor{gray!20} \first{0.446} & \first{0.436}\\
        &  336 & 0.502 & 0.473 & \rowcolor{gray!20} \first{0.488} & \first{0.464}\\
        &  720 & 0.517 & 0.494 & \rowcolor{gray!20} \first{0.485} & \first{0.481}\\
        \cmidrule{2-6}
        &  Avg.& 0.463 & 0.452 & \rowcolor{gray!20} \first{0.451} & \first{0.446}\\
        \midrule
        \multirow{5.5}{*}{ETTh2}
        &  96 &  0.310 & \first{0.352} & \rowcolor{gray!20} \first{0.306} & \first{0.352}\\
        &  192 & 0.390 & 0.403 & \rowcolor{gray!20} \first{0.379} & \first{0.397}\\
        &  336 &  0.454 & 0.443 & \rowcolor{gray!20} \first{0.438} & \first{0.437}\\
        &  720 &  0.420 & \first{0.437} & \rowcolor{gray!20} \first{0.417} & \first{0.437}\\
        \cmidrule{2-6}
        &  Avg.&  0.394 & 0.409 & \rowcolor{gray!20} \first{0.385} & \first{0.406}\\
        \midrule
        \multirow{5.5}{*}{ETTm1}
        &  96 &  0.341 & 0.371 & \rowcolor{gray!20} \first{0.325} & \first{0.362}\\
        &  192 & 0.380 & 0.390 & \rowcolor{gray!20} \first{0.370} & \first{0.385}\\
        &  336 & \first{0.437} & \first{0.424} & \rowcolor{gray!20} 0.441 & 0.430\\
        &  720 & 0.507 & 0.462 & \rowcolor{gray!20} \first{0.479} & \first{0.454}\\
        \cmidrule{2-6}
        &  Avg.& 0.416 & 0.412 & \rowcolor{gray!20} \first{0.404} & \first{0.408}\\
        \midrule
        \multirow{5.5}{*}{ETTm2}
        &  96 &  0.180 & \first{0.260} & \rowcolor{gray!20} \first{0.179} & \first{0.260}\\
        &  192 & 0.243 & \first{0.303} & \rowcolor{gray!20} \first{0.242} & \first{0.303}\\
        &  336 & 0.309 & 0.345 & \rowcolor{gray!20} \first{0.304} & \first{0.343}\\
        &  720 & 0.407 & 0.403 & \rowcolor{gray!20} \first{0.406} & \first{0.403}\\
        \cmidrule{2-6}
        &  Avg.& 0.285 & 0.328 & \rowcolor{gray!20} \first{0.283} & \first{0.327}\\
        \midrule
        \multirow{5.5}{*}{Exchange}
        &  96 &  \first{0.096} & \first{0.227} & \rowcolor{gray!20} \first{0.096} & 0.229\\
        &  192 & 0.222 & 0.353 & \rowcolor{gray!20} \first{0.220} & \first{0.351}\\
        &  336 & 0.463 & 0.516 & \rowcolor{gray!20} \first{0.416} & \first{0.487}\\
        &  720 & 1.251 & 0.856 & \rowcolor{gray!20} \first{1.130} & \first{0.824}\\
        \cmidrule{2-6}
        &  Avg.& 0.508 & 0.488 & \rowcolor{gray!20} \first{0.465} & \first{0.472}\\
        \midrule
        \multirow{5.5}{*}{Weather}
        &  96 &  0.174 & 0.213 & \rowcolor{gray!20} \first{0.166} & \first{0.207}\\
        &  192 & 0.228 & 0.261 & \rowcolor{gray!20} \first{0.217} & \first{0.253}\\
        &  336 & 0.281 & 0.299 & \rowcolor{gray!20} \first{0.274} & \first{0.295}\\
        &  720 & 0.357 & 0.349 & \rowcolor{gray!20} \first{0.351} & \first{0.346}\\
        \cmidrule{2-6}
        &  Avg.&  0.260 & 0.281 & \rowcolor{gray!20} \first{0.252} & \first{0.275}\\
        \midrule
        \multirow{5.5}{*}{Solar}
        &  96 &  0.202 & \first{0.231} & \rowcolor{gray!20} \first{0.200} & \first{0.231}\\
        &  192 & \first{0.227} & \first{0.248} & \rowcolor{gray!20} 0.229 & 0.249\\
        &  336 & 0.245 & 0.264 & \rowcolor{gray!20} \first{0.239} & \first{0.262}\\
        &  720 & 0.246 & 0.267 & \rowcolor{gray!20} \first{0.243} & \first{0.266}\\
        \cmidrule{2-6}
        &  Avg.& 0.230 & 0.253 & \rowcolor{gray!20} \first{0.228} & \first{0.252}\\
        \bottomrule
        \end{NiceTabular}
        \end{adjustbox}
        \captionsetup{type=table}
    \end{minipage}
    \hfill
    \begin{minipage}{0.44\textwidth}
        \centering
        \begin{adjustbox}{max width=1.00\textwidth}
        \begin{NiceTabular}{c|c|cc|cc}
\toprule
\multicolumn{2}{c}{\multirow{2.5}{*}{Metric}} & 
\multicolumn{2}{c}{Minusformer} &
\multicolumn{2}{c}{+ CM} 
\\
\cmidrule(lr){3-4} \cmidrule(lr){5-6}
\multicolumn{2}{c}{ }  & {MSE} & {MAE}  & {MSE} & {MAE}  \\
\toprule
\multirow{5.5}{*}{PEMS03}
&  12 & 0.067 & 0.173 & \rowcolor{gray!20} \first{0.063} & \first{0.167}\\
&  24 & 0.095 & 0.206 & \rowcolor{gray!20} \first{0.085} & \first{0.193}\\
&  48 & 0.149 & 0.261 & \rowcolor{gray!20} \first{0.123} & \first{0.236}\\
&  96 & 0.239 & 0.341 & \rowcolor{gray!20} \first{0.185} & \first{0.297}\\
\cmidrule{2-6}
&  Avg.& 0.138 & 0.245 & \rowcolor{gray!20} \first{0.114} & \first{0.223}\\
\midrule
\multirow{5.5}{*}{PEMS04}
&  12 & 0.085 & 0.190 & \rowcolor{gray!20} \first{0.076} & \first{0.180}\\
&  24 & 0.118 & 0.226 & \rowcolor{gray!20} \first{0.095} & \first{0.205}\\
&  48 &  0.179 & 0.282 & \rowcolor{gray!20} \first{0.127} & \first{0.241}\\
&  96 &  0.303 & 0.382 & \rowcolor{gray!20} \first{0.181} & \first{0.296}\\
\cmidrule{2-6}
&  Avg.& 0.171 & 0.270 & \rowcolor{gray!20} \first{0.120} & \first{0.230}\\
\midrule
\multirow{5.5}{*}{PEMS07}
&  12 & 0.063 & 0.160 & \rowcolor{gray!20} \first{0.057} & \first{0.152}\\
&  24 & 0.090 & 0.192 & \rowcolor{gray!20} \first{0.074} & \first{0.175}\\
&  48 & 0.137 & 0.238 & \rowcolor{gray!20} \first{0.095} & \first{0.200}\\
&  96 & 0.208 & 0.304 & \rowcolor{gray!20} \first{0.130} & \first{0.237}\\
\cmidrule{2-6}
&  Avg.& 0.125 & 0.224 & \rowcolor{gray!20} \first{0.089} & \first{0.191}\\
\midrule
\multirow{5.5}{*}{PEMS08}
&  12 & 0.077 & 0.177 & \rowcolor{gray!20} \first{0.074} & \first{0.174}\\
&  24 & 0.113 & 0.213 & \rowcolor{gray!20} \first{0.101} & \first{0.203}\\
&  48 & 0.182 & 0.272 & \rowcolor{gray!20} \first{0.147} & \first{0.242}\\
&  96 & 0.308 & 0.354 & \rowcolor{gray!20} \first{0.245} & \first{0.318}\\
\cmidrule{2-6}
&  Avg.& 0.170 & 0.254 & \rowcolor{gray!20} \first{0.142} & \first{0.234}\\
\midrule
\multirow{5.5}{*}{ECL}
&  96 & 0.144 & 0.236 & \rowcolor{gray!20} \first{0.136} & \first{0.231}\\
&  192 & 0.161 & 0.252 & \rowcolor{gray!20} \first{0.154} & \first{0.248}\\
&  336 & 0.174 & 0.267 & \rowcolor{gray!20} \first{0.169} & \first{0.265}\\
&  720 & 0.204 & 0.294 & \rowcolor{gray!20} \first{0.196} & \first{0.290}\\
\cmidrule{2-6}
&  Avg.& 0.171 & 0.262 & \rowcolor{gray!20} \first{0.164} & \first{0.258}\\
\midrule
\multirow{5.5}{*}{Traffic}
&  96 &  0.400 & 0.267 & \rowcolor{gray!20} \first{0.393} & \first{0.264}\\
&  192 & 0.400 & 0.264 & \rowcolor{gray!20} \first{0.397} & \first{0.262}\\
&  336 & 0.416 & 0.271 & \rowcolor{gray!20} \first{0.406} & \first{0.267}\\
&  720 & 0.434 & 0.284 & \rowcolor{gray!20} \first{0.425} & \first{0.280}\\
\cmidrule{2-6}
&  Avg.& 0.413 & 0.272 & \rowcolor{gray!20} \first{0.405} & \first{0.268}\\
\midrule
\midrule
\multicolumn{2}{c}{Best Count (/52)} &\rowcolor{yellow!20} 4 & 8 & 51 & 49 \\
\bottomrule
\end{NiceTabular}
\end{adjustbox}
\captionsetup{type=table}
\end{minipage}
\captionsetup{type=table}
\caption{TS forecasting results with \textbf{Minusformer} with $L=96$.}
\label{tbl:minusformer_full96}
\end{figure*}

\clearpage

\begin{figure*}[h]
    \centering
    \begin{minipage}{0.44\textwidth}
        \centering
        \begin{adjustbox}{max width=1.00\textwidth}
        \begin{NiceTabular}{c|c|cc|cc}
        \toprule
        \multicolumn{2}{c}{\multirow{2.5}{*}{Metric}} & 
        \multicolumn{2}{c}{Minusformer} &
        \multicolumn{2}{c}{+ CM} 
        \\
        \cmidrule(lr){3-4} \cmidrule(lr){5-6}
        \multicolumn{2}{c}{ }  & {MSE} & {MAE}  & {MSE} & {MAE}  \\
        \toprule
        \multirow{5.5}{*}{ETTh1}
        &  96 &  0.384 & 0.406 & \rowcolor{gray!20} \first{0.381} & \first{0.404}\\
        &  192 & 0.431 & 0.432 & \rowcolor{gray!20} \first{0.424} & \first{0.427}\\
        &  336 & 0.527 & 0.498 & \rowcolor{gray!20} \first{0.514} & \first{0.491}\\
        &  720 & 0.469 & 0.476 & \rowcolor{gray!20} \first{0.451} & \first{0.470}\\
        \cmidrule{2-6}
        &  Avg.& 0.453 & 0.453 & \rowcolor{gray!20} \first{0.442} & \first{0.448}\\
        \midrule
        \multirow{5.5}{*}{ETTh2}
        &  96 &  0.289 & 0.348 & \rowcolor{gray!20} \first{0.287} & \first{0.347}\\
        &  192 & 0.356 & 0.388 & \rowcolor{gray!20} \first{0.349} & \first{0.386}\\
        &  336 &  0.378 & 0.411 & \rowcolor{gray!20} \first{0.374} & \first{0.408}\\
        &  720 &  0.418 & 0.442 & \rowcolor{gray!20} \first{0.413} & \first{0.439}\\
        \cmidrule{2-6}
        &  Avg.&  0.360 & 0.397 & \rowcolor{gray!20} \first{0.356} & \first{0.395}\\
        \midrule
        \multirow{5.5}{*}{ETTm1}
        &  96 &  0.302 & 0.352 & \rowcolor{gray!20} \first{0.296} & \first{0.348}\\
        &  192 & 0.346 & 0.380 & \rowcolor{gray!20} \first{0.344} & \first{0.378}\\
        &  336 & 0.379 & 0.397 & \rowcolor{gray!20} \first{0.369} & \first{0.392}\\
        &  720 & 0.449 & 0.444 & \rowcolor{gray!20} \first{0.437} & \first{0.437}\\
        \cmidrule{2-6}
        &  Avg.& 0.369 & 0.393 & \rowcolor{gray!20} \first{0.361} & \first{0.389}\\
        \midrule
        \multirow{5.5}{*}{ETTm2}
        &  96 &  0.175 & 0.262 & \rowcolor{gray!20} \first{0.173} & \first{0.261}\\
        &  192 & 0.237 & 0.310 & \rowcolor{gray!20} \first{0.226} & \first{0.301}\\
        &  336 & 0.286 & 0.340 & \rowcolor{gray!20} \first{0.279} & \first{0.335}\\
        &  720 & 0.400 & 0.401 & \rowcolor{gray!20} \first{0.376} & \first{0.391}\\
        \cmidrule{2-6}
        &  Avg.& 0.275 & 0.328 & \rowcolor{gray!20} \first{0.263} & \first{0.322}\\
        \midrule
        \multirow{5.5}{*}{Exchange}
        &  96 &  0.157 & 0.297 & \rowcolor{gray!20} \first{0.136} & \first{0.273}\\
        &  192 & \first{0.291} & \first{0.417} & \rowcolor{gray!20} 0.305 & 0.427 \\
        &  336 & 0.560 & 0.582 & \rowcolor{gray!20} \first{0.460} & \first{0.516}\\
        &  720 & 1.125 & 0.821 & \rowcolor{gray!20} \first{0.997} & \first{0.784}\\
        \cmidrule{2-6}
        &  Avg.& 0.533 & 0.529 & \rowcolor{gray!20} \first{0.474} & \first{0.500}\\
        \midrule
        \multirow{5.5}{*}{Weather}
        &  96 &  0.161 & 0.210 & \rowcolor{gray!20} \first{0.154} & \first{0.205}\\
        &  192 & 0.206 & 0.250 & \rowcolor{gray!20} \first{0.199} & \first{0.247}\\
        &  336 & 0.256 & 0.291 & \rowcolor{gray!20} \first{0.251} & \first{0.285}\\
        &  720 & 0.346 & 0.348 & \rowcolor{gray!20} \first{0.337} & \first{0.344}\\
        \cmidrule{2-6}
        &  Avg.&  0.242 & 0.275 & \rowcolor{gray!20} \first{0.235} & \first{0.270}\\
        \midrule
        \multirow{5.5}{*}{Solar}
        &  96 &  \first{0.216} & \first{0.238} & \rowcolor{gray!20} 0.218 & 0.240\\
        &  192 & 0.206 & 0.255 & \rowcolor{gray!20} \first{0.198} & \first{0.249}\\
        &  336 & 0.218 & 0.268 & \rowcolor{gray!20} \first{0.216} & \first{0.265}\\
        &  720 & 0.220 & 0.269 & \rowcolor{gray!20} \first{0.218} & \first{0.266}\\
        \cmidrule{2-6}
        &  Avg.& 0.215 & 0.258 & \rowcolor{gray!20} \first{0.212} & \first{0.255}\\
        \bottomrule
        \end{NiceTabular}
        \end{adjustbox}
        \captionsetup{type=table}
    \end{minipage}
    \hfill
    \begin{minipage}{0.44\textwidth}
        \centering
        \begin{adjustbox}{max width=1.00\textwidth}
        \begin{NiceTabular}{c|c|cc|cc}
\toprule
\multicolumn{2}{c}{\multirow{2.5}{*}{Metric}} & 
\multicolumn{2}{c}{Minusformer} &
\multicolumn{2}{c}{+ CM} 
\\
\cmidrule(lr){3-4} \cmidrule(lr){5-6}
\multicolumn{2}{c}{ }  & {MSE} & {MAE}  & {MSE} & {MAE}  \\
\toprule
\multirow{5.5}{*}{PEMS03}
&  12 & 0.059 & 0.158 & \rowcolor{gray!20} \first{0.056} & \first{0.156}\\
&  24 & 0.071 & 0.174 & \rowcolor{gray!20} \first{0.070} & \first{0.172}\\
&  48 & 0.104 & 0.213 & \rowcolor{gray!20} \first{0.098} & \first{0.204}\\
&  96 & \first{0.114} & \first{0.216} & \rowcolor{gray!20} 0.120 & 0.220 \\
\cmidrule{2-6}
&  Avg.& 0.087 & 0.190 & \rowcolor{gray!20} \first{0.086} & \first{0.188}\\
\midrule
\multirow{5.5}{*}{PEMS04}
&  12 & 0.072 & 0.171 & \rowcolor{gray!20} \first{0.070} & \first{0.169}\\
&  24 & 0.085 & 0.186 & \rowcolor{gray!20} \first{0.080} & \first{0.182}\\
&  48 &  0.116 & 0.222 & \rowcolor{gray!20} \first{0.109} & \first{0.218}\\
&  96 &  0.125 & 0.221 & \rowcolor{gray!20} \first{0.114} & \first{0.214}\\
\cmidrule{2-6}
&  Avg.& 0.100 & 0.200 & \rowcolor{gray!20} \first{0.093} & \first{0.195}\\
\midrule
\multirow{5.5}{*}{PEMS07}
&  12 & 0.052 & 0.143 & \rowcolor{gray!20} \first{0.050} & \first{0.142}\\
&  24 & 0.061 & 0.154 & \rowcolor{gray!20} \first{0.058} & \first{0.153}\\
&  48 & 0.083 & \first{0.184} & \rowcolor{gray!20} \first{0.080} & \first{0.184}\\
&  96 & \first{0.076} & \first{0.179} & \rowcolor{gray!20} 0.079 & \first{0.179}\\
\cmidrule{2-6}
&  Avg.& 0.068 & 0.165 & \rowcolor{gray!20} \first{0.067} & \first{0.164}\\
\midrule
\multirow{5.5}{*}{PEMS08}
&  12 & 0.066 & 0.170 & \rowcolor{gray!20} \first{0.064} & \first{0.162}\\
&  24 & 0.083 & \first{0.175} & \rowcolor{gray!20} \first{0.080} & 0.176\\
&  48 & 0.137 & 0.221 & \rowcolor{gray!20} \first{0.132} & \first{0.212}\\
&  96 & \first{0.157} & \first{0.213} & \rowcolor{gray!20} 0.159 & 0.216\\
\cmidrule{2-6}
&  Avg.& 0.111 & 0.198 & \rowcolor{gray!20} \first{0.109} & \first{0.191} \\
\midrule
\multirow{5.5}{*}{ECL}
&  96 & 0.131 & 0.224 & \rowcolor{gray!20} \first{0.127} & \first{0.223}\\
&  192 & 0.153 & 0.246 & \rowcolor{gray!20} \first{0.152} & \first{0.245}\\
&  336 & 0.166 & 0.260 & \rowcolor{gray!20} \first{0.162} & \first{0.258}\\
&  720 & 0.194 & 0.287 & \rowcolor{gray!20} \first{0.188} & \first{0.261}\\
\cmidrule{2-6}
&  Avg.& 0.161 & 0.254 & \rowcolor{gray!20} \first{0.157} & \first{0.252}\\
\midrule
\multirow{5.5}{*}{Traffic}
&  96 &  0.350 & 0.250 & \rowcolor{gray!20} \first{0.338} & \first{0.246}\\
&  192 & 0.377 & \first{0.261} & \rowcolor{gray!20} \first{0.376} & \first{0.261}\\
&  336 & 0.377 & 0.261 & \rowcolor{gray!20} \first{0.361} & \first{0.257}\\
&  720 & \first{0.386} & 0.268 & \rowcolor{gray!20} 0.389 & \first{0.269}\\
\cmidrule{2-6}
&  Avg.& 0.373 & 0.260 & \rowcolor{gray!20} \first{0.366} & \first{0.259}\\
\midrule
\midrule
\multicolumn{2}{c}{Best Count (/52)} &\rowcolor{yellow!20} 6 & 8 & 46 & 47 \\
\bottomrule
\end{NiceTabular}
\end{adjustbox}
\captionsetup{type=table}
\end{minipage}
\captionsetup{type=table}
\caption{TS forecasting results with \textbf{Minusformer} with $L=336$.}
\label{tbl:minusformer_full336}
\end{figure*}

\clearpage

\section{Application to UniTS}
To demonstrate the effectiveness of our method on a TS foundation model, we apply it to four different TS tasks using UniTS~\cite{gao2024units} on datasets from various domains, under multiple settings, including multi-task, few-shot, and zero-shot settings.
All experimental settings follow those outlined in UniTS~\cite{gao2024units}.
The sections and tables outlining the full experiment results are listed in Table~\ref{tbl:outline}.

\begin{table*}[h]
\centering
\begin{adjustbox}{max width=0.999\textwidth}
\begin{NiceTabular}{c|c|c|c|c|c}
\toprule
\multirow{2.5}{*}{Settings} & \multirow{2.5}{*}{Section} & \multicolumn{4}{c}{TS downstream tasks}\\
\cmidrule(lr){3-6}
& & FCST &CLS & IMP & AD  \\
\midrule
Multi-task & \ref{sec:units1} & Table~\ref{tbl:exp1_FCST} & Table~\ref{tbl:exp1_CLS} & - & - \\
\midrule
Few-shot & \ref{sec:units2}  & Table~\ref{tbl:fewshot_FCST5},\ref{tbl:fewshot_FCST15},\ref{tbl:fewshot_FCST20} & Table~\ref{tbl:fewshot_CLS5},\ref{tbl:fewshot_CLS15},\ref{tbl:fewshot_CLS20} & Table~\ref{tbl:fewshot_imp} & Table~\ref{tbl:fewshot_ad} \\
\midrule
Zero-shot & \ref{sec:app_TSFM}  & Table~\ref{tbl:zero_data}, \ref{tbl:zero_horizon} & - & - & - \\
\bottomrule
\end{NiceTabular}
\end{adjustbox}
\caption{Summary of experiments.}
\label{tbl:outline}
\end{table*}
\vspace{20pt}

\subsection{Multi-task Learning}
\label{sec:units1}
For experiments under multi-task settings, we perform 20 TS forecasting and 18 classification tasks, where the full results are shown in Table~\ref{tbl:exp1_FCST} and Table~\ref{tbl:exp1_CLS}, respectively.

\newcolumntype{g}{>{\color{gray}}c}
\begin{table*}[h]
\vspace{5pt}
\centering
\begin{adjustbox}{max width=0.999\textwidth}
\begin{NiceTabular}{cc|cccc|cccc|cccccc|gg}
\toprule
\multicolumn{2}{c}{\multirow{4}{*}{20 Tasks}} & \multicolumn{8}{c}{Shared (1 model)} & \multicolumn{8}{c}{Task-specific (20 models)} \\
\cmidrule(lr){3-10} \cmidrule(lr){11-18}
&
& \multicolumn{4}{c}{$\text{UniTS + CM}$} 
& \multicolumn{4}{c}{$\text{UniTS}$} 
& \multicolumn{2}{c}{{iTransformer}} 
& \multicolumn{2}{c}{{TimesNet}} 
& \multicolumn{2}{c}{{PatchTST}} 
& \multicolumn{2}{c}{{GPT4TS}} \\
\cmidrule(lr){3-6} \cmidrule(lr){7-10} \cmidrule(lr){11-16} \cmidrule(lr){17-18} 
& 
& \multicolumn{2}{c}{Sup.} 
& \multicolumn{2}{c}{PT} 
& \multicolumn{2}{c}{Sup.} 
& \multicolumn{2}{c}{PT} 
& \multicolumn{6}{c}{Sup.} 
& \multicolumn{2}{c}{FT} 
\\
\cmidrule{1-2} \cmidrule(lr){3-4} \cmidrule(lr){5-6} \cmidrule(lr){7-8} \cmidrule(lr){9-10} \cmidrule(lr){11-16} \cmidrule(lr){17-18} 
Dataset & $H$ & 
MSE & MAE & 
MSE & MAE & 
MSE & MAE & 
MSE & MAE & 
MSE & MAE & 
MSE & MAE & 
MSE & MAE & 
MSE & MAE \\
\midrule
NN5 & 112 & {0.641} & {0.568} & \textcolor{red}{\textbf{0.586}} & \textcolor{red}{\textbf{0.536}} & {0.635} & {0.556}  & \textcolor{blue}{\underline{0.611}} & \textcolor{blue}{\underline{0.552}} & 0.623 & 0.554 & 0.629 & {0.541} & 0.634 & 0.568 & 0.623 & 0.545\\
\midrule
\multirow{4}{*}{ECL} & 96 & {0.176} & {0.278} & \textcolor{red}{\textbf{0.168}} & \textcolor{red}{\textbf{0.272}} & \textcolor{blue}{\underline{0.172}} & \textcolor{blue}{\underline{0.273}} & {0.174} & {0.277} & 0.204 & 0.288 & 0.184 & 0.289 & 0.212 & 0.299 & 0.198 & 0.285 \\
 & 192 & {0.188} & {0.287}& \textcolor{red}{\textbf{0.184}} & \textcolor{blue}{\underline{0.286}} & \textcolor{blue}{\underline{0.185}} & \textcolor{red}{\textbf{0.284}} & {0.189} & {0.289}  & 0.208 & 0.294 & 0.204 & 0.307 & 0.213 & 0.303 & 0.200 & 0.288 \\
 & 336 & \textcolor{blue}{\underline{0.199}} & \textcolor{red}{\textbf{0.295}}& \textcolor{blue}{\underline{0.199}} & {0.301} & \textcolor{red}{\textbf{0.196}} & \textcolor{blue}{\underline{0.297}} & {0.205} & {0.304} &  0.224 & 0.310 & 0.217 & 0.320 & 0.228 & 0.317 & 0.214 & 0.302\\
 & 720 & \textcolor{red}{\textbf{0.230}} & \textcolor{red}{\textbf{0.321}} & \textcolor{blue}{\underline{0.231}} & \textcolor{blue}{\underline{0.326}}& {0.238} & \textcolor{red}{\textbf{0.321}} & {0.251} & {0.340} &  0.265 & 0.341 & 0.284 & 0.363 & 0.270 & 0.348 & 0.254 & 0.333\\
 \midrule
\multirow{4}{*}{ETTh1} & 96 & \textcolor{blue}{\underline{0.388}} & \textcolor{blue}{\underline{0.405}} & {0.389} & {0.408}& {0.390} & {0.408} &  0.390 & 0.411 & \textcolor{red}{\textbf{0.382}} & \textcolor{red}{\textbf{0.399}} & 0.478 & 0.448 & 0.389 & {0.400} & 0.396 & 0.413\\
 & 192 & {0.438} & {0.436} & {0.432} & \textcolor{blue}{\underline{0.432}}& \textcolor{red}{\textbf{0.428}} & \textcolor{blue}{\underline{0.432}} & {0.432} & 0.439 & \textcolor{blue}{\underline{0.431}} & \textcolor{red}{\textbf{0.426}} & 0.561 & 0.504 & 0.440 & {0.43} & 0.458 & 0.448\\
 & 336 & {0.478} & {0.455} & \textcolor{blue}{\underline{0.475}} & \textcolor{blue}{\underline{0.451}}& \textcolor{red}{\textbf{0.462}} & \textcolor{blue}{\underline{0.451}} & {0.480} & 0.460 & {0.476} & \textcolor{red}{\textbf{0.449}} & 0.612 & 0.537 & 0.482 & {0.453} & 0.508 & 0.472\\
 & 720 & \textcolor{red}{\textbf{0.483}} & \textcolor{red}{\textbf{0.472}} & {0.515} & {0.492}& {0.489} & \textcolor{blue}{\underline{0.476}} & 0.532 & 0.500 & 0.495 & 0.487 & 0.601 & 0.541 & \textcolor{blue}{\underline{0.486}} & \textcolor{blue}{\underline{0.479}}  & 0.546 & 0.503\\
\midrule
\multirow{2}{*}{Exchange} & 192 & {0.231} & {0.340} & {0.210} & {0.330}& {0.239} & {0.342} & 0.221 & 0.337  & \textcolor{red}{\textbf{0.175}} & \textcolor{red}{\textbf{0.297}} & 0.259 & 0.370 & \textcolor{blue}{\underline{0.178}} & \textcolor{blue}{\underline{0.301}} & 0.177 & 0.300\\
 & 336 & {0.431} & {0.472} & {0.387} & {0.451}& {0.479} & {0.486} & 0.387 & 0.453 & \textcolor{red}{\textbf{0.322}} & \textcolor{red}{\textbf{0.409}} & 0.478 & 0.501 & \textcolor{blue}{\underline{0.328}} & \textcolor{blue}{\underline{0.415}} & 0.326 & 0.414\\
\midrule
ILI & 60 & \textcolor{blue}{\underline{2.02}} & \textcolor{red}{\textbf{0.885}} & {2.15} & {0.923}& {2.48} & {0.944} & 2.45 & 0.994 & \textcolor{red}{\textbf{1.99}} & \textcolor{blue}{\underline{0.905}} & 2.37 & 0.966 & 2.31 & 0.970 & 1.90 & 0.868\\
\midrule
\multirow{4}{*}{Traffic} & 96 & \textcolor{blue}{\underline{0.486}} & \textcolor{red}{\textbf{0.322}} & \textcolor{red}{\textbf{0.483}} & \textcolor{blue}{\underline{0.324}}& {0.496} & {0.325}  & {0.502} & {0.330} & 0.606 & 0.389 & 0.611 & 0.336 & 0.643 & 0.405 & 0.524 & 0.351\\
 & 192 & \textcolor{red}{\textbf{0.492}} & \textcolor{red}{\textbf{0.325}}& {0.500} & {0.330} & \textcolor{blue}{\underline{0.497}} & \textcolor{blue}{\underline{0.327}} & {0.523} & {0.331} &  0.592 & 0.382 & 0.643 & 0.352 & 0.603 & 0.387 & 0.519 & 0.346\\
 & 336 & \textcolor{red}{\textbf{0.506}} & \textcolor{blue}{\underline{0.331}}& {0.520} & {0.337} & \textcolor{blue}{\underline{0.509}} & \textcolor{red}{\textbf{0.328}} & {0.552} & {0.338} &  0.600 & 0.384 & 0.662 & 0.363 & 0.612 & 0.389 &  0.530 & 0.350\\
 & 720 & \textcolor{red}{\textbf{0.523}} & \textcolor{red}{\textbf{0.340}} & {0.575} & {0.362}& \textcolor{blue}{\underline{0.525}} & \textcolor{blue}{\underline{0.350}} & {0.626} & {0.369} &  0.633 & 0.401 & 0.678 & 0.365 & 0.652 & 0.406 & 0.562 & 0.366\\
\midrule 
\multirow{4}{*}{Weather} & 96 & \textcolor{blue}{\underline{0.165}} & \textcolor{red}{\textbf{0.211}} & {0.166} & {0.219} & \textcolor{red}{\textbf{0.161}} & \textcolor{red}{\textbf{0.211}} & {0.175} & \textcolor{blue}{\underline{0.214}}  & 0.193 & 0.232 & 0.169 & 0.220 & 0.194 & 0.233 & 0.182 & 0.222\\
 &192 & \textcolor{red}{\textbf{0.210}} & \textcolor{red}{\textbf{0.254}} & {0.216} & {0.261}& \textcolor{blue}{\underline{0.212}} & \textcolor{blue}{\underline{0.255}} & {0.226} & {0.266} &  0.238 & 0.269 & 0.223 & 0.264 & 0.238 & 0.268 & 0.228 & 0.261\\
 &336 & \textcolor{red}{\textbf{0.266}} & \textcolor{red}{\textbf{0.294}}& \textcolor{blue}{\underline{0.273}} & {0.300} & \textcolor{red}{\textbf{0.266}} & \textcolor{blue}{\underline{0.295}} & {0.280} & {0.303} &  0.291 & 0.306 & {0.279} & 0.302 & 0.290 & 0.304 & 0.282 & 0.299 \\
 & 720 & \textcolor{red}{\textbf{0.342}} & \textcolor{red}{\textbf{0.343}}& {0.350} & {0.349} & \textcolor{blue}{\underline{0.343}} & \textcolor{blue}{\underline{0.344}}  & {0.352} & {0.350} & 0.365 & 0.354 & 0.359 & 0.355 & 0.363 & {0.35} & 0.359 & 0.349 \\
\midrule
\multicolumn{2}{c}{\cellcolor{gray!20} Best Count (/20)} \cellcolor{gray!20} &  \rowcolor{gray!20} 8 & 11 & 4 & 2 & 5 & 4 & 0 & 0 & 4 & 5 & 0 & 0 & 0 & 0 & - & -\\
\midrule 
\multicolumn{2}{c}{\cellcolor{gray!20} Average} \cellcolor{gray!20} & \rowcolor{gray!20} \textcolor{red}{\textbf{0.445}} & \textcolor{red}{\textbf{0.382}} & \textcolor{blue}{\underline{0.452}} & \textcolor{blue}{\underline{0.384}} & {0.469} & 0.386 & {0.478} & {0.393} & 0.466 & 0.394 & 0.525 & 0.412 & 0.488 & 0.401 & 0.449 & 0.386\\
\bottomrule
\end{NiceTabular}
\end{adjustbox}
\caption{
Results of multi-task forecasting.
}
\label{tbl:exp1_FCST}
\vspace{-10pt}

\end{table*}

\clearpage
\begin{table*}[t]
\centering
\begin{adjustbox}{max width=0.999\textwidth}
\begin{NiceTabular}{l|cc|cc|ccccc|g}
\toprule
\multicolumn{1}{c}{\multirow{4}{*}{18 Tasks}} & \multicolumn{4}{c}{Shared (1 model)} & \multicolumn{6}{c}{Task-specific (18 models)} \\
\cmidrule(lr){2-5} \cmidrule(lr){6-11} 
& \multicolumn{2}{c}{UniTS + CM} 
& \multicolumn{2}{c}{UniTS} 
& \multicolumn{1}{c}{{iTransformer}} 
& \multicolumn{1}{c}{{TimesNet}} 
& \multicolumn{1}{c}{{PatchTST}} 
& \multicolumn{1}{c}{{Pyraformer}} 
& \multicolumn{1}{c}{{Autoformer}} 
& \multicolumn{1}{c}{{GPT4TS}} \\
\cmidrule(lr){2-3}
\cmidrule(lr){4-5}
\cmidrule(lr){6-10}
\cmidrule(lr){11-11}
& Sup. & PT
& Sup. & PT
& \multicolumn{5}{c}{Sup.} 
& \multicolumn{2}{c}{FT} 
\\
\midrule
Heartbeat & 67.3 & 70.2 & 59.0 & 69.3 & 66.8 & \textcolor{red}{\textbf{72.7}} & 65.9 & \textcolor{red}{\textbf{72.7}} & \textcolor{blue}{\underline{71.7}} & 69.8\\
JapaneseVowels & 94.1 & 93.2 & 93.5 & 90.8 & \textcolor{blue}{\underline{95.9}} & \textcolor{red}{\textbf{97.6}} & 94.1 & 85.4 & 94.1& 94.6\\
PEMS-SF & \textcolor{blue}{\underline{83.2}} & 82.1 & \textcolor{blue}{\underline{83.2}} & 85.0 & \textcolor{blue}{\underline{83.2}} & 77.5 & \textcolor{red}{\textbf{83.8}} & \textcolor{blue}{\underline{83.2}} & 79.2& 79.2\\
SelfRegulationSCP2 & \textcolor{red}{\textbf{58.3}} & 51.7 & 47.8 & 53.3 & 48.9 & 52.8 & 48.9 & \textcolor{blue}{\underline{56.7}} & 45.0& 45.6\\
SpokenArabicDigits & 97.1 & 93.5 & 97.5 & 92.0 & \textcolor{blue}{\underline{97.8}} & \textcolor{red}{\textbf{98.7}} & 97.5 & 92.1 & 97.3& 97.5\\
UWaveGestureLibrary & \textcolor{red}{\textbf{84.4}} & \textcolor{blue}{\underline{83.8}} & 79.1 & 75.6 & 82.2 & \textcolor{red}{\textbf{84.4}}  & 81.9 & 72.2 & 42.2& 81.9\\
ECG5000 & \textcolor{blue}{\underline{93.4}} & \textcolor{blue}{\underline{93.4}} & 92.6 & \textcolor{blue}{\underline{93.4}}& \underline{93.3} & 92.6 & \textcolor{red}{\textbf{94.3}} & 91.4 & 91.9& 93.0\\
NonInvasiveFetalECGThorax1 & \textcolor{blue}{\underline{89.5}} & 55.2 & \textcolor{red}{\textbf{90.5}} & 27.1 & 88.2 & \underline{88.9} & 86.5 & 21.4 & 21.7 & 89.7\\
Blink & \textcolor{red}{\textbf{99.1}} & \textcolor{blue}{\underline{95.6}} & \textcolor{red}{\textbf{99.1}} & 91.1 & \underline{93.3} & 87.6 & 89.6 & 88.2 & 63.1 & 92.4\\
FaceDetection & 64.7 & 54.6 & 64.1 & 57.6 & 66.0 & \textcolor{blue}{\underline{66.2}} & 63.9 & \textcolor{red}{\textbf{67.3}} & 59.2 & 66.1\\
ElectricDevices & \textcolor{blue}{\underline{62.4}} & 60.5 & 60.3 & 55.4 & 57.3 & 49.5 & 59.5 & \textcolor{red}{\textbf{65.4}}& 56.1 & 62.9\\
Trace & \textcolor{red}{\textbf{99.0}} & \textcolor{blue}{\underline{93.0}} & 91.0 & 82.0 & 79.0 & 91.0 & 77.0 & 74.0 & 60.0& 96.0\\
FordB & \textcolor{red}{\textbf{76.2}} & 64.2 & \textcolor{blue}{\underline{76.0}}& 62.8 & 72.7 & 68.9 & 61.4 & 55.3 & 66.4& 77.7\\
MotionSenseHAR & 92.8 & \textcolor{red}{\textbf{94.3}} & 92.8 & 93.2 & \textcolor{blue}{\underline{93.6}} & 90.6 & 75.8 & 88.7 & 30.2 &96.2\\
EMOPain & 75.5 & \textcolor{blue}{\underline{80.8}} & 78.0 & 80.3 & 79.4 & 78.0 & 79.2 & \textcolor{red}{\textbf{81.4}} & 69.9& 79.4\\
Chinatown & \textcolor{blue}{\underline{97.7}}  & \textcolor{red}{\textbf{98.0}} & \textcolor{blue}{\underline{97.7}}  &\textcolor{red}{\textbf{98.0}} & 97.4 & \textcolor{blue}{\underline{97.7}}  & \textcolor{blue}{\underline{97.7}} & 27.4 & 96.8& 96.5\\
MelbournePedestrian & \textcolor{blue}{\underline{89.3}} & 78.3 & 87.3 & 77.0 & \textcolor{blue}{\underline{89.3}} & \textcolor{red}{\textbf{95.7}} & 80.4 & 52.3 & 75.0& 94.0\\
SharePriceIncrease & 62.9 & 66.6 & 61.9 & \textcolor{red}{\textbf{68.4}}& 61.9 & 65.0 & \textcolor{blue}{\underline{68.0}} & 63.1 & 61.5& 63.7\\ \midrule
\midrule
 \rowcolor{gray!20} 1st Count (/18) & 5 & 2 & 2 & 2 & 0 & 5 & 2 & 4 & 0 & -  \\
\rowcolor{gray!20}  2nd Count (/18) & 6 & 5 & 3 & 1 & 5 & 2 & 2 & 2 & 1 & -   \\
\midrule
\rowcolor{gray!20}  Average Score & \textcolor{red}{\textbf{82.0}} & 78.3 & 80.6 & 75.1 & 80.3 & \textcolor{blue}{\underline{80.9}}& 78.1 & 68.8 & 65.6 & 82.0\\
\bottomrule
\end{NiceTabular}
\end{adjustbox}
\caption{Results of multi-task classification.}
\label{tbl:exp1_CLS}
\end{table*}

\clearpage

\subsection{Few-shot Learning}
\label{sec:units2}
For the few-shot tasks, we conduct four distinct tasks: forecasting (FCST), classification (CLS), imputation (IMP), and anomaly detection (AD), which are discussed in Sections~\ref{sec:fewshot_FCST}, \ref{sec:fewshot_CLS}, \ref{sec:fewshot_imp}, and \ref{sec:fewshot_ad}, respectively.

\subsubsection{Few-shot Forecasting}
\label{sec:fewshot_FCST}

The results of few-shot forecasting with data ratios of 5\%, 15\%, and 20\% are shown in Tables~\ref{tbl:fewshot_FCST5}, \ref{tbl:fewshot_FCST15}, and \ref{tbl:fewshot_FCST20}, respectively.

\begin{table*}[h]
\vspace{10pt}
\centering
\begin{adjustbox}{max width=0.9\textwidth}
\begin{NiceTabular}{cc|cccccc|cccc}
\toprule 
\multicolumn{2}{c}{\multirow{2.5}{*}{5\%}}
& \multicolumn{2}{c}{iTransformer}
& \multicolumn{4}{c}{UniTS}
& \multicolumn{4}{c}{UniTS + CM}
\\
\cmidrule(lr){3-4} \cmidrule(lr){5-8} \cmidrule(lr){9-12} 
\multicolumn{2}{c}{} & \multicolumn{2}{c}{FT} & \multicolumn{2}{c}{PT} & \multicolumn{2}{c}{FT} & \multicolumn{2}{c}{PT} & \multicolumn{2}{c}{FT}  \\
\cmidrule(lr){1-2}  \cmidrule(lr){3-4} \cmidrule(lr){5-6} \cmidrule(lr){7-8} \cmidrule(lr){9-10} \cmidrule(lr){11-12}
Data & $H$ & 
MSE & MAE & 
MSE & MAE & 
MSE & MAE & 
MSE & MAE & 
MSE & MAE \\
\midrule  
\multirow{4}{*}{ETTh2} & 96 & 0.554 & 0.500 &  \textcolor{red}{\textbf{0.405}} & \textcolor{red}{\textbf{0.417}} & \textcolor{blue}{\underline{0.418}} & \textcolor{blue}{\underline{0.424}} & \cellcolor{gray!20} 0.421 & \cellcolor{gray!20} 0.427 & \cellcolor{gray!20} 0.421 & \cellcolor{gray!20} 0.425 \\
 & 192 & 0.440 & 0.438 & 0.400 & 0.406 & 0.377 & 0.397 & \cellcolor{gray!20} \textcolor{blue}{\underline{0.386}} & \cellcolor{gray!20} \textcolor{blue}{\underline{0.402}} &  \cellcolor{gray!20} \textcolor{red}{\textbf{0.370}} & \cellcolor{gray!20} \textcolor{red}{\textbf{0.389}}  \\
 & 336 & 0.478 & 0.467 & 0.425 & 0.433 & \textcolor{blue}{\underline{0.420}} & 0.433 & \cellcolor{gray!20} 0.423 & \cellcolor{gray!20} \textcolor{blue}{\underline{0.431}} & \cellcolor{gray!20} \textcolor{red}{\textbf{0.416}} & \cellcolor{gray!20} \textcolor{red}{\textbf{0.425}}\\
 & 720 & 0.483 & 0.480 & 0.446 & 0.457 & 0.439 & 0.452 & \cellcolor{gray!20} \textcolor{red}{\textbf{0.424}} & \cellcolor{gray!20} \textcolor{blue}{\underline{0.444}} & \cellcolor{gray!20} \textcolor{blue}{\underline{0.428}} & \cellcolor{gray!20} \textcolor{red}{\textbf{0.443}}\\
 \midrule
RiverFlow & 24 & 1.141 & 0.514 &  1.115 &  0.504 &  \second{1.112} &  0.504 & \cellcolor{gray!20} \textcolor{red}{\textbf{1.097}} & \cellcolor{gray!20} \textcolor{blue}{\underline{0.503}} & \cellcolor{gray!20} \textcolor{red}{\textbf{1.097}} & \cellcolor{gray!20} \textcolor{red}{\textbf{0.500}} \\
\midrule
\multirow{4}{*}{ETTm1} & 96 & 0.504 & 0.462 & 0.436 & 0.434 & \textcolor{blue}{\underline{0.384}} & \textcolor{blue}{\underline{0.404}} & \cellcolor{gray!20} 0.428 & \cellcolor{gray!20} 0.436 & \cellcolor{gray!20} \textcolor{red}{\textbf{0.354}} & \cellcolor{gray!20} \textcolor{red}{\textbf{0.384}} \\
 & 192 & 0.555 & 0.485 & 0.462 & 0.448 & \textcolor{blue}{\underline{0.414}} & \textcolor{blue}{\underline{0.418}} & \cellcolor{gray!20} 0.475 & \cellcolor{gray!20} 0.458 & \cellcolor{gray!20} \textcolor{red}{\textbf{0.393}} & \cellcolor{gray!20} \textcolor{red}{\textbf{0.405}}\\
 & 336 & 0.567 & 0.496 & 0.560 & 0.494 & \textcolor{blue}{\underline{0.453}}& \textcolor{blue}{\underline{0.442}} &\cellcolor{gray!20} 0.550 & \cellcolor{gray!20} 0.493 & \cellcolor{gray!20} \textcolor{red}{\textbf{0.420}} & \cellcolor{gray!20} \textcolor{red}{\textbf{0.423}} \\
 & 720 & 0.659 & 0.539 & 0.703 & 0.558 & \textcolor{blue}{\underline{0.526}} & \textcolor{blue}{\underline{0.483}} & \cellcolor{gray!20} 0.689 & \cellcolor{gray!20} 0.554 & \cellcolor{gray!20} \textcolor{red}{\textbf{0.483}} & \cellcolor{gray!20} \textcolor{red}{\textbf{0.455}}\\
\midrule
\multicolumn{2}{c}{Average}  & 0.598 & 0.487 & 0.549 & 0.461 & \textcolor{blue}{\underline{0.505}} & \textcolor{blue}{\underline{0.440}} & \cellcolor{gray!20} 0.546 & \cellcolor{gray!20} 0.462 & \cellcolor{gray!20} \textcolor{red}{\textbf{0.489}}& \cellcolor{gray!20} \textcolor{red}{\textbf{0.429}} \\
\bottomrule
\end{NiceTabular}
\end{adjustbox}
\caption{Results of few-shot forecasting (5\%).}
\label{tbl:fewshot_FCST5}
\end{table*}

\begin{table*}[h]
\vspace{15pt}
\centering
\begin{adjustbox}{max width=0.9\textwidth}
\begin{NiceTabular}{cc|cccccc|cccc}
\toprule 
\multicolumn{2}{c}{\multirow{2.5}{*}{15\%}}
& \multicolumn{2}{c}{iTransformer}
& \multicolumn{4}{c}{UniTS}
& \multicolumn{4}{c}{UniTS + CM}
\\
\cmidrule(lr){3-4} \cmidrule(lr){5-8} \cmidrule(lr){9-12} 
\multicolumn{2}{c}{} & \multicolumn{2}{c}{FT} & \multicolumn{2}{c}{PT} & \multicolumn{2}{c}{FT} & \multicolumn{2}{c}{PT} & \multicolumn{2}{c}{FT}  \\
\cmidrule(lr){1-2}  \cmidrule(lr){3-4} \cmidrule(lr){5-6} \cmidrule(lr){7-8} \cmidrule(lr){9-10} \cmidrule(lr){11-12}
Data & $H$ & 
MSE & MAE & 
MSE & MAE & 
MSE & MAE & 
MSE & MAE & 
MSE & MAE \\
\midrule 
\multirow{4}{*}{ETTh2} & 96 & 0.441 & 0.440 & 0.403 & 0.412 &  \textcolor{red}{\textbf{0.399}} & \textcolor{red}{\textbf{0.409}} & \cellcolor{gray!20} 0.416 & \cellcolor{gray!20}0.423 & \cellcolor{gray!20} \textcolor{blue}{\underline{0.403}} & \cellcolor{gray!20} \textcolor{blue}{\underline{0.411}}\\
 & 192 & 0.398 & 0.410 & 0.396 & 0.404 & 0.394 & \textcolor{blue}{\underline{0.399}} & \cellcolor{gray!20} \textcolor{blue}{\underline{0.388}} & \cellcolor{gray!20} 0.403 &  \cellcolor{gray!20} \textcolor{red}{\textbf{0.387}} & \cellcolor{gray!20} \textcolor{red}{\textbf{0.399}}
 \\
 & 336 & 0.436 & 0.441 & 0.432 & 0.435 & 0.441 & 0.435 &  \cellcolor{gray!20} \textcolor{red}{\textbf{0.419}} & \cellcolor{gray!20} \textcolor{blue}{\underline{0.435}} & \cellcolor{gray!20} \textcolor{blue}{\underline{0.430}} & \cellcolor{gray!20} \textcolor{red}{\textbf{0.431}} \\
 & 720 & 0.438 & 0.453 & 0.448 & 0.457 & 0.449 & 0.453 & \cellcolor{gray!20} \textcolor{red}{\textbf{0.415}} & \cellcolor{gray!20} \textcolor{red}{\textbf{0.442}} & \cellcolor{gray!20} \textcolor{blue}{\underline{0.433}} & \cellcolor{gray!20} \textcolor{blue}{\underline{0.446}}\\
 \midrule
RiverFlow & 24 & \textcolor{red}{\textbf{1.067}} & \textcolor{red}{\textbf{0.467}} & 1.077 & 0.492 & \textcolor{blue}{\underline{1.069}} & 0.489 & \cellcolor{gray!20} 1.073 & \cellcolor{gray!20} 0.492 & \cellcolor{gray!20} 1.072 & \cellcolor{gray!20} \textcolor{blue}{\underline{0.487}} \\
\midrule
\multirow{4}{*}{ETTm1} & 96 & 0.423 & 0.419 & 0.407 & 0.420 & \textcolor{blue}{\underline{0.353}} & \textcolor{blue}{\underline{0.386}} & \cellcolor{gray!20} 0.408 & \cellcolor{gray!20} 0.426 & \cellcolor{gray!20} \textcolor{red}{\textbf{0.342}} & \cellcolor{gray!20} \textcolor{red}{\textbf{0.380}}  \\
 & 192 & 0.464 & 0.439 & 0.434 & 0.432 & \textcolor{blue}{\underline{0.384}} & \textcolor{blue}{\underline{0.400}} & \cellcolor{gray!20} 0.449 & \cellcolor{gray!20} 0.447 & \cellcolor{gray!20} \textcolor{red}{\textbf{0.377}} & \cellcolor{gray!20} \textcolor{red}{\textbf{0.399}} \\
 & 336 &  0.492 & 0.457 & 0.490 & 0.464 & \textcolor{blue}{\underline{0.416}} & \textcolor{blue}{\underline{0.420}} & \cellcolor{gray!20} 0.502 & \cellcolor{gray!20} 0.475 & \cellcolor{gray!20} \textcolor{red}{\textbf{0.406}} & \cellcolor{gray!20} \textcolor{red}{\textbf{0.148}} \\
 & 720 & 0.558 & 0.493 & 0.641 & 0.537 & \textcolor{blue}{\underline{0.480}} & \textcolor{blue}{\underline{0.455}} & \cellcolor{gray!20} 0.621 & \cellcolor{gray!20} 0.530 & \cellcolor{gray!20} \textcolor{red}{\textbf{0.470}} & \cellcolor{gray!20} \textcolor{red}{\textbf{0.451}} \\
\midrule
\multicolumn{2}{c}{Average}  &0.524 & 0.450 & 0.525 & 0.450 & \textcolor{blue}{\underline{0.487}} & \textcolor{blue}{\underline{0.428}} & \cellcolor{gray!20} 0.522 & \cellcolor{gray!20} 0.452 & \cellcolor{gray!20} \textcolor{red}{\textbf{0.481}} & \cellcolor{gray!20} \textcolor{red}{\textbf{0.425}} \\
\bottomrule
\end{NiceTabular}
\end{adjustbox}
\caption{Results of few-shot forecasting (15\%).}
\label{tbl:fewshot_FCST15}
\end{table*}

\clearpage

\begin{table*}[t]
\centering
\begin{adjustbox}{max width=0.9\textwidth}
\begin{NiceTabular}{cc|cccccc|cccc}
\toprule 
\multicolumn{2}{c}{\multirow{2.5}{*}{20\%}}
& \multicolumn{2}{c}{iTransformer}
& \multicolumn{4}{c}{UniTS}
& \multicolumn{4}{c}{UniTS + CM}
\\
\cmidrule(lr){3-4} \cmidrule(lr){5-8} \cmidrule(lr){9-12} 
\multicolumn{2}{c}{} & \multicolumn{2}{c}{FT} & \multicolumn{2}{c}{PT} & \multicolumn{2}{c}{FT} & \multicolumn{2}{c}{PT} & \multicolumn{2}{c}{FT}  \\
\cmidrule(lr){1-2}  \cmidrule(lr){3-4} \cmidrule(lr){5-6} \cmidrule(lr){7-8} \cmidrule(lr){9-10} \cmidrule(lr){11-12}
Data & $H$ & 
MSE & MAE & 
MSE & MAE & 
MSE & MAE & 
MSE & MAE & 
MSE & MAE \\
\midrule 
\multirow{4}{*}{ETTh2} & 96 & 0.418 & 0.426 & 0.411 & 0.414 &  \textcolor{red}{\textbf{0.391}} & \textcolor{red}{\textbf{0.405}}  & \cellcolor{gray!20} 0.411 & \cellcolor{gray!20} 0.422 & \cellcolor{gray!20} \textcolor{blue}{\underline{0.395}} & \cellcolor{gray!20} \textcolor{blue}{\underline{0.409}}\\
 & 192 & 0.395 & 0.407 & 0.383 & \textcolor{red}{\textbf{0.398}} & 0.395 & 0.403 &  \cellcolor{gray!20} \textcolor{red}{\textbf{0.381}} & \cellcolor{gray!20} \second{0.400}& \cellcolor{gray!20} \textcolor{blue}{\underline{0.390}} & \cellcolor{gray!20} \textcolor{blue}{\underline{0.400}} \\
 & 336 & 0.431 & 0.438 &  \textcolor{red}{\textbf{0.419}} & \textcolor{blue}{\underline{0.431}} & 0.430 & \first{0.430} & \cellcolor{gray!20} \textcolor{blue}{\underline{0.423}} & \cellcolor{gray!20} \textcolor{red}{\textbf{0.430}} & \cellcolor{gray!20} 0.438 & \cellcolor{gray!20} 0.433 \\
 & 720 & \textcolor{blue}{\underline{0.431}} & \textcolor{blue}{\underline{0.449}} & 0.440 & 0.453 & 0.444 & 0.449 &  \cellcolor{gray!20} \textcolor{red}{\textbf{0.418}} & \cellcolor{gray!20} \textcolor{red}{\textbf{0.422}}& \cellcolor{gray!20} 0.456 & \cellcolor{gray!20} 0.456 \\
 \midrule
RiverFlow & 24 & \textcolor{red}{\textbf{1.056}} & \textcolor{red}{\textbf{0.462}} & 1.069 & \textcolor{blue}{\underline{0.487}}  & 1.069 & 0.489 & \cellcolor{gray!20} 1.071 & \cellcolor{gray!20} 0.487 & \cellcolor{gray!20} \textcolor{blue}{\underline{1.067}} & \cellcolor{gray!20} 0.489\\
\midrule
\multirow{4}{*}{ETTm1} & 96 & 0.408 & 0.410 & 0.409 & 0.421 & \textcolor{blue}{\underline{0.344}} & \textcolor{blue}{\underline{0.379}} & \cellcolor{gray!20} 0.403 & \cellcolor{gray!20} 0.425 & \cellcolor{gray!20} \textcolor{red}{\textbf{0.339}} & \cellcolor{gray!20} \textcolor{red}{\textbf{0.376}} \\
 & 192 & 0.444 & 0.428 & 0.443 & 0.439 & \textcolor{blue}{\underline{0.377}}& \textcolor{blue}{\underline{0.397}} & \cellcolor{gray!20} 0.450& \cellcolor{gray!20} 0.450 & \cellcolor{gray!20} \textcolor{red}{\textbf{0.375}} & \cellcolor{gray!20} \textcolor{red}{\textbf{0.396}} \\
 & 336 & 0.471 & 0.445 & 0.505 & 0.472 & \textcolor{blue}{\underline{0.408}} & \textcolor{blue}{\underline{0.418}}& \cellcolor{gray!20} 0.507 & \cellcolor{gray!20} 0.481 & \cellcolor{gray!20} \textcolor{red}{\textbf{0.403}} & \cellcolor{gray!20} \textcolor{red}{\textbf{0.415}}\\
 & 720 & 0.536 & 0.482 & 0.648 & 0.536 & \textcolor{blue}{\underline{0.472}}& \textcolor{blue}{\underline{0.453}} & \cellcolor{gray!20} 0.621 & \cellcolor{gray!20} 0.531 & \cellcolor{gray!20} \textcolor{red}{\textbf{0.466}} & \cellcolor{gray!20} \textcolor{red}{\textbf{0.448}}\\
\midrule
\multicolumn{2}{c}{Average}  & 0.510 & 0.438 & 0.525 & 0.450 & \textcolor{blue}{\underline{0.486}} & \textcolor{blue}{\underline{0.425}} & \cellcolor{gray!20} 0.521 & \cellcolor{gray!20} 0.453 & \cellcolor{gray!20} \textcolor{red}{\textbf{0.482}}& \cellcolor{gray!20} \textcolor{red}{\textbf{0.425}} \\
\bottomrule
\end{NiceTabular}
\end{adjustbox}
\caption{Results of few-shot forecasting (20\%).}
\label{tbl:fewshot_FCST20}
\end{table*}

\clearpage
\subsubsection{Few-shot Classification}
\label{sec:fewshot_CLS}

The results of few-shot classification with data ratios of 5\%, 15\%, and 20\% are shown in Tables~\ref{tbl:fewshot_CLS5}, \ref{tbl:fewshot_CLS15}, and \ref{tbl:fewshot_CLS20}, respectively.

\begin{table*}[h]
\vspace{15pt}
\centering
\begin{adjustbox}{max width=0.999\textwidth}
\begin{NiceTabular}{c|ccc|cc}
\toprule 
\cmidrule(lr){2-6}
 \multirow{2.5}{*}{5\%} & iTransformer
  & \multicolumn{2}{c}{UniTS}
  & \multicolumn{2}{c}{UniTS + CM}
  
  \\
  \cmidrule(lr){2-2} \cmidrule(lr){3-4} \cmidrule(lr){5-6} 
& 
FT & PT & FT & PT & FT \\
\midrule
ECG200 & \textcolor{blue}{\underline{78.0}} & 67.0 & 77.0 & \cellcolor{gray!20} \textcolor{red}{\textbf{80.0}} & \cellcolor{gray!20} 77.0\\
Handwriting & \textcolor{blue}{\underline{5.4}} & 4.6 & 4.7 & \cellcolor{gray!20} 4.8 & \cellcolor{gray!20} \textcolor{red}{\textbf{5.5}} \\
SelfRegulationSCP1 & 62.8 & 66.2 & \textcolor{blue}{\underline{74.7}} & \cellcolor{gray!20} \textcolor{red}{\textbf{77.8}} & \cellcolor{gray!20} 73.7\\ 
RacketSports & 37.5 & 31.6 & 35.5 & \cellcolor{gray!20} \textcolor{blue}{\underline{39.5}} & \cellcolor{gray!20} \textcolor{red}{\textbf{47.4}} \\
Epilepsy & 39.9 & 44.9 & \textcolor{blue}{\underline{47.1}} & \cellcolor{gray!20} 44.9 & \cellcolor{gray!20} \textcolor{red}{\textbf{57.2}} \\
StarLightCurves & 85.1 & 82.3 & 83.8 & \cellcolor{gray!20} \textcolor{red}{\textbf{86.3}} & \cellcolor{gray!20} \textcolor{blue}{\underline{85.4}} \\
\midrule
Average & 51.4 & 49.4 & 53.8 & \cellcolor{gray!20} \textcolor{red}{\textbf{54.9}} & \cellcolor{gray!20} \textcolor{blue}{\underline{54.8}}\\
\bottomrule
\end{NiceTabular}
\end{adjustbox}
\caption{Results of few-shot classification (5\%).}
\label{tbl:fewshot_CLS5}
\end{table*}

\begin{table*}[h]
\vspace{15pt}
\centering
\begin{adjustbox}{max width=0.999\textwidth}
\begin{NiceTabular}{c|ccc|cc}
\toprule 
\cmidrule(lr){2-6}
\multirow{2.5}{*}{15\%}& iTransformer
  & \multicolumn{2}{c}{UniTS}
  & \multicolumn{2}{c}{UniTS + CM}
  
  \\
  \cmidrule(lr){2-2} \cmidrule(lr){3-4} \cmidrule(lr){5-6} 
& 
FT & PT & FT & PT & FT \\
\midrule
ECG200 & \textcolor{blue}{\underline{81.0}} & 74.0 & 78.0 &  \cellcolor{gray!20} 73.2 &  \cellcolor{gray!20} \textcolor{red}{\textbf{82.0}} \\
Handwriting & \textcolor{red}{\textbf{9.8}} & 7.3 & 8.1 & \cellcolor{gray!20} \textcolor{blue}{\underline{9.2}} & \cellcolor{gray!20} 8.5 \\
SelfRegulationSCP1 & 67.9 & 59.0 & \textcolor{red}{\textbf{76.5}} & \cellcolor{gray!20} \textcolor{blue}{\underline{69.3}} & \cellcolor{gray!20} 68.6  \\ 
RacketSports & \textcolor{red}{\textbf{54.6}} & 40.1 & 50.7 & \cellcolor{gray!20} 44.7 & \cellcolor{gray!20} \textcolor{blue}{\underline{51.3}}  \\
Epilepsy & 41.3 & 52.9 & 58.0 & \cellcolor{gray!20} \textcolor{blue}{\underline{61.6}} & \cellcolor{gray!20} \textcolor{red}{\textbf{68.1}} \\
StarLightCurves  & 84.2 & 85.8 & \textcolor{red}{\textbf{87.1}} & \cellcolor{gray!20} \textcolor{blue}{\underline{85.9}} & \cellcolor{gray!20} 85.5 \\
\midrule
Average  & 56.5 & 53.2 & \textcolor{blue}{\underline{59.7}} & \cellcolor{gray!20} 55.4 & \cellcolor{gray!20} \textcolor{red}{\textbf{60.4}}  \\
\bottomrule
\end{NiceTabular}
\end{adjustbox}
\caption{Results of few-shot classification (15\%).}
\label{tbl:fewshot_CLS15}
\end{table*}

\begin{table*}[h]
\vspace{15pt}
\centering
\begin{adjustbox}{max width=0.999\textwidth}
\begin{NiceTabular}{c|ccc|cc}
\toprule 
\cmidrule(lr){2-6}
\multirow{2.5}{*}{20\%}& iTransformer
  & \multicolumn{2}{c}{UniTS}
  & \multicolumn{2}{c}{UniTS + CM}
  
  \\
  \cmidrule(lr){2-2} \cmidrule(lr){3-4} \cmidrule(lr){5-6} 
& 
FT & PT & FT & PT & FT \\
\midrule
ECG200 & 81.0 & 76.0 & 77.0 & \cellcolor{gray!20} \textcolor{red}{\textbf{85.0}} & \cellcolor{gray!20} \textcolor{blue}{\underline{82.0}} \\
Handwriting & \textcolor{red}{\textbf{11.8}} & 8.0 & 8.5 & \cellcolor{gray!20} 7.6 & \cellcolor{gray!20} \textcolor{blue}{\underline{9.8}} \\
SelfRegulationSCP1 &  \textcolor{blue}{\underline{77.1}} & 68.6 & 70.6 & \cellcolor{gray!20} \textcolor{red}{\textbf{77.8}} & \cellcolor{gray!20} 74.4 \\ 
RacketSports & \textcolor{blue}{\underline{54.6}} & 51.3 & \textcolor{red}{\textbf{57.9}} & \cellcolor{gray!20} 38.8 & \cellcolor{gray!20} 50.7 \\
Epilepsy &  62.3 & \textcolor{blue}{\underline{81.9}} & 72.5 & \cellcolor{gray!20} \textcolor{red}{\textbf{84.1}} & \cellcolor{gray!20} 61.6 \\
StarLightCurves  & 84.8 & 87.3 & 86.0 & \cellcolor{gray!20} \textcolor{red}{\textbf{90.0}} & \cellcolor{gray!20} \textcolor{blue}{\underline{87.8}} \\
\midrule
Average  & 59.9 & 58.9 & \textcolor{blue}{\underline{63.6}} & \cellcolor{gray!20} 60.0 & \cellcolor{gray!20} \textcolor{red}{\textbf{64.8}} \\
\bottomrule
\end{NiceTabular}
\end{adjustbox}
\caption{Results of few-shot classification (20\%).}
\label{tbl:fewshot_CLS20}
\vspace{-40pt}
\end{table*}

\clearpage
\subsubsection{Few-shot Imputation}
\label{sec:fewshot_imp}

The results of few-shot imputation with data ratios of 25\% and 50\% are shown in Table~\ref{tbl:fewshot_imp}

\begin{table*}[h]
\vspace{15pt}
\centering
\begin{adjustbox}{max width=0.999\textwidth}
\begin{NiceTabular}{c|cc|cccccc|c}
\toprule
Ratio  &  \multicolumn{2}{c}{} &  ECL & ETTh1  & ETTh2  & ETTm1  & ETTm2  & Weather & Avg. \\
\midrule
\multirow{9}{*}{25\%} & TimesNet & \multirow{3}{*}{FT} &  0.245 & 0.369 & 0.193 & 0.442 & 0.119 & 0.106 & 0.246\\
& PatchTST &   &  0.195 & 0.315 & 0.147 & 0.309 & \textcolor{blue}{\underline{0.092}} & 0.089 & 0.191\\
& iTransformer &   &  0.174 & 0.301 & 0.185 & 0.254 & 0.113 & 0.087 & 0.186 \\
\cmidrule{2-10}
& \multirow{2}{*}{UniTS} & PT  &  \textcolor{blue}{\underline{0.139}} & 0.311 &  0.178 &  0.268 &  0.102 &  0.078 &  0.179 \\
&   & FT  &  0.160 & \textcolor{blue}{\underline{0.284}} &  \textcolor{blue}{\underline{0.150}} &  \textcolor{blue}{\underline{0.241}} &  \textcolor{red}{\textbf{0.090}} &  \textcolor{blue}{\underline{0.077}} &  \textcolor{blue}{\underline{0.167}}  \\
\cmidrule{2-10}
& \multirow{2}{*}{UniTS + CM}  & PT  &  \rowcolor{gray!20} \textcolor{blue}{\underline{0.139}} & 0.310  &  0.176 &  0.262 &  0.100 &  0.078 &  0.179  \\
&   & FT  &  \rowcolor{gray!20} \textcolor{red}{\textbf{0.129}} & \textcolor{red}{\textbf{0.275}} &  \textcolor{red}{\textbf{0.149}} &  \textcolor{red}{\textbf{0.231}} &  \textcolor{red}{\textbf{0.090}} &  \textcolor{red}{\textbf{0.073}} &  \textcolor{red}{\textbf{0.158}} \\
\midrule
\multirow{9}{*}{50\%} & TimesNet &  \multirow{3}{*}{FT} &  0.258 & 0.412 & 0.211 & 0.607 & 0.140 & 0.125 & 0.292 \\
 & PatchTST &  &  0.230 & 0.353 & 0.175 & 0.442 & \textcolor{red}{\textbf{0.111}} & 0.105 & 0.236\\
 & iTransformer &  &  0.203 & 0.332 & 0.205 & 0.372 & 0.136 & 0.106 & 0.226\\
 \cmidrule{2-10}
 & \multirow{2}{*}{UniTS}  & PT  &  0.172 & 0.352 &  0.251 &  0.380 &  0.134 &  0.103 &  0.232 \\
 &   & FT  &  0.191 & \textcolor{blue}{\underline{0.322}} &  \textcolor{blue}{\underline{0.198}} &  \textcolor{blue}{\underline{0.352}} &  0.118 &  \textcolor{blue}{\underline{0.095}} &  \textcolor{blue}{\underline{0.213}}  \\
 \cmidrule{2-10}
 & \multirow{2}{*}{UniTS + CM}  & PT  & \rowcolor{gray!20}  \textcolor{blue}{\underline{0.162}} & 0.353 &  0.240 &  0.370 &  0.128 &  0.097 &  0.225\\
 &   & FT  &  \rowcolor{gray!20} \textcolor{red}{\textbf{0.151}} & \textcolor{red}{\textbf{0.307}} &  \textcolor{red}{\textbf{0.197}} &  \textcolor{red}{\textbf{0.345}} &  \textcolor{blue}{\underline{0.116}} &  \textcolor{red}{\textbf{0.093}} &  \textcolor{red}{\textbf{0.201}}  \\
\bottomrule
\end{NiceTabular}
\end{adjustbox}
\caption{Results of few-shot imputation.}
\label{tbl:fewshot_imp}
\end{table*}

\vspace{30pt}
\subsubsection{Few-shot Anomaly Detection}
\label{sec:fewshot_ad}

The results of few-shot anomaly detection with data ratio of 5\% are shown in Table~\ref{tbl:fewshot_ad}.

\begin{table*}[h]
\vspace{15pt}
\centering
\begin{adjustbox}{max width=0.999\textwidth}
\begin{NiceTabular}{cc|ccccc|c}
\toprule
 &   & \multicolumn{1}{c}{MSL}& \multicolumn{1}{c}{PSM} & \multicolumn{1}{c}{SMAP}   & \multicolumn{1}{c}{SMD} & \multicolumn{1}{c}{SWAT} & \multicolumn{1}{c}{Avg.}\\
\midrule
Anomaly Trans. & -  & 78.0 & 90.2 & 68.3 & 77.8 & 81.5 & 79.2  \\
TimesNet & FT  & 33.9 & 91.0 & 68.5 & 84.0 & \textcolor{red}{\textbf{93.4}} & 74.2 \\
iTransfomer & FT   & \textcolor{blue}{\underline{80.4}}  & 96.5 & 67.2 & 82.4 & 89.0 &  83.1 \\
PatchTST & FT  & 79.9 &  \textcolor{blue}{\underline{96.6}} & 68.7 & 83.8 & 92.6  & 84.3 \\
\midrule
\multirow{2}{*}{UniTS}  & PT  &  73.2 & 95.5 &  65.9 &  81.2 &   \textcolor{blue}{\underline{92.9}} &  81.7  \\
 & FT  &  \textcolor{red}{\textbf{81.3}}& \textcolor{red}{\textbf{97.3}} &   \textcolor{blue}{\underline{71.6}}&   \textcolor{blue}{\underline{85.5}} &  92.5 &   \textcolor{blue}{\underline{85.6}} \\
 \midrule
\multirow{2}{*}{UniTS + CM} & PT  &  \rowcolor{gray!20} 73.7 & 95.5 &  66.0 &  82.0 &   \textcolor{blue}{\underline{92.9}} &  82.0\\
 & FT  &  \rowcolor{gray!20} \textcolor{red}{\textbf{81.3}} & \textcolor{red}{\textbf{97.3}} &  \textcolor{red}{\textbf{75.9}} &  \textcolor{red}{\textbf{86.2}} &  92.6 &  \textcolor{red}{\textbf{86.6}} \\
\bottomrule
\end{NiceTabular}
\end{adjustbox}
\caption{Results of few-shot anomaly detection.}
\label{tbl:fewshot_ad}
\end{table*}

\clearpage
\section{Application to TimeSiam}
\label{sec:timesiam}

To demonstrate the effectiveness of our proposed model on TimeSiam~\cite{dong2024timesiam}, 
which uses a self-supervised pretraining framework for TS with Siamese networks, we conduct experiments using iTransformer~\cite{liu2023itransformer} as the backbone, with two datasets that vary in channel size: Exchange, with a small number of channels (8), 
and ECL, with a large number of channels (321). 
Specifically, we apply variants of our method by using the domain parameter only during the fine-tuning stage and during both pretraining and fine-tuning stages.
The results, shown in Table~\ref{tbl:timesiam}, validate both components of our method, with the best performance achieved when using domain parameters at both pretraining and fine-tuning stages.

\begin{table*}[h]
\centering
\vspace{10pt}
\begin{adjustbox}{max width=1.00\textwidth}
\begin{NiceTabular}{c|c|cc|cc|cc|cc}
\toprule
\multicolumn{2}{c}{ } & \multicolumn{2}{c}{TimeSiam} & \multicolumn{6}{c}{+ CM} \\
\midrule
 \multicolumn{2}{c}{Correlation matrix} & \multicolumn{2}{c}{-} & \multicolumn{2}{c}{\cmark}& \multicolumn{2}{c}{\cmark}& \multicolumn{2}{c}{\cmark} \\
\midrule
\multirow{2}{*}{Domain parameters} & Pretrain & \multicolumn{2}{c}{-} & \multicolumn{2}{c}{-}& \multicolumn{2}{c}{-}& \multicolumn{2}{c}{\cmark} \\
 & Fine-tune & \multicolumn{2}{c}{-} & \multicolumn{2}{c}{-}& \multicolumn{2}{c}{\cmark}& \multicolumn{2}{c}{\cmark} \\
\midrule
\midrule
 Dataset & $H$  & MSE & MAE & MSE & MAE & MSE & MAE & MSE & MAE \\
 \midrule
 \multirow{5}{*}{\shortstack{\\\\Exchange\\($C=8$)}} & 96& 0.092 & 0.215 & 
\second{0.089} & \first{0.207} & 
\first{0.088} & \first{0.207} & \first{0.088} & \second{0.209} \\

  & 192 & \first{0.182} & 0.306 &
\first{0.182} & \second{0.304} &
\first{0.182} & \first{0.303} &
\first{0.182} & 0.305 \\
  & 336 & 0.341 & 0.426 
& 0.336 & \second{0.422} &
\second{0.332} & \first{0.417} &
\first{0.329} & \first{0.417}
 \\
  & 720 & 0.806 & 0.679 &
{0.792} & {0.670} &
\second{0.788} & \second{0.668} &
\first{0.783} & \first{0.666} 
 \\
  \cmidrule{2-10}
  & \rowcolor{gray!20} Avg. & 0.356 & 0.407 &
{0.350} & {0.401} &
\second{0.349} & \second{0.399} &
\first{0.346} & \first{0.398}
 \\
  \midrule
  \multirow{5}{*}{\shortstack{\\\\ECL\\($C=321$)}} & 96 & 0.147 & 0.239 & 
\first{0.140} & \first{0.236} & 
\first{0.140} & \first{0.236} & \second{0.141} & \second{0.237} \\

  & 192 &0.162 & 0.253 &
\first{0.157} & \second{0.251} &
\first{0.157} & \second{0.251} &
\first{0.157} & \first{0.250}
 \\
  & 336 & 0.175 & 0.269 &
\second{0.173} & \second{0.268} &
\second{0.173} & \second{0.268} &
\first{0.172} & \first{0.267}
 \\
  & 720 & 0.215 & 0.304 &
\first{0.203} & \second{0.297} &
\first{0.203} & \second{0.297} &
\first{0.203} & \first{0.296}
 \\
  \cmidrule{2-10}
 & \rowcolor{gray!20} Avg. & 0.175 & 0.266 &
\first{0.168} & \second{0.263} &
\first{0.168} & \second{0.263} &
\first{0.168} & \first{0.262}
 \\
 
\bottomrule
\end{NiceTabular}
\end{adjustbox}
\caption{Results of TS forecasting with TimeSiam.}
\label{tbl:timesiam}
\vspace{10pt}
\end{table*}

\vspace{20pt}
\section{Application to Moirai}
\label{sec:why_not_moirai}
Although Moirai \cite{woo2024unified} is a CI method, CM can still be adapted to it, as Moirai performs channel flattening followed by time-axis attention. The CM can be applied by reshaping it from $C \times C$ to $C \cdot P \times C \cdot P$ with duplicated intra-channel values, where $P$ is the number of patches. However, we exclude it for the following reasons:

\begin{itemize}[leftmargin=0.3cm,itemsep=0pt,topsep=0pt, partopsep=0pt]
    \item \textbf{Limited task coverage}: Moirai supports only forecasting task, whereas UniTS handles four tasks.
    \item \textbf{Reproducibility issues}: Its code lacks pretraining details, and our reproduced fine-tuning results were significantly worse than reported, which is also shared with many users in the official GitHub.
    \item \textbf{Resource constraints}: Moirai’s large data (231B) and model size (91M) make experiments infeasible under our resource limits.
\end{itemize}

\clearpage
\section{Lookback Window Size vs. Performance}
\begin{figure*}[h]
\vspace{15pt}
\centering
\includegraphics[width=.83\textwidth]{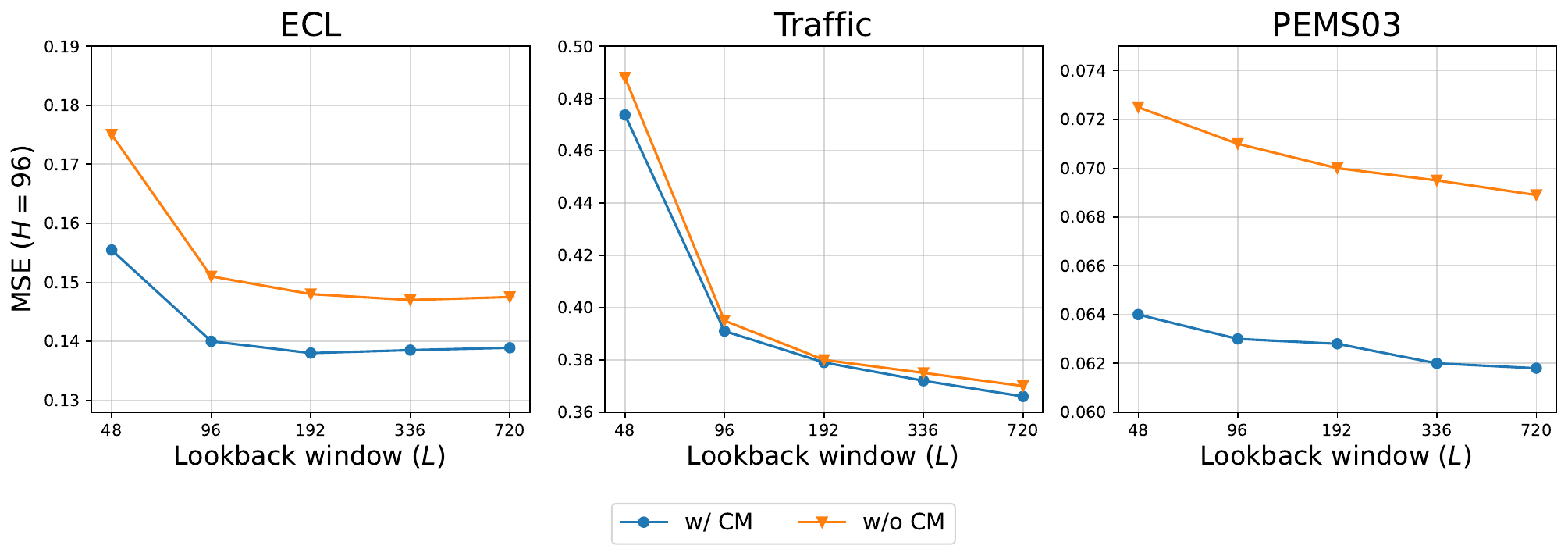} 
\caption{\textbf{Effect of CM under various lookback window sizes.} Forecasting performance with the lookback length $L \in \{48, 96, 192, 336, 720\}$ , with forecast horizon $H= 12$ for PEMS03 and $H = 96$ for other datasets.}
\label{fig:lookback}
\end{figure*}

Following the previous work~\cite{liu2023itransformer}, we conduct an experiment to evaluate the effect of varying the lookback window size ($L$) on performance, using three datasets: ECL~\cite{wu2021autoformer}, Traffic~\cite{wu2021autoformer}, and PEMS03~\cite{liu2022scinet} with iTransformer~\cite{liu2023itransformer} as the backbone. 
The results, shown in Figure~\ref{fig:lookback}, indicate that the effectivness of CM remains robust to the choice of $L$ for all three datasets.

\vspace{30pt}
\section{CM under Extreme Cases}
\begin{figure*}[h]
\vspace{15pt}
\centering
\includegraphics[width=.83\textwidth]{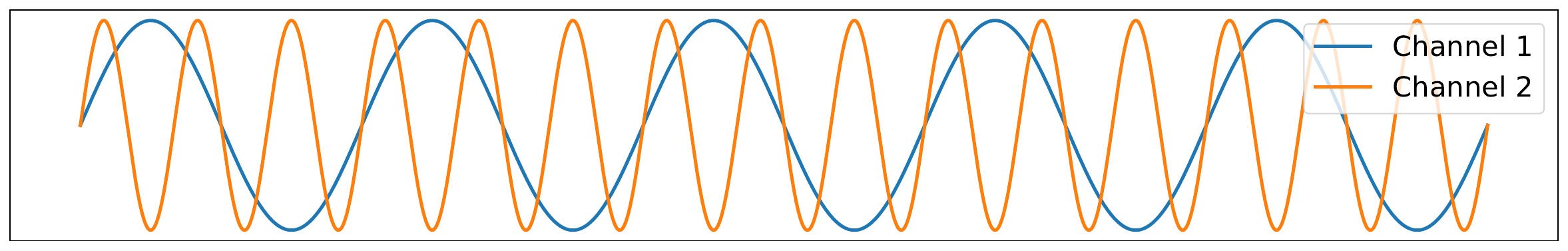} 
\caption{Synthetic dataset with two uncorrelated channels}
\label{fig:extreme}
\end{figure*}

To evaluate the effectiveness of CM under extreme cases, we design a scenario where the channels in TS exhibit no correlation. Specifically, we generate a synthetic TS dataset with two channels using sine waves oscillating at frequencies of 0.5 and 2.0 over a length of 18,000 (similar to ETTh~\cite{zhou2021informer}), as shown in Figure~\ref{fig:extreme}. 
We conduct TS forecasting using this dataset with iTransformer~\cite{liu2023itransformer} as the backbone, 
with an input window size and forecasting horizon of 96, following the experimental protocol used in ETTh1.
The result yields a CD ratio of CM approximately 0.018 and a forecasting MSE of around 0.0014, confirming strong channel independence and demonstrating the effectiveness of our method even under extreme CI conditions.

\clearpage
\section{Masked Channel Prediction}
\label{sec:appendix_mcp}
Tables~\ref{tbl:appendix_mcp1} and \ref{tbl:appendix_mcp2} show the results of masked channel prediction for five datasets~\cite{wu2021autoformer,liu2022scinet}, indicating significant improvement when a CM is applied to iTransformer compared to when it is not used.

\begin{table*}[h]
\vspace{20pt}
\centering
\begin{adjustbox}{max width=0.999\textwidth}
\begin{NiceTabular}{c|ccc|ccc}
\toprule
\multirow{4}{*}{Horizon} & \multicolumn{3}{c}{Exchange} & \multicolumn{3}{c}{ECL} \\
\cmidrule(lr){2-4} \cmidrule(lr){5-7} 
& \multicolumn{3}{c}{Avg. MSE(C1$\sim$C8)} & \multicolumn{3}{c}{Avg. MSE(C1$\sim$C321)}\\
\cmidrule(lr){2-4} \cmidrule(lr){5-7} 
& iTrans. & + CM &  Imp.
& iTrans. & + CM &  Imp.\\
\midrule
96 & 
0.139 & \cellcolor{gray!20} \first{0.138} & \first{1.2\%} &
0.846 & \cellcolor{gray!20} \first{0.526} & \first{37.8\%} 
\\
192 & 
0.236 & \cellcolor{gray!20} \first{0.232} & \first{1.5\%} &
0.849 & \cellcolor{gray!20} \first{0.563} & \first{33.7\%} 
\\
336 & 
0.383 & \cellcolor{gray!20} \first{0.374} & \first{2.4\%} &
0.861 & \cellcolor{gray!20} \first{0.594} & \first{31.0\%} 
\\
720 & 
0.934 & \cellcolor{gray!20} \first{0.917} & \first{1.8\%} &
0.891 & \cellcolor{gray!20} \first{0.741} & \first{16.8\%} 
\\
\midrule
Avg. & 0.423 & \cellcolor{gray!20} \first{0.415} & \first{1.8\%} 
& 0.862 & \cellcolor{gray!20} \first{0.606} & \first{29.7\%} \\
\bottomrule
\end{NiceTabular}
\end{adjustbox}
\caption{Results of masked channel prediction (Exchange, ECL).}
\label{tbl:appendix_mcp1}
\end{table*}

\begin{table*}[h]
\vspace{15pt}
\centering
\begin{adjustbox}{max width=0.999\textwidth}
\begin{NiceTabular}{c|ccc|ccc|ccc}
\toprule
\multirow{4}{*}{Horizon}  & \multicolumn{3}{c}{PEMS04} & \multicolumn{3}{c}{PEMS07} & \multicolumn{3}{c}{PEMS08}\\
\cmidrule(lr){2-4} \cmidrule(lr){5-7} \cmidrule(lr){8-10} 
& \multicolumn{3}{c}{Avg. MSE(C1$\sim$C307)} & \multicolumn{3}{c}{Avg. MSE(C1$\sim$C883)} & \multicolumn{3}{c}{Avg. MSE(C1$\sim$C170)}\\
\cmidrule(lr){2-4} \cmidrule(lr){5-7} \cmidrule(lr){8-10} 
& iTrans. & + CM &  Imp.
& iTrans. & + CM &  Imp.
& iTrans. & + CM &  Imp.\\
\midrule
12 & 
0.549 & \cellcolor{gray!20} \first{0.300} & \first{45.4\%} &
0.835 & \cellcolor{gray!20} \first{0.343} & \first{58.9\%} &
0.628 & \cellcolor{gray!20} \first{0.200} & \first{68.1\%}

\\
24 & 
0.718 & \cellcolor{gray!20} \first{0.351} & \first{51.1\%} &
0.865 & \cellcolor{gray!20} \first{0.448} & \first{48.1\%} &
0.678 & \cellcolor{gray!20} \first{0.241} & \first{64.5\%} 
\\
48 & 
0.750 & \cellcolor{gray!20} \first{0.409} & \first{45.5\%} &
1.038 & \cellcolor{gray!20} \first{0.511} & \first{50.8\%} &
1.197 & \cellcolor{gray!20} \first{1.059} & \first{11.5\%}
\\
96 & 
0.758 & \cellcolor{gray!20} \first{0.513} & \first{32.3\%} &
1.040 & \cellcolor{gray!20} \first{0.640} & \first{38.5\%} &
1.375 & \cellcolor{gray!20} \first{1.217} & \first{11.5\%}
\\
\midrule
Avg.  &  0.694 & \cellcolor{gray!20} \first{0.393} & \first{43.3\%} 
&0.945 & \cellcolor{gray!20} \first{0.486} & \first{48.6\%} 
&0.970 & \cellcolor{gray!20} \first{0.679} & \first{29.9\%} \\
\bottomrule
\end{NiceTabular}
\end{adjustbox}
\caption{Results of masked channel prediction (PEMS datasets).}
\label{tbl:appendix_mcp2}
\end{table*}

\end{document}